%% file: main.tex
\PassOptionsToPackage{table}{xcolor}
\documentclass{article}
\usepackage{iclr2027_conference,times}
\iclrfinalcopy
\input{math_commands.tex}

\usepackage[T1]{fontenc}
\usepackage[utf8]{inputenc}
\DeclareUnicodeCharacter{2203}{\ensuremath{\exists}}
\DeclareUnicodeCharacter{2200}{\ensuremath{\forall}}
\DeclareUnicodeCharacter{211D}{\ensuremath{\mathbb{R}}}
\DeclareUnicodeCharacter{2286}{\ensuremath{\subseteq}}
\DeclareUnicodeCharacter{2264}{\ensuremath{\leq}}
\DeclareUnicodeCharacter{2265}{\ensuremath{\geq}}
\DeclareUnicodeCharacter{2192}{\ensuremath{\to}}
\DeclareUnicodeCharacter{00AC}{\ensuremath{\lnot}}
\DeclareUnicodeCharacter{00D7}{\ensuremath{\times}}
\DeclareUnicodeCharacter{2295}{\ensuremath{\oplus}}
\DeclareUnicodeCharacter{2227}{\ensuremath{\wedge}}

\usepackage{hyperref}
\usepackage{url}
\hypersetup{
  hidelinks,
  pdftitle={TCSExam: Evaluating Frontier LLMs on the Full Range of TCS Research},
  pdfauthor={Yifan Luo, Zhengying Liu, Yang Yuan, Zhilin Yang, Andrew Chi-Chih Yao},
  pdfsubject={},pdfkeywords={}
}
\usepackage{booktabs}
\usepackage{array,tabularx}
\usepackage{longtable}
\usepackage{etoolbox}
\makeatletter
\patchcmd{\@maketitle}{\vskip 0.3in minus 0.1in}{\vskip 8pt}{}%
  {\PackageError{tcsexam}{Title spacing could not be applied}{Check the title macro.}}
\makeatother
\makeatletter
\patchcmd{\LT@makecaption}{\reset@font}{\reset@font\normalsize}{}%
  {\PackageError{tcsexam}{Longtable caption formatting could not be applied}{Check the longtable package definition.}}
\makeatother
\usepackage{multirow}
\usepackage{graphicx}
\usepackage{wrapfig}
\usepackage{float}
\usepackage{needspace}
\usepackage{enumitem}
\usepackage{pifont}
\usepackage{xcolor}
\newif\ifmarkarxivchanges
\markarxivchangesfalse
\ifdefined\TCSEXAMClean\markarxivchangesfalse\fi
\newcommand{\arxivadded}[1]{\ifmarkarxivchanges{\color{red}#1}\else#1\fi}
\newcommand{\tablehead}[2][c]{\begin{tabular}[c]{@{}#1@{}}#2\end{tabular}}
\newcommand{\qwenagent}{\arxivadded{Qwen3.8 Max (0902)}}
\newcommand{\deepseeknewagent}{\arxivadded{DeepSeek V4.1 Flash 1017}}
\newcommand{\deepseekagent}{\arxivadded{DeepSeek V4.1 Flash 0910}}

\newenvironment{arxivaddition}{\begingroup\ifmarkarxivchanges\color{red}\fi}{\endgroup}
\newcommand{\arxivfigure}[1]{\includegraphics[width=\linewidth]{#1.pdf}}
\newcommand{\projectrepository}{https://github.com/MoonshotAI/TCSExam}
\newcommand{\projectwebsiteurl}{https://moonshotai.github.io/TCSExam/}
\newcommand{\projectwebsite}{\ifdefempty{\projectwebsiteurl}%
  {\projectglobe\, Website {\color{red}\scriptsize[TODO]}}%
  {\href{\projectwebsiteurl}{\projectglobe\, Website}}}
\definecolor{projectlink}{HTML}{315C83}
\usepackage{tikz}
\usetikzlibrary{positioning,arrows.meta,fit,backgrounds,calc}
\newcommand{\projectglobe}{\tikz[baseline=-.55ex,scale=.13,line width=.6pt,color=black]{%
  \draw (0,0) circle (1);\draw (0,0) ellipse (.42 and 1);
  \draw (-1,0)--(1,0);\draw (-.87,.5)--(.87,.5);\draw (-.87,-.5)--(.87,-.5);}}
\newcommand{\trackmargin}[1]{\makebox[0pt][r]{%
  \smash{\raisebox{-.16em}{\includegraphics[height=1.05em]{figures/icons/track-#1.pdf}}}\hspace{.4em}}}
\newcommand{\trackheading}[2]{\noindent\trackmargin{#1}\textbf{#2}}
\DeclareRobustCommand{\trackinline}[1]{%
  \smash{\raisebox{-.13em}{\includegraphics[height=.9em]{figures/icons/track-#1.pdf}}}\hspace{.28em}}
\newcommand{\tracksubsection}[2]{\subsection[#2]{\trackinline{#1}#2}}
\newcommand{\trackparagraph}[2]{\paragraph[#2]{\trackinline{#1}#2}}
\newcommand{\projectcode}{\raisebox{-.12em}{\includegraphics[height=.95em]{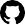}}}

\title{TCSExam: Evaluating Frontier LLMs on the\\Full Range of TCS Research}

\author{%
Yifan Luo\textsuperscript{1,2}\quad
Zhengying Liu\textsuperscript{2}\quad
Yang Yuan\textsuperscript{1,3}\quad
Zhilin Yang\textsuperscript{2}\quad
Andrew Chi-Chih Yao\textsuperscript{1,3}\\[4pt]
\normalfont\textsuperscript{1}Tsinghua University\quad
\textsuperscript{2}Moonshot AI\quad
\textsuperscript{3}Shanghai Qizhi Institute\\
\normalfont\texttt{luoyf24@mails.tsinghua.edu.cn}
}

\begin{document}
\raggedbottom
\maketitle
\fancyhead{}
\renewcommand{\headrulewidth}{0pt}

\noindent\makebox[\textwidth][c]{\small
  \ifmarkarxivchanges\color{red}\else\color{projectlink}\fi
  \projectwebsite\hspace{1em}\textperiodcentered\hspace{1em}%
  \href{\projectrepository}{\projectcode\, MoonshotAI/TCSExam}}\par
\vspace{3pt}

\begin{abstract}
\arxivadded{Can large language models (LLMs) advance theoretical computer science and resolve its long-standing open conjectures? To study progress toward this goal, we introduce TCSExam, a unified framework and benchmark covering the full range of TCS research. It evaluates \emph{theorem proving}, \emph{solver engineering}, and \emph{constructions} alongside direct attempts to resolve \emph{open conjectures}, capturing both concrete research accomplishments and progress toward long-standing research goals. TCSExam-Full contains 100 expert-reviewed tasks, with submissions checked by the Lean kernel or an executable task checker. TCSExam-Lite selects 20 tasks, five per track, reducing the number of task attempts by 80\% at the same per-attempt budget. We evaluate eight agents on Full and nine on Lite. The TCSExam Index combines all four tracks: GPT-6 Astra leads Full (71.9), and Opus 5.5 leads Lite (72.3). Full and Lite rankings are strongly aligned across eight shared agents (Spearman's $\rho=0.93$), supporting Lite as a compact, discriminative benchmark. The evaluations yield improved solver outputs and mathematical constructions, together with independently verified special cases and bounds from open-conjecture attempts. TCSExam provides an extensible framework with task sources, Docker build recipes, and evaluation tools for reproducible use.}
\end{abstract}

\input{sections/01-intro}
\input{sections/02-related}
\input{sections/03-benchmark}
\input{sections/04-setup}
\input{sections/05-results}
\input{sections/06-discussion}

\clearpage
\input{sections/ai-use}
\bibliography{refs}
\bibliographystyle{iclr2027_conference}

\appendix
\input{sections/appendix}

\end{document}

%% file: math_commands.tex
\usepackage{amsmath,amsfonts,bm}

\def\eqref#1{equation~\ref{#1}}

\def\1{\bm{1}}

\DeclareMathAlphabet{\mathsfit}{\encodingdefault}{\sfdefault}{m}{sl}
\SetMathAlphabet{\mathsfit}{bold}{\encodingdefault}{\sfdefault}{bx}{n}



%% file: sections/01-intro.tex
\section{Introduction}
\label{sec:intro}

Can large language models (LLMs) resolve long-standing \emph{open conjectures} and advance theoretical computer science? Studying progress toward this ambition poses a challenge for evaluation: when a conjecture remains unresolved, the final outcome alone reveals little about which research tasks an LLM can perform or where its limitations lie. An informative evaluation should therefore examine concrete research accomplishments alongside direct attempts to prove or refute open conjectures.

Three forms of research output provide concrete ways to examine these questions. \emph{Theorem proving} requires complete arguments for research-level mathematical statements. \emph{Solver engineering} produces reusable programs whose effectiveness is tested across problem instances. \emph{Constructions} provide explicit mathematical objects with verified properties that establish or improve bounds. Each output has independent research value, and the methods used to obtain them can interact: solvers search for candidate objects, proofs establish their properties, and constructions can supply witnesses or counterexamples. Evaluating these outputs helps TCS researchers judge which research work LLMs can undertake and helps model developers identify priorities for improvement.

\begin{figure}[t]
\setlength{\parskip}{0pt}
\setlength{\abovecaptionskip}{5pt}
\centering
\includegraphics[width=\linewidth]{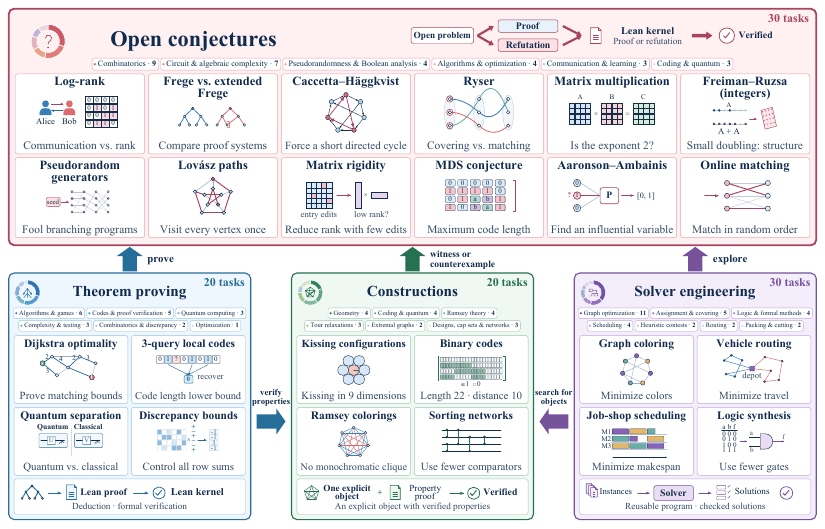}
\caption{TCSExam covers the full range of TCS research through four evaluation tracks. Arrows show how proving, searching, and constructing can contribute to one another and to resolving open conjectures. The examples illustrate tasks within each track; research areas can span multiple tracks.}
\label{fig:overview}
\end{figure}

Existing benchmarks address selected aspects of TCS research (Table~\ref{tab:positioning}). We introduce \textbf{TCSExam}, a unified framework and benchmark covering the full range of TCS research (Figure~\ref{fig:overview}). Its four tracks evaluate theorem proving, solver engineering, constructions, and open-conjecture settlement, with success criteria defined by the requested outputs. Research areas can span multiple tracks, and shared task interfaces and verification protocols support the incorporation of additional problems.

TCSExam comprises 100 tasks: 20 research theorems to prove in Lean 4 \citep{lean4}, 30 solver-engineering problems evaluated over instance families, 20 construction problems scored against documented records, and 30 open conjectures in prove-or-refute form. Authors with expertise in the relevant TCS areas reviewed all 100 tasks for research significance. Every submission is checked by the Lean kernel or an executable task checker. This combines expert judgment about which problems matter with reproducible verification of the submitted results.
\arxivadded{For smaller-scale evaluations, TCSExam-Lite provides a 20-task subset with five tasks per track, selected to cover diverse topics and performance profiles within the Full suite.}

\arxivadded{We evaluate eight agents on Full and nine on Lite.} On 20 research theorems without supplied intermediate lemmas, \arxivadded{GPT-6 Astra achieves a per-attempt complete-proof success rate of 70.0\%, followed by Kimi K3.1 Pro at 38.3\%; the other six agents achieve at most 11.7\%.} Solver performance varies across instance families, with submitted programs improving on public competition circuits. \deepseekagent{} produces a verified construction that exceeds a task reference despite completing none of its 60 theorem-proving attempts. Individual attempts also yield independently verified special cases and bounds (Section~\ref{sec:openresults}).

\arxivadded{Lite extends the comparison to Opus 5.5, which solves four of its five theorem tasks, compared with three for Astra. Full and Lite rankings agree on 26 of 28 pairs among the eight shared agents, while track-level comparisons reveal differences in their research capabilities (Section~\ref{sec:literesults}).}

Our contributions are as follows.
\begin{itemize}[leftmargin=1.5em,itemsep=2pt]
\item \textbf{A framework spanning the full range of TCS research.} Four tracks evaluate research outputs and open-conjecture settlement through shared task interfaces, distinct success criteria, and a common protocol for incorporating additional problems.
\item \textbf{An expert-reviewed benchmark with verifiable outcomes.} We curate 100 research tasks \arxivadded{and a 20-task Lite subset} with track-specific scoring and verification by the Lean kernel or executable task checkers.
\item \arxivadded{\textbf{A systematic evaluation of nine agents across Full and Lite.}} We characterize performance across the four tracks and examine submitted proofs, programs, and constructions to identify research accomplishments and remaining limitations.
\end{itemize}

%% file: sections/02-related.tex
\section{Related Work}
\label{sec:related}

\textbf{Theorem proving in mathematics and TCS.} TCS-BENCH \citep{tcsbench2026} evaluates research-level natural-language proofs using an LLM verifier. Despite reported accuracy above 90\% against expert judgments, acceptance remains subject to verifier errors and provides no formal correctness guarantee. FormalTCS \citep{formaltcs2026} evaluates end-to-end formalization and proving of TCS results. Their benchmark scope does not include reusable solver performance or construction quality. Formal evaluation also spans competition mathematics, undergraduate mathematics, and repository-context proofs \citep{minif2f,putnambench,proofnet,leandojo,minictx}. CombiBench provides formal combinatorial problems in Lean 4 \citep{combibench2025}, while LeanGeo-Bench evaluates competition geometry in the LeanGeo framework \citep{leangeo2025}. Kimina-Prover uses reinforcement learning to train structured reasoning for Lean 4 proof generation \citep{kiminaprover2025}. Proof guidance and decomposition are studied by Draft, Sketch, and Prove, DeepSeek-Prover-V2, and Seed-Prover \citep{draftsketchprove,deepseekproverv2,seedprover}.

\textbf{Solver programs and explicit constructions.} AlphaTensor and AlphaDev demonstrate algorithm discovery through reinforcement learning \citep{alphatensor,alphadev}; FunSearch and AlphaEvolve use program search for mathematical and computational improvements \citep{funsearch,alphaevolve}. FrontierCS evaluates executable programs on open-ended algorithmic and research tasks \citep{mang2026frontiercs,frontiercs1}. \arxivadded{FrontierCS 2 compares agents' reimplementations of research contributions with the original code across 48 public tasks \citep{mang2026frontiercs2benchmarkcomputer}.} EdgeBench includes formal proving and combinatorial optimization in long-interaction evaluation \citep{edgebench2026}. \arxivadded{TCSExam separately evaluates reusable solvers across instance families and the parameters attained by explicit constructions.}

\textbf{Open conjectures and research-level evaluation.} FrontierMath evaluates advanced mathematical reasoning through automatically checkable answers \citep{frontiermath}. Formal Conjectures supplies formalized research problems and frozen evaluations of verified discovery \citep{formalconjectures2026}; the proving-ground repository reports complete solutions and verified reductions on its open tier \citep{provingground2026}. TCSExam evaluates complete proof or refutation of open TCS conjectures, retaining intermediate outputs for separate analysis.

\textbf{Positioning of TCSExam.} Research-agent benchmarks also evaluate ML engineering and research replication \citep{mlebench,rebench,paperbench}. Table~\ref{tab:positioning} compares the closest TCS benchmarks. TCSExam brings four forms of TCS research output into one extensible framework, with distinct success criteria and common procedures for task development, verification, and version control.

\begin{table}[t]
\centering
\newcommand{\coverageyes}{\textcolor[HTML]{2E7D32}{\ding{51}}}
\newcommand{\coverageno}{\textcolor[HTML]{A0A0A0}{\ding{55}}}
\newcommand{\coveragerelated}{\textcolor[HTML]{667788}{$\circ$}}
\newcommand{\coveragehead}[2][c]{{\renewcommand{\arraystretch}{1}\bfseries\begin{tabular}[c]{@{}#1@{}}#2\end{tabular}}}
\caption{\arxivadded{Coverage of selected benchmarks.}}
\label{tab:positioning}
\small
\setlength{\tabcolsep}{4pt}
\renewcommand{\arraystretch}{1.0}
\begin{tabular*}{\linewidth}{@{\extracolsep{\fill}}lrcccc@{}}
\toprule
\coveragehead[l]{Benchmark / track} & \coveragehead[r]{\# Tasks} & \coveragehead{Theorem\\proving} & \coveragehead{Solver\\engineering} & \coveragehead{Constructions} & \coveragehead{Open\\conjectures} \\
\midrule
TCS-BENCH & 300 & \coverageyes & \coverageno & \coverageno & \coverageno \\
FormalTCS & 143 & \coverageyes & \coverageno & \coverageno & \coverageno \\
FrontierCS & 240 & \coverageno & \coverageyes & \coverageyes & \coverageno \\
\arxivadded{FrontierCS 2} & \arxivadded{48} & \coverageno & \coveragerelated & \coverageno & \coverageno \\
EdgeBench-Formal & 13 & \coverageyes & \coverageno & \coverageno & \coveragerelated \\
EdgeBench-Optimization & 19 & \coverageno & \coverageyes & \coveragerelated & \coverageno \\
\midrule
\textbf{TCSExam} & \textbf{100} & \coverageyes & \coverageyes & \coverageyes & \coverageyes \\
\bottomrule
\end{tabular*}
\par\smallskip
\begin{minipage}{\linewidth}
\footnotesize
\coverageyes\ evaluated; \coverageno\ outside scope; \coveragerelated\ related tasks with a different scope.\par
\coveragerelated: \arxivadded{code optimization (FrontierCS 2); }\arxivadded{machine-specific halting proofs and graph coloring (EdgeBench).}
\end{minipage}
\end{table}

%% file: sections/03-benchmark.tex
\section{TCSExam}
\label{sec:benchmark}

TCSExam organizes tasks across TCS areas by their required outputs and success criteria into the four tracks in Figure~\ref{fig:overview}. Search can produce candidate objects, proofs establish their properties, and both can contribute to resolving open conjectures. Appendix~\ref{app:tasklist} lists the 100 tasks and their sources.

\subsection{Research scope and task interfaces}
\label{sec:overview}
\label{sec:composition}

\trackheading{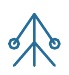}{Theorem proving (20 tasks).} Given a formal statement derived from a research theorem, the model submits a Lean 4 proof. We vary the supplied context through three settings with a common target: \emph{bare} provides the main statement, fixed definitions, and mathematical assumptions; \emph{ladder} additionally provides intermediate lemma statements; \emph{guided} further supplies proof ideas. This design is related to TCS-BENCH's use of intermediate-result masking to vary proof-task difficulty \citep{tcsbench2026}. Our main evaluation uses bare to measure complete-proof success. Ladder and guided support a separate analysis of proof guidance, with partial checkpoint credit and independent complete-proof checks (Section~\ref{sec:scoring}).

\trackheading{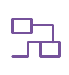}{Solver engineering (30 tasks).} The model receives a problem specification, a naive baseline, and public instances, then submits a reusable solver program. Final judging evaluates the program across the task's designated instance set under task-specific resource limits. Checkers validate solutions or certificates and recompute objective values.

\trackheading{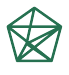}{Constructions (20 tasks).} The model submits an explicit mathematical object and a certificate of its required properties. Examples include codes, geometric configurations, and graph colorings. Tasks use Lean 4 formalizations and executable checks to verify the object and its achieved parameter. Documented construction records provide reference values against which progress is measured; the model may use mathematical reasoning, search code, or both to produce its submission.

\trackheading{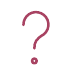}{Open conjectures (30 tasks).} The model receives a formalized conjecture and two Lean proof slots, \texttt{conj} and \texttt{refuted}, for the statement and its negation. A submission must prove either side in full. A counterexample must come with a proof that it refutes the stated conjecture. This interface accepts either direction of a resolution.

\subsection{Task curation and development}
\label{sec:tasks}

We select tasks for their \textbf{research significance}, \textbf{coverage across TCS areas}, and \textbf{verifiable outcomes}. Theorem tasks come from recent TCS papers (2023--2026) and span 13 subfields. Solver tasks draw on established competitions and instance libraries. Task-specific safeguards, including withholding hidden instances from the agent during development and applying structure-preserving transformations, mitigate direct reuse of stored answers. Construction tasks target open problems with documented records, while the open-conjecture track includes problems such as log-rank and Ryser. Expert review assesses the significance and representativeness of the selected problems.

\begin{figure}[t]
\centering
\includegraphics[width=\linewidth]{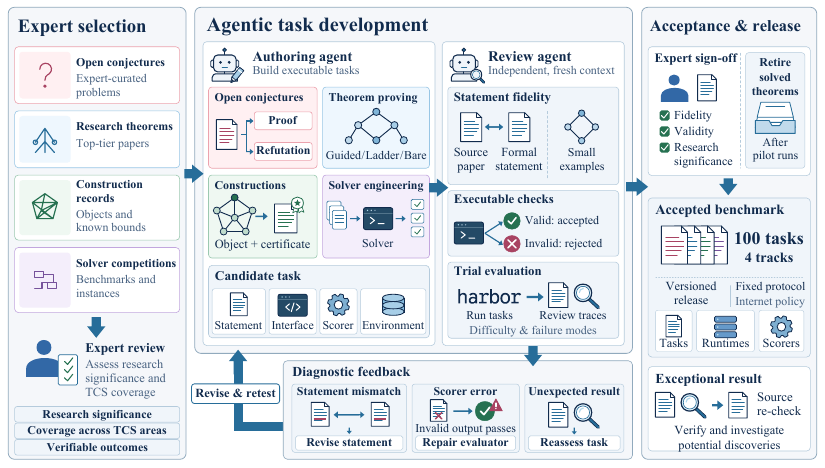}
\caption{Task curation and validation. Expert selection is followed by agent-assisted authoring and review, executable checks, and trial evaluation. Statement mismatches, scorer errors, and unexpected results prompt revision and re-testing. Tasks are included after expert approval; exceptional results trigger verification and source re-checks.}
\label{fig:construction}
\end{figure}

Figure~\ref{fig:construction} summarizes the task development process. An authoring agent prepares the statement, submission interface, scorer, and environment. A separate review agent checks correspondence to the research source and the behavior of the verification procedure. Tests and pilot evaluations identify defects and guide revision. Across development rounds, over 40 theorem problems were retired after being solved in pilot evaluations. Tasks are included after expert approval.

\subsection{Track-specific scoring}
\label{sec:scoring}

Scores reflect the submitted output and the task's notion of success. Complete proof obligations use binary scores; optimization and construction tasks provide graded credit. The score definitions and integrity checks are specified per track (Appendix~\ref{app:scoring}). \arxivadded{Reported scores use percentages rounded to one decimal; formulas use the original score units.}

\noindent\textbf{Theorem proving.} A valid proof of the complete bare-level statement scores \arxivadded{100.0\%}; an incomplete proof scores \arxivadded{0.0\%}. Guided and ladder tasks receive partial credit for the fraction of checkpoints passed; a verified complete proof receives full credit. Each checkpoint is checked in isolation with the other checkpoint statements supplied as assumptions; this also applies to the main-theorem checkpoint. Coverage therefore measures conditional proof obligations. Complete proofs are checked separately without unproved checkpoint assumptions (Appendix~\ref{app:scoring-lean}).

\noindent\textbf{Solver engineering.} Optimization tasks measure the fraction of the difference between a naive baseline $B$ and a task reference $K$ recovered by the submitted solver. These scores are not clipped: values below \arxivadded{0.0\%} indicate performance worse than the baseline, and values above \arxivadded{100.0\%} indicate improvement over the reference. Decision tasks use solve rates. A separate Kimi K3-relative score standardizes raw instance outcomes against repeated Kimi K3 runs and applies a Gaussian cumulative-distribution transform (Appendix~\ref{app:scoring-solver}). Scores are averaged over instances, then attempts within tasks, then tasks. Improvements over $K$ undergo artifact and source verification before being described as new records.

\noindent\textbf{Constructions.} Scores measure attainment relative to a trivial construction and a documented record: matching the record scores \arxivadded{100.0\%}, and surpassing it can score above \arxivadded{100.0\%}. Task-specific shaping is cubic for 14 tasks and linear for the remainder. We report the achieved mathematical parameters alongside these scores, since the score transformation changes the numerical size of an improvement. A record provides a comparison point even when the optimum remains unknown.

\noindent\textbf{Open conjectures.} A verified proof of either the statement or its negation scores \arxivadded{100.0\%}. All other submissions score \arxivadded{0.0\%} on the resolution criterion. \arxivadded{We also analyze intermediate bounds, special cases, and counterexamples to related statements alongside the resolution score.}

\begin{arxivaddition}
\textbf{TCSExam Index.} Let $P$, $S$, $C$, and $O$ be the mean theorem, Solver, construction, and open-conjecture scores, respectively, on their original scales. The Solver input $S$ is the relative score in $[0,1]$. We define
\begin{equation}
I=30P+30S+30C+10O.
\label{eq:index}
\end{equation}
The three tracks with graded observed performance receive equal weight; open-conjecture resolution contributes the remaining ten points. These weights are the same for Full and Lite. Construction scores above 1 remain unchanged, so the Index can exceed 100. We report the four track means alongside the Index to show which capabilities account for an aggregate difference.
\end{arxivaddition}

\begin{arxivaddition}
\subsection{TCSExam-Lite}
\label{sec:lite}
Lite contains five tasks per track, selected from Full.\footnote{\arxivadded{The task list was fixed before evaluating Kimi K3.1 Pro and \deepseeknewagent{}.}} Selection considered subject coverage, difficulty, execution cost, and the performance profiles of the original six agents. The theorem subset spans additive combinatorics, games, coding theory, circuit complexity, and quantum cryptography. Solver tasks cover packing, scheduling, graph optimization, resource assignment, and certified Boolean reasoning. The construction and open-conjecture subsets likewise cover several mathematical objects and research areas (Appendix~\ref{app:lite}).

Every selected task retains its original scoring function, and the five task means receive equal weight within each track. Lite uses the same 12-hour development budget per attempt: one attempt on each task has a maximum development allocation of 240 hours, compared with 1,200 hours for Full. Task execution can run in parallel, and final judging has its own resource budget. Both Solver protocols use public instances for development. Full provides a parallel self-test service; Lite runs self-tests in the development environment, simplifying deployment. Both use the same final judging rules. Section~\ref{sec:setup} details these protocols. Section~\ref{sec:literesults} compares the observed rankings under the two suite protocols.
\end{arxivaddition}

\subsection{Task and result validation}
\label{sec:fidelity}

Before a task enters the benchmark, task validation checks whether the encoded problem matches its research source. After submission, result validation checks outputs against the task rules, with additional review for candidate advances. Appendix~\ref{app:fidelity} documents these checks and supporting evidence.

\textbf{Formal statement fidelity.} Agent-assisted reviews compare definitions, assumptions, quantifiers, and parameter ranges in the Lean statement with the source. Small examples and boundary cases test for degenerate or unintended interpretations, and re-translation provides an additional cross-check. Shared definitions and statements are checked across disclosure levels. Successful proofs and unexpected outcomes prompt renewed scrutiny of the statement and its assumptions. These checks also flagged potential inconsistencies in source statements or proofs; candidate tasks with unresolved concerns were excluded.

\textbf{Result validation.} Every submission is checked by the Lean kernel or an executable task checker. Candidate advances additionally undergo agent-assisted audits of the rollouts and submitted artifacts. These audits check compliance with task rules, proof dependencies or object validity, and the conditions of the claimed improvement. For solver results, this includes comparing instance sets and execution budgets. Source checks distinguish improvements over task references from new literature records; claims of new records require evidence for the prior best value and a matching problem setting.

\textbf{Task versions and extensions.} Each reported result is associated with the task statement, environment, verifier, and reference values used in evaluation. Corrections and additional problems enter new task versions, with changes and affected results recorded. Earlier versions and evaluation records are retained so that results remain interpretable as the benchmark expands.

%% file: sections/04-setup.tex
\section{Experimental Setup}
\label{sec:setup}

\begin{arxivaddition}
\textbf{Agents and execution.} Each evaluated agent combines an LLM with an execution harness that manages tool use and interaction with the environment. Full includes GPT-6 Astra \citep{gpt6astra}, Kimi K3 \citep{kimik3}, GLM-5.3 \citep{glm53}, \qwenagent{} \citep{qwen38max}, Grok 4.6 \citep{grok46}, \deepseekagent{} and \deepseeknewagent{} \citep{deepseek41flash}, and Kimi K3.1 Pro \citep{kimik31pro}. Lite includes these eight agents and Opus 5.5 \citep{opus55}. The configurations appear in Appendix~\ref{app:protocol}.

\textbf{Tasks and resources.} Full contains 20 bare-level theorems, 30 Solver problems, 20 constructions, and 30 open conjectures; Lite contains five tasks per track. Kimi K3, Kimi K3.1 Pro, and both DeepSeek versions contribute three attempts per task; the other agents contribute one. An attempt is one agent run on one task, with a 12-hour wall-clock development budget. Theorem and Solver development have no general Internet access. Construction and open-conjecture tasks permit literature access. Tasks use the Harbor format \citep{harbor} to package environments, submission interfaces, and verifiers. Construction prompts provide published reference bounds. Solver judging uses separate per-instance or shared-pool budgets (Appendix~\ref{app:solvertasks}).

\noindent\textbf{Solver development protocols.} Both protocols provide the same public instances and self-test information during development. Full uses Online: an external service evaluates submitted solvers, parallelizing independent instances to accelerate iteration. Lite uses Offline: agents run public self-tests in their development environment, with parallelism limited by its resources. This removes the need to deploy a parallel judging service. Both protocols reserve hidden-set judging for the final submission and retain the same input split, checker, final execution limits, and fixed Kimi K3 reference. Online and Offline describe how public self-tests run; task network policies are unchanged.

\textbf{Score aggregation and shared tasks.} We first average attempts within each task, then average tasks equally within a track. Theorem success rates divide complete proofs by all scheduled attempts: 20 or 60 on Full, and 5 or 15 on Lite. The Index combines the four track means using Equation~\ref{eq:index}. The 15 non-Solver tasks shared by Full and Lite reuse the same submissions for each of the eight common agents. Solver results come from separate evaluations under each suite's development protocol. Opus contributes 20 Lite results. Appendix~\ref{app:lite-provenance} describes these result sets.
\end{arxivaddition}

\textbf{Proof-guidance comparison.} We compare Kimi K3 and \deepseekagent{} on the same 20 theorems, with three attempts per theorem in each setting. Bare supplies the target statement and definitions; ladder adds intermediate lemma statements; guided also supplies proof ideas. Complete-proof success requires a verified assembled or direct proof without unproved checkpoint assumptions. We report success per attempt and the number of theorems solved at least once; ladder and guided also receive partial checkpoint credit (Appendix~\ref{app:guidancepopulation}).

%% file: sections/05-results.tex
\section{Results}
\label{sec:results}

\subsection{Overall performance}
\label{sec:profiles}

\begin{figure}[t]
\centering
\arxivfigure{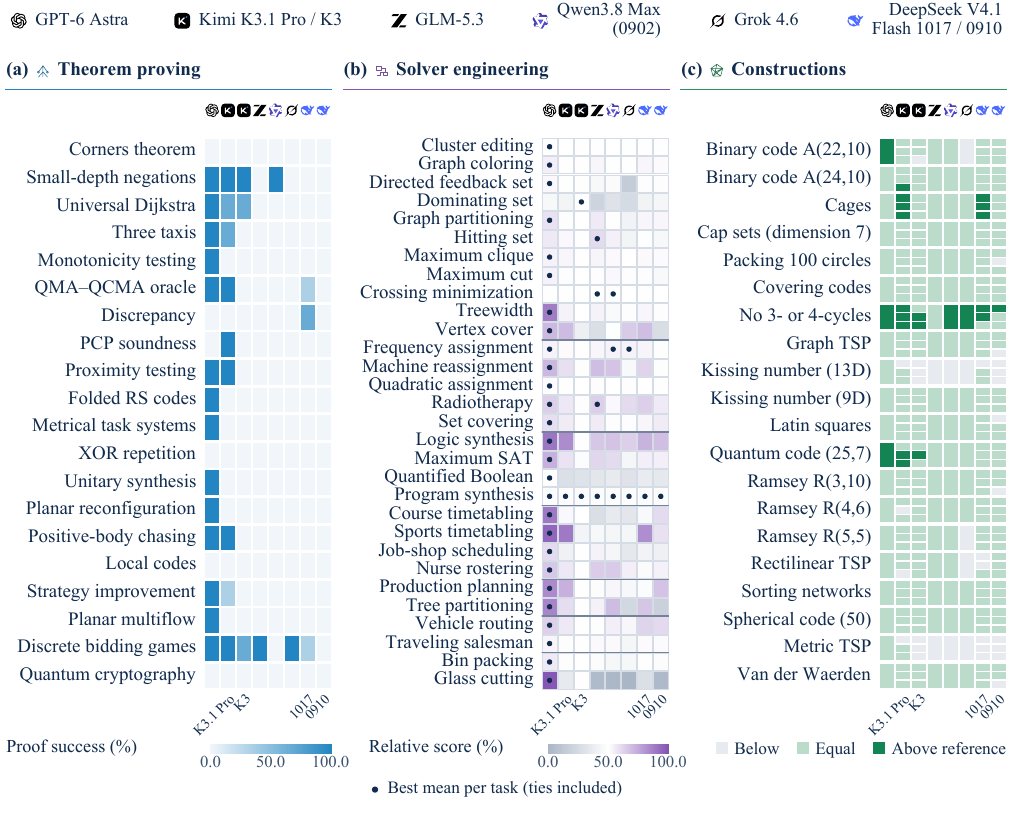}
\caption{\arxivadded{\textbf{Full task-level results for eight agents.} Rows are tasks; columns are agents. (a) Complete-proof success rates. (b) Kimi K3-relative Solver means; dots mark task-best means, including ties. (c) Cells stack one to three construction attempts vertically, colored below, at, or above the 100.0\% task reference (relative tolerance $10^{-6}$). Scores appear in Appendix~\ref{app:results}.}}
\label{fig:taskdistributions}
\end{figure}

Table~\ref{tab:mainresults} summarizes performance across the four tracks. Figure~\ref{fig:taskdistributions} shows task-level results for theorem proving, solver engineering, and constructions.

\begin{table}[t]
\centering
\begin{arxivaddition}
\input{tables/main-results-arxiv}
\end{arxivaddition}
\end{table}

\arxivadded{\textbf{Research proofs distinguish the evaluated agents.} GPT-6 Astra completes 14 of 20 theorems (70.0\%) without supplied intermediate lemmas. Kimi K3.1 Pro succeeds in 23 of 60 attempts (38.3\%), solving nine distinct theorems. It completes all three attempts on the PCP alphabet-soundness task, which Astra does not solve. Kimi K3 succeeds in 7 of 60 attempts (11.7\%), covering three theorems. GLM-5.3, \qwenagent{}, and Grok 4.6 each succeed in 1 of 20 attempts (5.0\%); \deepseekagent{} succeeds in none of 60, while \deepseeknewagent{} succeeds in 4 of 60 attempts (6.7\%) across three theorems. Across the eight agents, four theorems remain unproved.}

\textbf{Construction means reflect reference attainment across tasks.} Astra has the highest mean at \arxivadded{104.9\%}, matching the reference on 17 tasks and exceeding it on three. \arxivadded{The other agents range from 86.2\% to 98.6\%, with Kimi K3.1 Pro second. GLM-5.3 averages 97.7\% without exceeding any reference, illustrating that broad reference attainment and individual improvements are distinct aspects of performance.} Section~\ref{sec:outputs} examines the verified improvements on specific tasks.

\tracksubsection{solver}{Solver performance across instance families}
\label{sec:solverresults}

\textbf{Performance differs across problem families.} On each instance, a relative score of \arxivadded{50.0\%} denotes a valid outcome equal to the Kimi K3 reference mean. \arxivadded{Astra averages 64.7\%, followed by Kimi K3.1 Pro at 54.1\% and \deepseeknewagent{} at 53.1\%. The other five agents range from 43.8\% to 49.2\%} (Table~\ref{tab:mainresults}). Astra has the highest unrounded task mean, including ties, on 27 of 30 problems (Figure~\ref{fig:taskdistributions}b). It trails GLM-5.3 on hitting set, Kimi K3 on dominating set, \arxivadded{and GLM-5.3, \qwenagent{}, Kimi K3.1 Pro, and \deepseeknewagent{} marginally on crossing minimization.} Appendix~\ref{app:solverdiagnostics} gives detailed results.

\textbf{Relative scoring limits baseline--reference scaling effects.} The original score $(B-A)/(B-K)$ compares objective $A$ with baseline $B$ and reference $K$. A large $|B-K|$ compresses score differences near $K$; a small denominator can amplify individual contributions. We introduce a Kimi K3-relative metric that compares raw outcomes with repeated Kimi K3 results, retains sensitivity near their mean, and bounds contributions to $[0,100]$ on the reported scale. For example, \qwenagent{}\textquotesingle s valid IWLS 2023 circuit for \texttt{ex37} uses 273,020 AND gates, versus baseline 698 and reference 106. The ratio $(698-273{,}020)/(698-106)\approx-460$ allows this single oversized circuit to depress the original mean. \qwenagent{} averages \arxivadded{64.7\%} on that scale and \arxivadded{49.2\%} on the bounded relative scale. Appendix~\ref{app:scoring-solver} defines the metric; Appendix~\ref{app:solverdiagnostics} gives raw outcomes.

\textbf{Solver improvements across a benchmark family.} On the IWLS 2023 logic-synthesis task, Astra's submitted program produces valid circuits on all 100 evaluation instances: it improves the task reference on 22, matches it on 31, and uses more AND gates than the reference on 47. \arxivadded{In the original six-agent case-study cohort, each of the other agents' submitted programs also passes all 100 validity checks}, but only one Kimi K3 attempt and Grok 4.6 improve a reference, each on one instance. These counts describe the same program applied across a family of inputs. Appendix~\ref{app:solvercases} gives the circuit-level improvements and reference sources.

\begin{arxivaddition}
\subsection{Lite results and consistency with Full}
\label{sec:literesults}

\begin{table}[t]
\centering
\begin{arxivaddition}
\input{tables/lite-results}
\end{arxivaddition}
\end{table}

\textbf{Lite compares nine agents across all four tracks.} In Table~\ref{tab:literesults}, Opus 5.5 completes four of five theorem tasks and has an Index of 72.3; Astra completes three and scores 67.0. Their construction means are equal at 100.4\%, while Astra has the higher Solver score, 63.0\% versus 60.5\%. Kimi K3.1 Pro scores 53.5, followed by Kimi K3 at 48.6. All five Lite open conjectures remain unresolved.

\textbf{Lite closely matches Full in aggregate ranking.} Across the eight shared agents, Spearman's rank correlation is $\rho=0.93$, and 26 of 28 agent pairs have the same order (Figure~\ref{fig:full-lite-index}). The top three remain Astra, Kimi K3.1 Pro, and Kimi K3. Kimi K3.1 Pro scores 57.4 on Full and 53.5 on Lite. The two reversed pairs involve \deepseeknewagent{}, which ranks above GLM-5.3 and \qwenagent{} on Full and below them on Lite. The other seven agents retain their relative order.

\begin{figure}[H]
\setlength{\parskip}{0pt}
\setlength{\abovecaptionskip}{5pt}
\centering
\arxivfigure{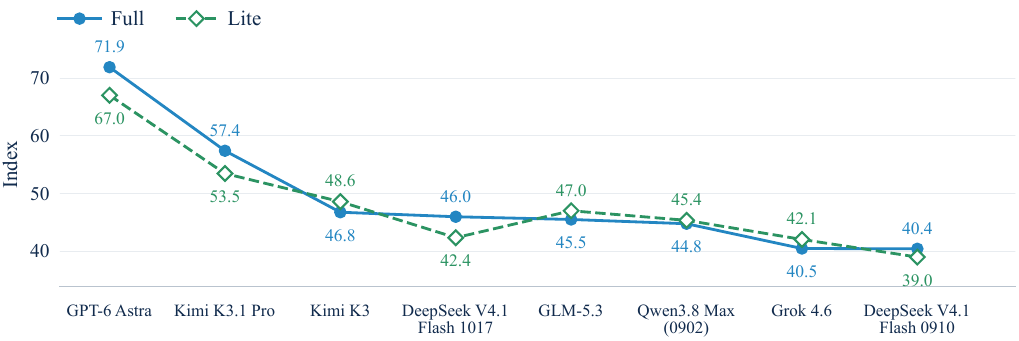}
\vspace{-8pt}
\caption{\arxivadded{\textbf{Full and Lite rankings are strongly aligned.} Among the eight shared agents, both suites give the same top three and agree on 26 of 28 pairs ($\rho=0.93$). Agents are ordered by Full Index.}}
\label{fig:full-lite-index}
\end{figure}

\textbf{Track scores show where the suites differ.} Among the 28 shared-agent pairs, theorem rankings reverse three, Solver rankings six, and construction rankings three. The two conjecture tasks with accepted refutations on Full are outside the Lite subset, where all agents score zero. These comparisons show how differences in task coverage and Solver self-test execution accompany changes in individual track rankings, while the overall ordering remains closely aligned (Appendix~\ref{app:lite-ranking}).
\par
\end{arxivaddition}

\tracksubsection{theorem}{Understanding proof performance}
\label{sec:proofdiagnostics}
\label{sec:trackfindings}

\textbf{More theorems are solved with additional statements and proof ideas.} With three attempts on each of 20 theorems, Kimi K3 solves 3, 4, and 7 theorems at least once under bare, ladder, and guided, respectively; \deepseekagent{} solves 0, 2, and 6. Guided also has the highest per-attempt complete-proof success: Kimi K3 succeeds in 18/60 attempts, compared with 7/60 under both ladder and bare. \deepseekagent{} succeeds in 0/60 bare, 2/60 ladder, and 11/60 guided attempts. Appendix~\ref{app:checkpoints} reports complete-proof outcomes, partial-credit scores, and individual proof attempts.

\tracksubsection{construction}{Verified construction improvements}
\label{sec:outputs}
\label{sec:behavior}

\begin{figure}[t]
\centering
\includegraphics[width=\linewidth]{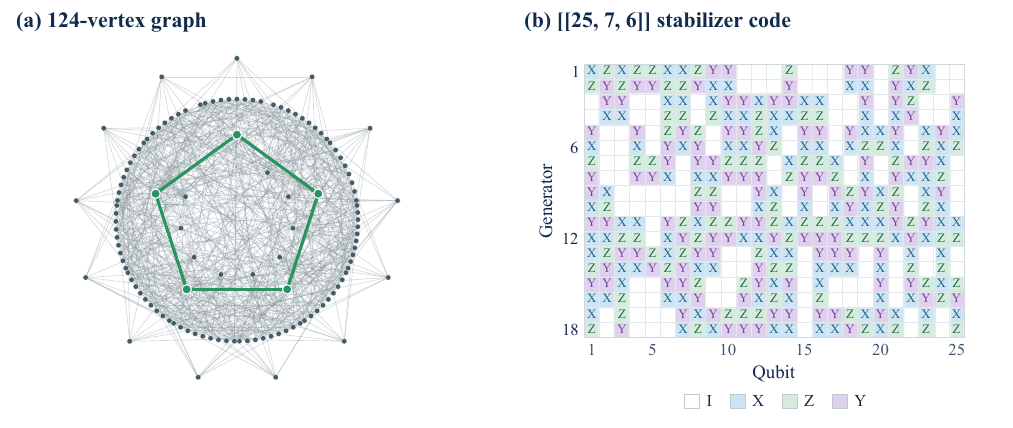}
\caption{\textbf{Two constructions submitted by Kimi K3.} (a) All 124 vertices and 630 edges of a graph with no 3- or 4-cycles, exceeding the 629-edge task reference. Green highlights a shortest cycle. (b) The 18 stabilizer generators of a $[[25,7,6]]$ quantum code, improving the reference distance from 5 to 6. Each row specifies an operator on 25 qubits; white cells denote the identity $I$. Both panels are drawn directly from submitted certificates.}
\label{fig:researchoutputs}
\end{figure}

\textbf{Graph and code results.} In the Erd\H{o}s girth-5 extremal-graph task, Kimi K3 constructs a 124-vertex, 630-edge graph without 3- or 4-cycles, exceeding the 629-edge reference \citep{girthbounds}. In the quantum stabilizer-code task, it constructs a binary $[[25,7,6]]$ code, increasing the minimum distance from 5 to 6 for a code on 25 qubits encoding seven logical qubits \citep{codetables25} (Figure~\ref{fig:researchoutputs}). Astra also attains both parameters and proves that a binary code of length 22 and minimum Hamming distance 10 has at most 82 codewords ($A(22,10)\leq82$), improving the published upper bound of 84 \citep{gijswijt2012} (Appendix~\ref{app:constructionprotocol}). \qwenagent{}, Grok 4.6, and \deepseekagent{} each obtain a 630-edge graph. Appendix~\ref{app:cases} records provenance and shows Astra's constructions.

\tracksubsection{open}{Open conjectures within the evaluation budget}
\label{sec:openresults}

\arxivadded{\textbf{Use of published counterexamples.} During the network-enabled evaluations, Kimi K3.1 Pro and \deepseeknewagent{} retrieved OpenAI's published counterexamples and Lean proofs for Ryser and the modified integer round-up conjecture (MIRUP) \citep{openaiMathRyser,openaiMathBinPacking}.\footnote{\arxivadded{The tasks were formulated before these \href{https://openai.com/index/sharing-ai-progress-in-mathematics/}{OpenAI results} were released; the Kimi K3.1 Pro and DeepSeek 1017 evaluations took place afterward.}} Their submissions adapt these existing results to the task definitions. The accepted adaptations comprise Ryser submissions from Kimi K3.1 Pro (1/3) and DeepSeek 1017 (2/3), and MIRUP submissions from DeepSeek 1017 (3/3; Table~\ref{tab:full-open}). The other 28 conjecture tasks have no accepted complete solution. OpenAI's results highlight the potential for AI-assisted conjecture resolution. Future work could evaluate agents without Internet access on conjectures still open at the time of testing.}

\begingroup
\setlength{\intextsep}{6pt}
\begin{figure}[H]
\setlength{\parskip}{0pt}
\setlength{\abovecaptionskip}{0pt}
\centering
\includegraphics[width=.86\linewidth]{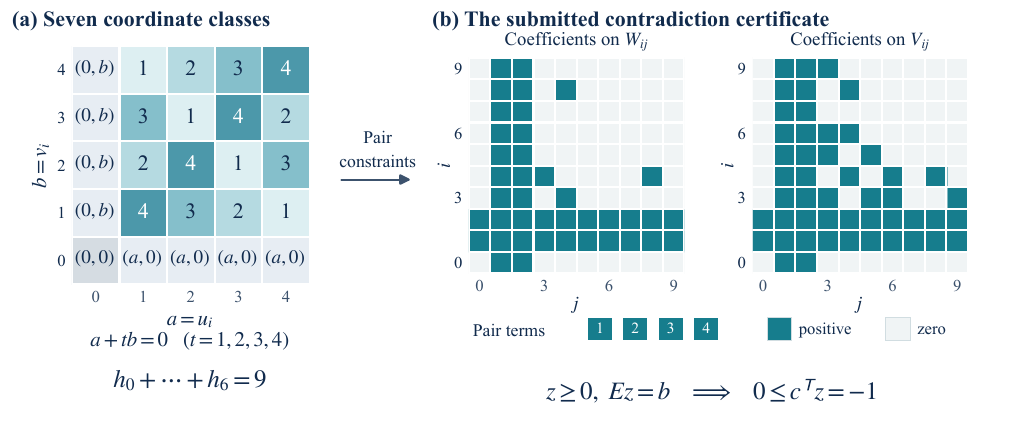}
\caption{\textbf{Astra's proof certificate for the nine-dimensional Alon--Jaeger--Tarsi case over $\mathbb{F}_5$.} (a) Coordinate pairs fall into seven classes; numbered cells give the unique nonzero $t$ with $a+tb=0$. Class counts sum to nine and yield pair constraints. (b) The actual certificate has nonnegative coefficients on two arrays of weight counts and four pair terms. Combined with the counting identities, it forces a nonnegative sum to equal $-1$ after positive rescaling, excluding a minimal counterexample. Colored cells mark positive coefficients; pale cells mark zeros.}
\label{fig:ajtcertificate}
\end{figure}
\endgroup

\arxivadded{\textbf{Verified special cases and bounds.}} While attempting the full conjecture, GPT-6 Astra proves the Alon--Jaeger--Tarsi conjecture over $\mathbb{F}_5$ for every $n\leq9$: every invertible $n\times n$ matrix $A$ admits a vector $x$ such that $x$ and $Ax$ have no zero coordinates. The $n=9$ case extends the dimensional ranges $n\leq7$ of \citet{yu1999permanent} and $n\leq8$ reported in a recent preprint \citep{zeng2026ajt}. The proof combines counting identities with an exact integer certificate to exclude a minimal counterexample (Figure~\ref{fig:ajtcertificate}). Independent Lean verification confirms the result using only standard axioms. Appendix~\ref{app:ajt} gives the proof structure, formal verification, and its relation to prior work. \arxivadded{Four further verified cases concern} a six-variable Fourier entropy--influence bound, an exact high-rank Hilbert-matrix rigidity formula, an influence bound for quadratic polynomials, and finite-instance online-matching bounds (Appendix~\ref{app:additionalopen}). \arxivadded{These five cases demonstrate that open-conjecture attempts can produce independently verifiable mathematical results. Analyzing their statements and proofs captures progress beyond the binary resolution score and identifies concrete results for further mathematical study.}

%% file: tables/main-results-arxiv.tex
\caption{\textbf{TCSExam-Full results on 100 tasks.} Tasks receive equal weight within each track. Solver uses the Kimi K3-relative metric; construction scores of 100.0\% match the task reference. Index is defined in Section~\ref{sec:scoring}. Accepted conjecture submissions adapt OpenAI\textquotesingle s published counterexamples and proofs (Section~\ref{sec:openresults}).}
\label{tab:mainresults}
\small\setlength{\tabcolsep}{2.5pt}\renewcommand{\arraystretch}{1.1}
\begin{tabular*}{\linewidth}{@{\extracolsep{\fill}}lrrrrr@{}}
\toprule
\tablehead[l]{\textbf{Agent}} & \tablehead[r]{\textbf{Theorem proving}\\Success (\%)\,$\uparrow$} & \tablehead[r]{\textbf{Solver}\\Relative (\%)\,$\uparrow$} & \tablehead[r]{\textbf{Constructions}\\Task score (\%)\,$\uparrow$} & \tablehead[r]{\textbf{Open conjectures}\\Resolved (\%)\,$\uparrow$} & \tablehead[r]{\textbf{Index}\\Points\,$\uparrow$} \\
\midrule
GPT-6 Astra & 70.0 & 64.7 & 104.9 & 0.0 & 71.9 \\
Kimi K3.1 Pro & 38.3 & 54.1 & 98.6 & 1.1 & 57.4 \\
Kimi K3 & 11.7 & 48.7 & 95.5 & 0.0 & 46.8 \\
\deepseeknewagent{} & 6.7 & 53.1 & 91.8 & 5.6 & 46.0 \\
GLM-5.3 & 5.0 & 49.1 & 97.7 & 0.0 & 45.5 \\
\qwenagent{} & 5.0 & 49.2 & 95.1 & 0.0 & 44.8 \\
Grok 4.6 & 5.0 & 43.8 & 86.2 & 0.0 & 40.5 \\
\deepseekagent{} & 0.0 & 48.3 & 86.6 & 0.0 & 40.4 \\
\bottomrule
\end{tabular*}

%% file: tables/lite-results.tex
\caption{\textbf{TCSExam-Lite results on 20 tasks, with five per track.} Tasks receive equal weight within each track. Solver uses the Kimi K3-relative metric; construction scores of 100.0\% match the task reference. Index is defined in Section~\ref{sec:scoring}.}
\label{tab:literesults}
\small\setlength{\tabcolsep}{2.5pt}\renewcommand{\arraystretch}{1.1}
\begin{tabular*}{\linewidth}{@{\extracolsep{\fill}}lrrrrr@{}}
\toprule
\tablehead[l]{\textbf{Agent}} & \tablehead[r]{\textbf{Theorem proving}\\Success (\%)\,$\uparrow$} & \tablehead[r]{\textbf{Solver}\\Relative (\%)\,$\uparrow$} & \tablehead[r]{\textbf{Constructions}\\Task score (\%)\,$\uparrow$} & \tablehead[r]{\textbf{Open conjectures}\\Resolved (\%)\,$\uparrow$} & \tablehead[r]{\textbf{Index}\\Points\,$\uparrow$} \\
\midrule
Opus 5.5 & 80.0 & 60.5 & 100.4 & 0.0 & 72.3 \\
GPT-6 Astra & 60.0 & 63.0 & 100.4 & 0.0 & 67.0 \\
Kimi K3.1 Pro & 40.0 & 50.0 & 88.3 & 0.0 & 53.5 \\
Kimi K3 & 33.3 & 48.9 & 79.8 & 0.0 & 48.6 \\
GLM-5.3 & 20.0 & 46.1 & 90.7 & 0.0 & 47.0 \\
\qwenagent{} & 20.0 & 51.0 & 80.3 & 0.0 & 45.4 \\
\deepseeknewagent{} & 6.7 & 54.3 & 80.2 & 0.0 & 42.4 \\
Grok 4.6 & 20.0 & 41.0 & 79.3 & 0.0 & 42.1 \\
\deepseekagent{} & 0.0 & 50.4 & 79.7 & 0.0 & 39.0 \\
\bottomrule
\end{tabular*}

%% file: sections/06-discussion.tex
\Needspace{24\baselineskip}
\section{Conclusion}
\label{sec:conclusion}

\arxivadded{TCSExam provides a unified, extensible framework for evaluating agents across the full range of TCS research, connecting theorem proving, solver engineering, and explicit constructions with open-conjecture settlement. Full comprises 100 expert-reviewed tasks with outcomes checked by the Lean kernel or executable checkers. Lite retains all four tracks in 20 tasks, reducing the number of task attempts by 80\% at the same per-attempt budget. Evaluations across nine agents reveal substantial differences in research capabilities, with strong Full--Lite rank agreement among the eight shared agents. Together, the suites support broad capability assessment and smaller-scale evaluation.}

\arxivadded{The results show complementary strengths in proof completion, solver performance, and construction quality. Submitted programs and mathematical objects improve selected task references. For both agents in the proof-guidance study, supplying intermediate lemmas and proof ideas increases the number of theorems solved. Open-conjecture attempts also yield independently verified special cases and bounds, including Astra's nine-dimensional Alon--Jaeger--Tarsi result over $\mathbb{F}_5$. These outputs help characterize the research work agents can undertake and identify directions for further progress. The accompanying task sources, Docker build recipes, and evaluation tools support reproducible use and extension. Researchers can add problems through the shared task interfaces and verification protocols, extending this foundation for evaluating agents and studying their contributions to TCS research.}

%% file: sections/ai-use.tex
\section*{AI Use Statement}

\arxivadded{AI coding assistants were used in task authoring, formalization, checker implementation, and iterative review (Section~\ref{sec:tasks}). AI assistants also supported data processing, result analysis, manuscript drafting and editing, figure code, and literature search. Evaluated model outputs are identified as experimental submissions and checked under the relevant task protocol. Result tables and plots are generated programmatically from experiment records. The authors take responsibility for the final content of this work.}

%% file: sections/appendix.tex
\clearpage
\input{sections/appendixA-tables}
\input{sections/appendix-a-examples}

\input{sections/appendix-b-scoring}
\input{sections/appendix-c-fidelity}
\input{sections/appendix-d-protocol}

\input{sections/appendix-e-results}
\input{sections/appendix-f-cases}
\input{sections/appendix-g-maintenance}

\input{sections/appendix-h-lite}

%% file: sections/appendixA-tables.tex
\section{Task Catalogue and Examples}
\label{app:tasklist}
The catalogue records the sources and research questions represented by the 100-task collection. Task counts refer to problem packages; solver instances and construction parameter settings are listed separately in the final evaluation manifest.
The references identify the mathematical sources; the evaluation protocol is specified in Appendix~\ref{app:protocol}.
\tracksubsection{theorem}{Theorem proving: 20 theorems}
{\scriptsize\emergencystretch=1.5em
\setlength{\tabcolsep}{3pt}
\begin{longtable}{@{}>{\raggedright\arraybackslash}p{0.19\linewidth}>{\raggedright\arraybackslash}p{0.52\linewidth}r>{\raggedright\arraybackslash}p{0.14\linewidth}@{}}
\toprule
Theorem & Source paper & Steps & Domain \\
\midrule\endfirsthead
\toprule
Theorem & Source paper & Steps & Domain \\
\midrule\endhead
\bottomrule\endfoot
\texttt{corners-\allowbreak{}quasipoly} & \emph{Quasipolynomial Bounds for the Corners Theorem} \citep{corners} & 8 & additive combinatorics \\
\texttt{dijkstra-\allowbreak{}universal} & \emph{Universal Optimality of Dijkstra via Beyond-Worst-Case Heaps} \citep{dijkstra} & 8 & algorithms \\
\texttt{discrepancy-\allowbreak{}beyond-\allowbreak{}banaszczyk} & \emph{Decoupling via Affine Spectral-Independence: Beck-Fiala and Komlós Bounds Beyond Banaszczyk} \citep{discrepancy} & 7 & discrepancy theory \\
\texttt{discrete-\allowbreak{}bidding-\allowbreak{}games} & \emph{Mean-Payoff and Energy Discrete-Bidding Games} \citep{bidding} & 8 & games \\
\texttt{folded-\allowbreak{}rs-\allowbreak{}gsb} & \emph{Explicit Folded Reed-Solomon and Multiplicity Codes Achieve Relaxed Generalized Singleton Bounds} \citep{foldedrs} & 8 & coding theory \\
\texttt{gg26-\allowbreak{}proximity} & \emph{Optimal Proximity Gaps for Subspace-Design Codes and (Random) Reed-Solomon Codes} \citep{ggproximity} & 6 & coding theory \\
\texttt{lcc-\allowbreak{}exp-\allowbreak{}lb} & \emph{An Exponential Lower Bound for Linear 3-Query Locally Correctable Codes} \citep{lcc} & 6 & coding theory \\
\texttt{monotonicity-\allowbreak{}test-\allowbreak{}lb} & \emph{Boolean Function Monotonicity Testing Requires (Almost) n\^{}\{1/2\} Queries} \citep{monotonicity} & 8 & property testing \\
\texttt{mts-\allowbreak{}diamond-\allowbreak{}lb} & \emph{The Randomized k-Server Conjecture is False!} \citep{kserver} & 8 & online algorithms \\
\texttt{multiflow-\allowbreak{}gap-\allowbreak{}planar} & \emph{Improved Lower Bounds on Multiflow-Multicut Gaps} \citep{multiflow} & 8 & approximation \\
\texttt{negations-\allowbreak{}small-\allowbreak{}depth} & \emph{Negations Are Powerful Even in Small Depth} \citep{negations} & 7 & circuit complexity \\
\texttt{pcp-\allowbreak{}alphabet-\allowbreak{}soundness} & \emph{Near Optimal Alphabet-Soundness Tradeoff PCPs} \citep{pcp} & 7 & PCP \\
\texttt{planar-\allowbreak{}is-\allowbreak{}reconfig-\allowbreak{}kernel} & \emph{A Linear Kernel for Independent Set Reconfiguration in Planar Graphs} \citep{planaris} & 7 & parameterized complexity \\
\texttt{posbody-\allowbreak{}chasing} & \emph{Chasing Positive Bodies} \citep{positivebodies} & 8 & online algorithms \\
\texttt{qma-\allowbreak{}qcma-\allowbreak{}oracle} & \emph{Separating QMA from QCMA with a Classical Oracle} \citep{qmaqcma} & 8 & quantum \\
\texttt{strategy-\allowbreak{}improvement-\allowbreak{}lb} & \emph{The Worst-Case Complexity of Symmetric Strategy Improvement} \citep{strategy} & 8 & games \\
\texttt{three-\allowbreak{}taxi} & \emph{Online 3-Taxi on General Metrics} \citep{threetaxi} & 8 & online algorithms \\
\texttt{unitary-\allowbreak{}synthesis-\allowbreak{}lb} & \emph{A One-Query Lower Bound for Unitary Synthesis and Breaking Quantum Cryptography} \citep{unitary} & 8 & quantum \\
\texttt{xor3-\allowbreak{}parallel-\allowbreak{}repetition} & \emph{Parallel Repetition for 3-Player XOR Games} \citep{xor} & 8 & PCP \\
\texttt{quantum-\allowbreak{}crypto-\allowbreak{}algorithmica} & \emph{Quantum cryptography in Algorithmica} \citep{algorithmica} & 8 & quantum computing \\
\end{longtable}
}
\tracksubsection{solver}{Solver engineering: 30 problems}
{\scriptsize\emergencystretch=1.5em
\setlength{\tabcolsep}{3pt}
\begin{longtable}{@{}>{\raggedright\arraybackslash}p{0.20\linewidth}>{\raggedright\arraybackslash}p{0.29\linewidth}>{\raggedright\arraybackslash}p{0.32\linewidth}>{\raggedright\arraybackslash}p{0.10\linewidth}@{}}
\toprule
Task & Problem & Source & Objective \\
\midrule\endfirsthead
\toprule
Task & Problem & Source & Objective \\
\midrule\endhead
\bottomrule\endfoot
\texttt{ahc058-\allowbreak{}atcoder} & Hierarchical production planning & AtCoder Heuristic Contest 058: Apple Incremental Game  & maximize \\
\texttt{bpp-\allowbreak{}bpplib} & One-dimensional bin packing & BPPLIB  & minimize \\
\texttt{ce-\allowbreak{}pace2021} & Cluster editing & PACE 2021 (heuristic track)  & minimize \\
\texttt{celar-\allowbreak{}rlfap} & Radio link frequency assignment (RLFAP) & CALMA/CELAR benchmarks (ZIB archive; MiniZinc Challenge)  & minimize \\
\texttt{christmas-\allowbreak{}tree-\allowbreak{}ahc041} & Christmas-tree forest partitioning & AtCoder Heuristic Contest 041: Christmas Tree Cutting  & maximize \\
\texttt{coloring-\allowbreak{}dimacs} & Graph coloring & DIMACS / COLOR02 benchmark (+ gc\_instances)  & minimize \\
\texttt{curriculum-\allowbreak{}itc2007} & Curriculum-based course timetabling & ITC 2007 Track 3  & minimize \\
\texttt{cvrp-\allowbreak{}cvrplib} & Capacitated vehicle routing & CVRPLIB  & minimize \\
\texttt{dfvs-\allowbreak{}pace2022} & Directed feedback vertex set & PACE 2022  & minimize \\
\texttt{ds-\allowbreak{}pace2025} & Dominating set & PACE 2025 (heuristic track)  & minimize \\
\texttt{glasscut-\allowbreak{}roadef2018} & 2D guillotine glass cutting with defects & ROADEF/EURO Challenge 2018  & minimize \\
\texttt{gp-\allowbreak{}dimacs10} & Balanced graph partitioning (k=4, 3\% balance cap) & DIMACS10 / Walshaw partition archive (34 graphs + 9 generated)  & minimize \\
\texttt{hs-\allowbreak{}pace2025} & Hitting set & PACE 2025 (heuristic track)  & minimize \\
\texttt{itc2021-\allowbreak{}sports-\allowbreak{}timetabling} & Sports timetabling & ITC 2021 (5th International Timetabling Competition)  & minimize \\
\texttt{jsplib-\allowbreak{}jssp} & Job-shop scheduling (makespan) & JSPLib (252 classic + 31 generated)  & minimize \\
\texttt{logic-\allowbreak{}syn-\allowbreak{}iwls2023} & Logic synthesis (minimum AIG) & IWLS 2023 Programming Contest \citep{iwls2023} & minimize \\
\texttt{maxclique-\allowbreak{}dimacs} & Maximum clique & DIMACS clique benchmark (classic 80-instance set)  & maximize \\
\texttt{maxcut-\allowbreak{}mqlib} & MaxCut & MQLib \citep{mqlib} & maximize \\
\texttt{maxsat-\allowbreak{}mse} & Weighted MaxSAT & MaxSAT Evaluation 2026 (anytime weighted set)  & minimize \\
\texttt{mr-\allowbreak{}roadef2012} & Machine reassignment & ROADEF/EURO Challenge 2012 \citep{roadef2012} & minimize \\
\texttt{nurse-\allowbreak{}inrc} & Nurse rostering & INRC-I (First International Nurse Rostering Competition, 2010)  & minimize \\
\texttt{oscm-\allowbreak{}pace2024} & One-sided crossing minimization & PACE 2024 (exact track)  & minimize \\
\texttt{qap-\allowbreak{}qaplib} & Quadratic assignment & QAPLIB \citep{qaplib} & minimize \\
\texttt{qbf-\allowbreak{}qbfeval} & QBF solving with Q-resolution certificates & QBFEVAL'20 (PCNF track)  & decision \\
\texttt{radiation-\allowbreak{}mznchallenge} & Radiotherapy planning & MiniZinc Challenge benchmarks  & minimize \\
\texttt{scp-\allowbreak{}orlib} & Set covering & OR-Library (full official 87-instance set) \citep{orlibrary} & minimize \\
\texttt{sygus-\allowbreak{}comp} & Program synthesis (SyGuS) & SyGuS-Comp 2017-2019 benchmarks (deduplicated)  & decision \\
\texttt{tsp-\allowbreak{}tsplib} & Traveling salesman (symmetric) & TSPLIB \citep{tsplib} & minimize \\
\texttt{tw-\allowbreak{}heuristic} & Treewidth decomposition & PACE 2017 (heuristic track)  & minimize \\
\texttt{vc-\allowbreak{}pace2019} & Vertex cover (exact) & PACE 2019 (exact track)  & decision \\
\end{longtable}
}
\tracksubsection{construction}{Constructions: 20 tasks}
{\scriptsize\emergencystretch=1.5em
\setlength{\tabcolsep}{3pt}
\begin{longtable}{@{}>{\raggedright\arraybackslash}p{0.18\linewidth}>{\raggedright\arraybackslash}p{0.24\linewidth}>{\raggedright\arraybackslash}p{0.37\linewidth}>{\raggedright\arraybackslash}p{0.12\linewidth}@{}}
\toprule
Task & Construction & Reference in source snapshot & Domain \\
\midrule\endfirsthead
\toprule
Task & Construction & Reference in source snapshot & Domain \\
\midrule\endhead
\bottomrule\endfoot
\texttt{ramsey-\allowbreak{}r55} & R(5,5) lower-bound construction & N = 42 (R(5,5) in [43, 46]) \citep{exoo1989r55} & Ramsey theory \\
\texttt{ramsey-\allowbreak{}r46} & R(4,6) lower-bound construction & N = 35 (R(4,6) in [36, 40]) \citep{exoo2012r46} & Ramsey theory \\
\texttt{ramsey-\allowbreak{}r3-\allowbreak{}10} & R(3,10) lower-bound construction & N = 39 (R(3,10) in [40, 41]) \citep{exoo1989r310} & Ramsey theory \\
\texttt{vdw-\allowbreak{}w27} & van der Waerden number W(2,7) lower bound & n = 3703 \citep{rabung2012} & Ramsey theory \\
\texttt{capset-\allowbreak{}d7} & Maximum cap set in dimension 7 & 236 points (upper bound 288; earlier bound 291) \citep{calderbank1994,thackeray2022,versluis2017} & additive combinatorics \\
\texttt{covering-\allowbreak{}code-\allowbreak{}k10} & Binary covering code K(10,1) (minimize codewords) & 120 codewords (K(10,1) in [107, 120]) \citep{ostergard1991} & coding theory \\
\texttt{binary-\allowbreak{}code-\allowbreak{}a22-\allowbreak{}10} & A(22,10) maximum binary code & 64 codewords (A(22,10) in [64, 84]) \citep{ostergard2005} & coding theory \\
\texttt{binary-\allowbreak{}code-\allowbreak{}a24-\allowbreak{}10} & A(24,10) maximum binary code & 136 codewords (A(24,10) in [136, 268]) \citep{laaksonen2016} & coding theory \\
\texttt{quantum-\allowbreak{}code-\allowbreak{}25-\allowbreak{}7} & Quantum stabilizer code [[25,7]] maximum minimum distance & d = 5 (upper bound 7) \citep{codetables25} & quantum codes \\
\texttt{mols-\allowbreak{}n10} & Mutually orthogonal Latin squares, per-n records (10 cells) & n=\{10,12,14,15,18,20,21,22,24,26\} -\ensuremath{>} k=\{2,5,4,4,5,4,5,3,7,4\} \citep{miller2024mols} & combinatorial designs \\
\texttt{cage-\allowbreak{}girth} & (k,g)-cage minimum order (cells (3,13)/(4,10)/(3,16), minimize N) & 272 / 320 / 936 (lower bounds 202/243/512) \citep{biggs1989,exoo2013cagesurvey,exoo2026cage,houseofgraphscages} & extremal graph theory \\
\texttt{sorting-\allowbreak{}network-\allowbreak{}size} & Sorting-network minimum comparators (8 cells n=13..20, minimize) & 45/51/56/60/71/77/85/91 \citep{dobbelaeresorting} & algorithms \\
\texttt{ex-\allowbreak{}c3c4-\allowbreak{}maxedges} & Erdős girth-5 extremal problem ex(n,\{C3,C4\}) (cells n=80/124/175/198) & 321/629/1012/1166 edges \citep{girthbounds} & extremal graph theory \\
\texttt{kissing-\allowbreak{}d9} & 9-dimensional kissing number lower bound & 306 points (d=9 in [306, 363]) \citep{leech1971} & sphere packing \\
\texttt{kissing-\allowbreak{}d13} & 13-dimensional kissing number lower bound & 1154 points (d=13 in [1154, 2064]) \citep{zinoviev1999contact} & sphere packing \\
\texttt{circles-\allowbreak{}in-\allowbreak{}square-\allowbreak{}n100} & 100 non-overlapping equal circles in a unit square (maximize radius) & r = 0.051401071774 \citep{packomaniacsq} & packing geometry \\
\texttt{spherical-\allowbreak{}code-\allowbreak{}d3-\allowbreak{}n50} & Tammes problem, 50 points on the sphere (minimize max inner product) & max inner product 0.868173209127 \citep{szekely1974tammes} & spherical codes \\
\texttt{tsp-\allowbreak{}gap} & Metric TSP subtour-LP integrality gap (n=15..20) & Per-n values: 15: 11/9, 16: 27/22, 17: 53/43, 18: 26/21, 19: 77/62, 20: 76/61 \citep{benoit2008} & approximation \\
\texttt{graph-\allowbreak{}tsp-\allowbreak{}gap} & graph-TSP subtour-LP integrality gap (n=11..20) & Per-n values, window max 26/21 \citep{desjardins2017} & approximation \\
\texttt{rectilinear-\allowbreak{}tsp-\allowbreak{}gap} & Rectilinear (L1) TSP subtour-LP integrality gap (n=11..16) & Zhong family per-n: 11: 17/15, 12: 8/7, 13: 23/20, 14: 22/19, 15: 7/6, 16: 142/121 \citep{zhong2025} & approximation \\
\end{longtable}
}
\tracksubsection{open}{Open conjectures: 30 conjectures}
{\scriptsize\emergencystretch=1.5em
\setlength{\tabcolsep}{3pt}
\begin{longtable}{@{}>{\raggedright\arraybackslash}p{0.20\linewidth}>{\raggedright\arraybackslash}p{0.35\linewidth}>{\raggedright\arraybackslash}p{0.24\linewidth}>{\raggedright\arraybackslash}p{0.12\linewidth}@{}}
\toprule
Task & Conjecture & Source & Domain \\
\midrule\endfirsthead
\toprule
Task & Conjecture & Source & Domain \\
\midrule\endhead
\bottomrule\endfoot
\texttt{aaronson-\allowbreak{}ambainis} & Aaronson-Ambainis conjecture & \citep{aaronson2014} & quantum computing \\
\texttt{alon-\allowbreak{}jaeger-\allowbreak{}tarsi} & Alon-Jaeger-Tarsi conjecture (prime-field form) & \citep{nagypach2025ajt} & combinatorics \\
\texttt{bin-\allowbreak{}packing-\allowbreak{}mirup} & MIRUP conjecture (modified integer round-up property, 1-D bin packing) & \citep{scheithauer1995} & integer programming \\
\texttt{caccetta-\allowbreak{}haggkvist} & Caccetta-Häggkvist conjecture & \citep{caccetta1978} & graph theory \\
\texttt{d-\allowbreak{}vs-\allowbreak{}r-\allowbreak{}total-\allowbreak{}functions} & D(f) \ensuremath{\leq} C·R(f)\^{}2 for all total Boolean functions & \citep{nisan1991} & query complexity \\
\texttt{evasiveness-\allowbreak{}graph-\allowbreak{}properties} & Evasiveness conjecture (Karp's conjecture) & \citep{rosenberg1973} & query complexity \\
\texttt{explicit-\allowbreak{}ramsey-\allowbreak{}graph-\allowbreak{}log} & Paley-graph Ramsey conjecture (polylogarithmic form) & \citep{graham1990} & pseudo\-randomness \\
\texttt{explicit-\allowbreak{}rigid-\allowbreak{}matrix} & Valiant's matrix-rigidity conjecture (Hilbert-matrix family) & \citep{valiant1977} & algebraic complexity \\
\texttt{fourier-\allowbreak{}entropy-\allowbreak{}influence} & Fourier Entropy-Influence conjecture (FEI) & \citep{friedgut1996} & Boolean analysis \\
\texttt{frege-\allowbreak{}not-\allowbreak{}pbounded} & Frege systems are not polynomially bounded & \citep{cook1979} & proof complexity \\
\texttt{frege-\allowbreak{}vs-\allowbreak{}extended-\allowbreak{}frege} & Frege does not polynomially simulate Extended Frege & \citep{cook1979} & proof complexity \\
\texttt{krw-\allowbreak{}composition-\allowbreak{}conjecture} & KRW composition conjecture & \citep{karchmer1995} & communication complexity \\
\texttt{log-\allowbreak{}rank-\allowbreak{}conjecture} & Log-rank conjecture & \citep{lovasz1988} & communication complexity \\
\texttt{lovasz-\allowbreak{}hamiltonicity} & Lovász's Hamiltonicity conjecture (connected vertex-transitive graphs have a Hamiltonian path) & \citep{babai1979} & graph theory \\
\texttt{mansour-\allowbreak{}conjecture} & Mansour's conjecture (DNF Fourier concentration) & \citep{mansour1995} & learning theory \\
\texttt{mds-\allowbreak{}conjecture-\allowbreak{}nonprime} & MDS main conjecture over non-prime finite fields & \citep{ball2012} & coding theory \\
\texttt{nexp-\allowbreak{}vs-\allowbreak{}tc0} & NEXP not in TC0 & \citep{chentell2019} & circuit complexity \\
\texttt{prg-\allowbreak{}constant-\allowbreak{}width-\allowbreak{}robp} & o(log\^{}2 n)-seed PRG for width-4 read-once branching programs & \citep{nisan1992,forbes2018} & pseudo\-randomness \\
\texttt{np-\allowbreak{}vs-\allowbreak{}acc} & NP not in ACC & \citep{williams2014} & circuit complexity \\
\texttt{omega-\allowbreak{}2} & Matrix multiplication exponent omega = 2 & \citep{strassen1969} & algebraic complexity \\
\texttt{online-\allowbreak{}bin-\allowbreak{}packing-\allowbreak{}ratio} & Optimal asymptotic competitive ratio of online bin packing (benchmark threshold R* \ensuremath{\leq} 1.56) & \citep{balogh2018upper,balogh2021} & online algorithms \\
\texttt{online-\allowbreak{}matching-\allowbreak{}random-\allowbreak{}order} & Competitive ratio of online bipartite matching under random arrival order & \citep{karp1990,goel2008} & online algorithms \\
\texttt{permanent-\allowbreak{}vs-\allowbreak{}determinant} & Permanent cannot be written as a polynomial-size determinant & \citep{valiant1979algebra} & algebraic complexity \\
\texttt{pfr-\allowbreak{}integers} & Polynomial Freiman-Ruzsa conjecture over the integers (Marton's conjecture) & \citep{lovett2017} & additive combinatorics \\
\texttt{planar-\allowbreak{}l1-\allowbreak{}gnrs} & GNRS conjecture (planar-graph l1 embedding with constant distortion) & \citep{gupta2004} & metric embeddings \\
\texttt{polar-\allowbreak{}scaling-\allowbreak{}exponent} & Polar-code scaling-exponent conjecture (mu = 3.627) & \citep{mondelli2016} & coding theory \\
\texttt{ryser-\allowbreak{}conjecture} & Ryser's conjecture (tau(H) \ensuremath{\leq} (r-1)*nu(H) for r-partite r-uniform hypergraphs) & \citep{debiasio2021} & combinatorics \\
\texttt{total-\allowbreak{}coloring-\allowbreak{}conjecture} & Total Coloring Conjecture (chi''(G) \ensuremath{\leq} Delta(G)+2) & \citep{vizing1968} & graph theory \\
\texttt{tuza-\allowbreak{}conjecture} & Tuza's conjecture (triangle edge cover \ensuremath{\leq} 2x triangle edge packing) & \citep{tuza1990} & graph theory \\
\texttt{xor-\allowbreak{}log-\allowbreak{}rank-\allowbreak{}conjecture} & XOR log-rank conjecture & \citep{montanaro2009,tsang2013} & communication complexity \\
\end{longtable}
}

%% file: sections/appendix-a-examples.tex
\subsection{Reading the four task interfaces}
\label{app:examples}

\arxivadded{The following examples explain what an agent submits and how it is evaluated in each track. The versioned task packages provide the full statements, input formats, and checker configurations.}

\trackheading{theorem}{Theorem proving: corners-quasipoly.}
\label{app:ex-corners}
The task concerns the quantitative bound for corner-free sets of \citet{corners} and provides formal definitions and a fixed target theorem. Bare attempts must prove the target from the supplied mathematical context. Ladder and guided attempts additionally receive intermediate statements; guided attempts receive a proof idea. A local checkpoint proof is evaluated under the other checkpoint statements, whereas the complete-proof check excludes unproved checkpoint assumptions. The fixed target declaration and supplied assumptions determine the statement whose complete proof is scored.

\trackheading{solver}{Solver engineering: maxcut-mqlib.}
\label{app:ex-maxcut}
A submitted program receives a weighted graph from the MQLib collection \citep{mqlib} and returns a vertex partition. The checker validates the partition and recomputes the total weight of edges crossing the cut. Final judging applies the same program across the designated instance family under the task's resource limits. \arxivadded{The evaluation measures the program\textquotesingle s performance across the instance family.} The final manifest specifies public and held-out instance counts and the limits used for each set.

\trackheading{construction}{Constructions: capset-d7.}
\label{app:ex-capset}
A cap set in $(\mathbb{Z}/3\mathbb{Z})^7$ contains no three distinct points whose sum is zero. The submission supplies a set of points and the required certificate. Verification checks membership, distinctness, and the cap-set property, then recomputes cardinality. \arxivadded{The task scores the verified cardinality using its construction-scoring function.}

\trackheading{open}{Open conjectures: log-rank-conjecture.}
\label{app:ex-logrank}
The task asks for a proof or refutation of a formal relationship between deterministic communication complexity and the rank of the communication matrix. Two slots hold proofs of $P$ and $\neg P$. Each is checked without access to the unproved opposite slot. \arxivadded{Complete proof or refutation earns the resolution score; verified special cases are retained as auxiliary mathematical results.}

%% file: sections/appendix-b-scoring.tex
\section{Scoring and Verification}
\label{app:scoring}

The task version fixes the submission interface, legal operations, checker, reference values, and resource limits. \arxivadded{Official scores require both mathematical validity and compliance with the submission rules. Separately verified mathematical outputs are reported with their originating task and verification conditions.}

\tracksubsection{theorem}{Theorem proving: conditional coverage and complete proofs}
\label{app:scoring-lean}

For a task with $m$ checkpoints, let $I_j$ indicate that checkpoint $j$ passes its isolated check. All checkpoints within a task receive equal credit, so conditional checkpoint coverage is
\begin{equation}
C=\frac{1}{m}\sum_{j=1}^{m} I_j.
\end{equation}
When checking checkpoint $j$, the verifier supplies the other checkpoints' reference statements as assumptions. Passing a checkpoint establishes that obligation under the supplied assumptions. \arxivadded{This rule applies equally to an intermediate lemma and to the main-theorem checkpoint, so $C$ measures conditional proof coverage.}

Complete-proof validation uses a separate check. The \texttt{assembly\_full} path verifies the assembled proof; the \texttt{main\_direct} path checks a direct proof of the main theorem. Acceptance requires the permitted fixed context and excludes unproved checkpoint assumptions and \texttt{sorryAx}. For guided and ladder tasks, an accepted complete proof receives full credit: $S=\max(F,C)$, where $F$ is the complete-proof indicator. Complete-proof outcomes use the separately recorded indicator $F$; the guidance analysis reports both complete-proof success and partial-credit task scores. Bare tasks have one complete target and use the binary complete-proof result.

Formal tasks use Lean 4 \citep{lean4} and, where required, the mathlib library \citep{mathlib}; task manifests pin their versions. \arxivadded{Acceptance combines successful compilation with checks of fixed statements, writable regions, and permitted proof dependencies.} Reports retain the check mode, declaration name, permitted assumptions, and the dependency output needed to interpret acceptance.

\tracksubsection{solver}{Solver engineering: instance performance and reference-model comparison}
\label{app:scoring-solver}

For an optimization instance with attained value $A$, naive-baseline value $B$, and reference value $K$, the ordinary case with a feasible baseline and a strictly better reference uses
\begin{equation}
s(A;B,K)=\frac{A-B}{K-B}.
\end{equation}
The expression applies to both optimization directions when $K$ is better than $B$. It assigns 0 at $B$ and 1 at $K$, and is unbounded in either direction. Checkers validate feasibility and recompute $A$; invalid solutions, solver crashes, and budget overruns receive the task's failure score, which is zero for the reported optimization tasks. A valid solution worse than $B$ can therefore score below an invalid submission; failure rates and raw objectives distinguish these outcomes. Decision tasks use their declared solve-rate or certificate-based rule.

Instances with equal baseline and reference values, a reference worse than the baseline, or no feasible baseline do not enter the ratio-score mean. Their raw outcomes remain in the evaluation records and may enter the Kimi K3-relative comparison. An absent reference value is distinct from an absent model outcome. Instances are not discarded because a submitted solver fails, and missing judging records are tracked separately from failed solutions.

The ratio score measures improvement relative to the chosen baseline and reference. Its sensitivity is $|\Delta s|=|\Delta A|/|K-B|$: a wide baseline--reference difference can make distinct outcomes near $K$ look numerically similar, while a narrow difference can yield very large contributions. The Kimi K3-relative score provides a second comparison using the raw outcomes of the three Kimi K3 attempts on each instance. Its scale depends on those observed outcomes and the floor below, rather than on $K-B$; the bounded transform limits the contribution of any one instance to the relative mean.

For optimization, set $a=A$ for maximization and $a=-A$ for minimization; for decision tasks, $a$ is the binary solve indicator. Let $\mu_i$ and $\sigma_i$ be the reference mean and sample standard deviation on instance $i$, with variance denominator $n_i-1$. The comparison uses
\begin{equation}
\widetilde\sigma_i=\max(\sigma_i,\,0.02|\mu_i|,\,10^{-9}),\qquad
z_i=\frac{a_i-\mu_i}{\widetilde\sigma_i},\qquad q_i=\Phi(z_i),
\end{equation}
\arxivadded{Here $\Phi$ is the standard Gaussian cumulative distribution function, and the reported percentage score is $100\Phi(z_i)$. A valid outcome equal to the Kimi K3 reference mean receives 50.0\%; better outcomes score higher and worse outcomes score lower. The transform is most sensitive near the reference mean and compresses differences far from it. The scale floor prevents small reference variation from magnifying tiny performance differences. Because the transform is nonlinear, averaging Kimi K3's transformed outcomes need not give 50.0\%. Raw objective values and baseline-relative ratio scores accompany this comparison.}

For optimization instances, failed Kimi K3 attempts supply no numerical reference value. The comparison includes only instances with at least two valid reference values; the same eligible instance set is used for all models. Decision failures supply a reference outcome of zero. A failed candidate solution receives relative score zero directly, rather than being interpreted as a zero-cost solution. Missing judging records are reported separately and are not filled with numerical zeros. We transform each instance outcome before averaging over instances, then average attempts within tasks and give the 30 tasks equal weight. Reference coverage, feasibility outcomes, and the unbounded ratio scores accompany this comparison.

Final solver execution follows the task's declared budget. Per-instance tasks limit each instance separately. Time-pool tasks permit allocation of a single total budget across the family; evaluation must preserve that allocation opportunity. Certificate requirements are also task-specific: for example, SAT unsatisfiability claims require the designated proof certificate and certificate checker. Final results use the final submitted program on the designated evaluation set.

\tracksubsection{construction}{Constructions: mathematical parameters and shaped scores}
\label{app:scoring-construction}

The checker verifies the submitted object and recomputes its mathematical parameter $X$. Let $B_t$ and $R_t$ denote the task's trivial baseline and documented reference. The task's calibrated attainment $u_t(X)$ assigns 0 at the baseline and 1 at the reference. For an ordinary linear calibration,
\begin{equation}
u_t(X)=\max\!\left(0,\frac{X-B_t}{R_t-B_t}\right),\qquad S_t(X)=u_t(X)^{p_t}.
\end{equation}
The same ratio accommodates minimization when $R_t<B_t$. Task-specific nonlinear calibrations replace $u_t$ where required. The scoring design uses $p_t=3$ for 14 tasks and $p_t=1$ for six, with no upper cap at the reference. Multi-parameter tasks apply their declared aggregation over all required settings, including missing submissions. \arxivadded{For accepted submissions, a score above 1 denotes improvement over the task reference under its validity and tolerance checks. The achieved parameter $X$ reports the mathematical improvement in its original units.}

\tracksubsection{open}{Open conjectures: full proof or refutation}
\label{app:scoring-open}

A task supplies a formal proposition $P$ and slots for $P$ and $\neg P$. A valid proof in either direction earns 1; otherwise the complete-resolution score is 0. Each main slot is checked in isolation after removing its contradictory twin, without treating either unproved main statement as an assumption. Auxiliary statements may support the submitted proof only within the permitted verification context. Special cases, reductions, and counterexamples to related stronger statements are reported with their precise scope and do not change the binary target score.

%% file: sections/appendix-c-fidelity.tex
\section{Task Development and Fidelity Audits}
\label{app:fidelity}

\subsection{From a research source to a task}

Agent-assisted source reviews identify the intended research claim, its quantifiers and parameter range, and the result or object accepted by the task. For formal statements, component reconciliation maps each source definition and assumption to the corresponding Lean declaration. Small examples and boundary cases check that definitions behave as intended. Re-translation provides an additional comparison between the source and its formalization. Some re-translation passes were performed by the authoring agent, while other reviews used a fresh context. Shared definitions and statements are checked across the three theorem-disclosure levels.

Checker review exercises both valid and invalid submissions: empty outputs, malformed objects, incorrect objectives, incomplete certificates, and attempts to change fixed statements. \arxivadded{Reference outputs are used where available. For open conjectures, development tests check the proof-or-refutation interface and its rejection of invalid submissions.}

\begin{table}[ht]
\centering\small
\caption{Task-validation checks and the evidence used to assess each failure mode.}
\begin{tabularx}{\linewidth}{@{}p{0.23\linewidth}XX@{}}
\toprule
Check & Evidence & Failure addressed \\
\midrule
Source correspondence & Source location, formal statement, reconciliation & Omitted assumptions or changed quantifiers \\
Definition behavior & Hand-checkable examples and executed checks & Degenerate or incorrectly encoded objects \\
Submission integrity & Fixed-region and proof-dependency checks & Changing the task or assuming the result \\
Checker behavior & Valid, invalid, and empty test outputs & Accepting an invalid answer or rejecting the documented interface \\
Version consistency & Versioned statements, environments, and scoring rules & Mixing incompatible statements or score definitions \\
\bottomrule
\end{tabularx}
\end{table}

\subsection{Review triggered by unexpected outcomes}

A full proof or a value beyond the reference triggers agent-assisted re-examination of the rollout, submitted artifact, and task statement. The review checks adherence to the task rules, the accepted proof's assumptions or the object's verified properties, and the scope of the claimed result. For numerical improvements, source comparison checks the prior reference, problem parameters, instance set, and resource conditions. Mathematical validity and literature novelty are documented separately. If a statement or checker changes, the corrected task receives a new version; results on different versions retain their original scope.

\paragraph{Potential inconsistencies in research sources.}
Development checks sometimes identified apparent missing boundary conditions, discrepancies between definitions and their use in proofs, or potential gaps in arguments. We treated these as concerns requiring source reconciliation, including checks for mistakes in our own formalization. Candidate tasks were excluded when such concerns could not be resolved. \arxivadded{This source-reconciliation process informed task selection and formalization.}

\subsection{Curation and evaluation population}

The theorem collection was screened through pilot evaluations, with over 40 theorem problems retired across development rounds. The evaluation records retain the screening model and configuration, candidate and included task lists, and reasons for retirement. \arxivadded{Because pilot performance informed task selection, the collection is conditioned on the screening agent\textquotesingle s outcomes, which can affect its comparison with other agents.} Task additions and retirements are recorded without changing historical evaluation manifests.

%% file: sections/appendix-d-protocol.tex
\section{Experimental Protocol and Reproduction}
\label{app:protocol}

\arxivadded{The project repository (\url{\projectrepository}) provides task statements, Full and Lite suite definitions, Dockerfiles, data recipes, judging code, and evaluation commands. Users build environments on a Linux Docker host and configure their model provider and harness. The 100 logical tasks are packaged as 70 Lean tasks and 60 Solver variants, with Online and Offline versions for each Solver problem.}

\Needspace{5\baselineskip}
\subsection{Agents, versions, and budgets}

\arxivadded{The evaluation includes eight agents on Full and nine on Lite.} Each attempt has a development budget of up to 12 hours. \arxivadded{General Internet access is disabled} for theorem and solver tasks and enabled for construction and conjecture tasks. Agents use the supplied tools and libraries within the task rules. Solver judging has separate task-specific time and hardware limits.

\begin{table}[ht]
\centering\small
\begin{arxivaddition}
\caption{Model and harness configurations. Counts are attempts per task in each suite.}
\label{tab:systems}
\begin{tabular*}{\linewidth}{@{\extracolsep{\fill}}lllrr@{}}
\toprule
Model & Harness & Effort & Full & Lite \\
\midrule
GPT-6 Astra & Codex 0.154.0 & max & 1 & 1 \\
Kimi K3.1 Pro & Kimi Code 0.43.0 & max & 3 & 3 \\
Kimi K3 & Kimi Code 0.43.0 & max & 3 & 3 \\
GLM-5.3 & Claude Code 2.1.257 & max & 1 & 1 \\
\qwenagent{} & Claude Code 2.1.257 & max & 1 & 1 \\
Grok 4.6 & Grok Build 1.0.41 & xhigh & 1 & 1 \\
\deepseekagent{} & Claude Code 2.1.257 & max & 3 & 3 \\
\deepseeknewagent{} & Claude Code 2.1.257 & max & 3 & 3 \\
Opus 5.5 & Claude Code 2.1.283 & max & -- & 1 \\
\bottomrule
\end{tabular*}
\end{arxivaddition}
\end{table}

Grok 4.6 uses the Grok Build harness with thinking enabled and reasoning effort set to xhigh. Agent development budgets and final solver execution budgets are recorded separately. Each result is associated with its task definition, submitted output, verification outcome, and resource use.

\input{sections/appendix-d-solver-tasks}

\subsection{Repetitions and aggregation}

For agent $m$, task $t$, and reported attempt $r$, let $s_{mtr}$ be the score and $R_m$ the number of attempts per task. \arxivadded{Across all four tracks, $R_m=3$ for Kimi K3, Kimi K3.1 Pro, and both DeepSeek versions, and $R_m=1$ for the other agents.} Within a track with task set $\mathcal T$, we report
\begin{equation}
\bar{s}_{m}=\frac{1}{|\mathcal T|}\sum_{t\in\mathcal T}
\left(\frac{1}{R_m}\sum_{r=1}^{R_m}s_{mtr}\right).
\end{equation}
Task-specific aggregation over instances or parameter settings precedes this calculation. Every task has equal weight. For binary theorem outcomes, the formula equals the number of complete proofs \arxivadded{divided by $20R_m$ on Full or $5R_m$ on Lite.} Case studies examine individual submissions; the main table retains average performance across attempts.

\trackparagraph{solver}{Solver evaluation.}
\label{app:solverprotocol}
\arxivadded{Full Solver evaluation uses the repetition counts in Table~\ref{tab:systems}: 480 submitted programs across eight agents. The 30 problems contain 1,870 final instances. The original six-agent cohort contributes ten attempts per problem in total: 300 programs and $10\times1{,}870=18{,}700$ instance outcomes. The detailed raw-outcome diagnostics in Appendix~\ref{app:solverdiagnostics} use this cohort. Table~\ref{tab:solverrelative} includes all eight agents. Lite has five Solver tasks. Both suites use public self-tests during development and a single final evaluation per attempt; Section~\ref{sec:setup} describes their service-based and local self-test execution.} Each submitted program is evaluated under the task's time and hardware limits. Independent checks of the evaluation records verify instance membership, solution validity, objective values, and aggregation.

\trackparagraph{theorem}{Proof-guidance evaluation.}
\label{app:guidancepopulation}
Kimi K3 and \deepseekagent{} each contribute three attempts in each of the three settings on 20 theorems, giving 60 bare, 60 ladder, and 60 guided attempts per agent. Guided and ladder share their task definitions and main targets, with guided additionally supplying proof ideas.

An accepted assembled proof or direct main proof counts as a complete-proof success. The per-attempt rate divides successful attempts by 60; theorem coverage counts targets with at least one success among the three attempts. All submitted proofs are checked using the task's fixed definitions and permitted axioms. Complete-proof outcomes and partial-credit task scores are reported separately.

Cross-setting case analyses also check that the main theorem and its definitions agree. The bare setting can differ in its supporting definitions, prompt wording, and verifier; Appendix~\ref{app:checkpoints} details these differences for the selected cases.

\trackparagraph{construction}{Construction evaluation.}
\label{app:constructionprotocol}
\arxivadded{The original six-agent cohort receives construction prompts} that describe published reference bounds but omit the numerical score transformation. The 12-hour development budget applies to producing the submitted object and certificate. \arxivadded{In this original cohort, post-submission verification evaluates mathematical validity with extended compilation where needed.} Acceptance requires the unchanged submitted artifact to satisfy the task statement, fixed-context checks, and permitted proof dependencies. Scores use these verified outcomes, and reference improvements are distinguished from matching the reference within the stated numerical tolerance.

\trackparagraph{open}{Open-conjecture evaluation.}
\arxivadded{The eight Full agents attempt all 30 conjectures; the nine Lite agents have results on the selected five, with the repetitions in Table~\ref{tab:systems}.} Success requires a complete proof or refutation of the stated conjecture; accepted refutations occur on the Ryser and MIRUP tasks. \arxivadded{The source note below Table~\ref{tab:full-open} identifies the published proofs used in these submissions.} Verified auxiliary results and special cases are analyzed separately as mathematical outputs. \arxivadded{Analysis focuses on the verified statements, proof methods, and relation to the target conjecture.}

\subsection{Interpreting variation}
\label{app:variance}

\arxivadded{The theorem and construction plots show variation across tasks; repeated attempts additionally show within-task variation for agents evaluated more than once. For agents evaluated once, task-level distributions describe differences across problems. Figure~\ref{fig:proofdiagnostics} highlights large task-mean differences and individual Kimi K3 outcomes across proof settings, while the aggregate analysis includes all 20 paired theorems. Attempts are independent across settings, with unmatched random seeds. The comparison evaluates complete agents: the model and harness listed in Table~\ref{tab:systems}, together with the tools and network access permitted for each track.}

\Needspace{6\baselineskip}
\subsection{Recorded token usage and duration}
\label{app:resources}

\arxivadded{Table~\ref{tab:resourceuse} reports accumulated input and output tokens and recorded duration per task attempt for all eight Full agents.} Each mean uses only attempts with an observed value for that field, with coverage reported alongside it. \arxivadded{All scheduled attempts remain in the performance evaluation, including those with missing telemetry.}

Usage measurements primarily use the recorded API usage totals. Where API totals are unavailable, counts are reconstructed from request traces or estimated from recorded text. \arxivadded{Accumulated input totals include context counted on successive requests. The totals follow each provider\textquotesingle s tokenization and accounting conventions, with cached input and reasoning tokens separated where recorded. Table~\ref{tab:resourceuse} summarizes this recorded workload by agent and track.}

For the original six agents, Solver durations are estimated from execution logs and output timestamps, excluding final judging delays where these can be identified; budget-exhausting attempts are recorded at 12 hours. \arxivadded{Kimi K3.1 Pro and DeepSeek 1017 durations use recorded agent-execution start and end times for all four tracks.}

Attempts can finish before the 12-hour development budget is exhausted. Recorded durations can also include termination and reporting overhead beyond the nominal limit.

\input{tables/resource-usage}

%% file: sections/appendix-d-solver-tasks.tex
\tracksubsection{solver}{Solver task construction and execution}
\label{app:solvertasks}

\paragraph{Sources and evaluation sets.}
The solver collection comprises 19 competition-derived tasks and 11 tasks based on instance libraries. Source type and execution budget are separate choices: either type can use per-instance limits or a shared time pool. Each task specifies a public development set, a final evaluation set, an executable submission interface, and a checker. Competition-derived tasks can retain the source's development/test distinction or define a new split of released instances. Library-derived tasks define their own splits; these are fixed in the task version rather than selected from model outcomes. Table~\ref{tab:solverprotocols} records the evaluated configurations.

Public instances and self-test results are available during program development. Hidden instance files and transformation parameters remain on the judging side, outside the \arxivadded{agent's development environment}. \arxivadded{Public development and final evaluation use disjoint sets of source instances.}

\paragraph{Adapting an instance library.}
The quadratic-assignment task illustrates the conversion of a public library into a solver evaluation. From 136 QAPLIB instances \citep{qaplib}, the versioned split assigns 41 to hidden evaluation, 75 to public development, and 20 to neither set. Hidden selection is stratified by instance family and size; withholding the third group prevents the public complement from identifying the evaluation set exactly. Hidden instances independently permute facility and location indices. If $\pi$ relabels facilities and $\sigma$ relabels locations, a feasible assignment $p$ becomes $p'=\sigma\circ p\circ\pi^{-1}$. Applying the same relabelings to the flow and distance matrices preserves every assignment's cost and the optimal value. The checker validates the submitted permutation and recomputes its cost on the transformed instance.

Other transformations follow the problem's structure, such as vertex relabeling for graph problems and item reordering for bin packing. \arxivadded{These changes obstruct direct reuse of answers in the original indexing. Training exposure to the public source instances is unknown. The submission rules require computation during judging and prohibit replay of stored final answers; hidden access controls, transformations, and rollout audits support compliance at complementary stages.}

\paragraph{Development and execution budgets.}
The 12-hour budget covers model interaction and program development. Afterward, the frozen program is rebuilt and evaluated under the task's separate execution limits. Twenty-four tasks impose limits independently on each instance. Six tasks instead share one total time pool: bin packing, graph partitioning, sports timetabling, logic synthesis, quadratic assignment, and set cover. The judge processes the evaluation family while tracking the remaining pool, which is supplied to the program on each invocation. Time saved on one instance is available to later instances; these tasks are evaluated as one group to preserve this allocation rule. For example, quadratic assignment has 1,230 seconds across its 41 final instances, with no separate 30-second cutoff per instance. Glass cutting uses two runs per instance, at 180 and 3,600 seconds, with weights 0.1 and 0.9. \arxivadded{Table~\ref{tab:solverprotocols} specifies each task's final execution time budget.}

\paragraph{Parallel evaluation.}
For per-instance evaluation, we use up to ten isolated sandboxes concurrently, using ten whenever the number of instances and available resources permit. Each instance retains the specified time, CPU, and memory limits. Shared-pool tasks remain in one evaluation group to preserve cross-instance time allocation.

\input{tables/solver-protocols}

%% file: tables/solver-protocols.tex
\begin{table}[htbp]
\centering
\caption{\arxivadded{Solver instance sets and final execution budgets. Budgets are in seconds: \emph{Each} is a per-instance limit; \emph{Pool} is the total for the final set.}}\label{tab:solverprotocols}
\small\setlength{\tabcolsep}{3pt}
\begin{tabular*}{\linewidth}{@{\extracolsep{\fill}}>{\raggedright\arraybackslash}p{0.29\linewidth}lrrl@{}}
\toprule Task & Source & Public & Hidden & Budget \\ \midrule
AtCoder AHC058 & Competition & 105 & 45 & Each 2 \\
Bin packing & Library & 90 & 39 & Pool 1,170 \\
Cluster editing & Competition & 100 & 100 & Each 600 \\
Frequency assignment & Library & 13 & 8 & Each 1,200 \\
AtCoder AHC041 & Competition & 105 & 45 & Each 2 \\
Graph coloring & Library & 132 & 72 & Each 180 \\
Course timetabling & Competition & 12 & 6 & Each 240 \\
Vehicle routing & Library & 80 & 34 & Each 180 \\
Directed feedback vertex set & Competition & 100 & 200 & Each 1,800 \\
Dominating set & Competition & 100 & 100 & Each 300 \\
Glass cutting & Competition & 35 & 15 & Each 180 / 3,600 \\
Graph partitioning & Library & 24 & 13 & Pool 780 \\
Hitting set & Competition & 100 & 100 & Each 300 \\
Sports timetabling & Competition & 29 & 16 & Pool 960 \\
Job-shop scheduling & Library & 198 & 85 & Each 600 \\
Logic synthesis & Competition & 100 & 100 & Pool 4,000 \\
Maximum clique & Library & 44 & 24 & Each 180 \\
Maximum cut & Library & 105 & 45 & Each 600 \\
Maximum satisfiability & Competition & 11 & 200 & Each 300 \\
Machine reassignment & Competition & 20 & 10 & Each 300 \\
Nurse rostering & Competition & 9 & 60 & Each 600 \\
Crossing minimization & Competition & 100 & 100 & Each 1,800 \\
Quadratic assignment & Library & 75 & 41 & Pool 1,230 \\
Quantified Boolean formulas & Competition & 13 & 57 & Each 900 \\
Radiation therapy & Competition & 15 & 36 & Each 1,200 \\
Set cover & Library & 48 & 26 & Pool 780 \\
Syntax-guided synthesis & Competition & 100 & 60 & Each 3,600 \\
Traveling salesperson & Library & 61 & 33 & Each 600 \\
Treewidth & Competition & 100 & 100 & Each 1,800 \\
Vertex-cover certificates & Competition & 100 & 100 & Each 1,800 \\
\midrule
\leavevmode\arxivadded{\textbf{Total}} & \arxivadded{--} & \arxivadded{2,124} & \arxivadded{1,870} & \arxivadded{--} \\
\bottomrule\end{tabular*}
\par\smallskip\raggedright Glass cutting evaluates separate 180- and 3,600-second runs, weighted 0.1 and 0.9. Named source collections are listed in Appendix~\ref{app:tasklist}.
\end{table}

%% file: tables/resource-usage.tex
\begingroup\small
\setlength{\tabcolsep}{4pt}
\begin{longtable}{@{}lrrrrrr@{}}
\caption{\arxivadded{Token usage and duration per attempt in the Full main evaluation. Means use recorded values; $n/N$ gives attempts with measurements over total attempts. M and k denote millions and thousands of tokens.}}\label{tab:resourceuse}\\
\toprule Agent & Input (M) & $n/N$ & Output (k) & $n/N$ & Time (h) & $n/N$ \\ \midrule\endfirsthead
\toprule Agent & Input (M) & $n/N$ & Output (k) & $n/N$ & Time (h) & $n/N$ \\ \midrule\endhead
\bottomrule\endfoot
\multicolumn{7}{@{}l}{\trackmargin{theorem}\textbf{Theorem proving (bare)}}\\*
GPT-6 Astra & 6.51 & 20/20 & 91.1 & 20/20 & 0.68 & 20/20 \\
\arxivadded{Kimi K3.1 Pro} & \arxivadded{253.45} & \arxivadded{60/60} & \arxivadded{1101.9} & \arxivadded{60/60} & \arxivadded{4.90} & \arxivadded{60/60} \\
Kimi K3 & 183.66 & 60/60 & 1599.8 & 60/60 & 9.32 & 60/60 \\
GLM-5.3 & 29.34 & 19/20 & 586.0 & 20/20 & 4.26 & 20/20 \\
\qwenagent{} & 43.03 & 15/20 & 655.3 & 15/20 & 8.22 & 20/20 \\
Grok 4.6 & 93.06 & 20/20 & 478.6 & 20/20 & 3.28 & 20/20 \\
\deepseekagent{} & 15.48 & 58/60 & 553.5 & 59/60 & 1.24 & 60/60 \\
\arxivadded{\deepseeknewagent{}} & \arxivadded{189.86} & \arxivadded{60/60} & \arxivadded{1581.8} & \arxivadded{60/60} & \arxivadded{2.91} & \arxivadded{60/60} \\
\addlinespace
\multicolumn{7}{@{}l}{\trackmargin{solver}\textbf{Solver engineering}}\\*
GPT-6 Astra & 58.00 & 30/30 & 398.1 & 30/30 & 5.22 & 30/30 \\
\arxivadded{Kimi K3.1 Pro} & \arxivadded{98.53} & \arxivadded{90/90} & \arxivadded{293.0} & \arxivadded{90/90} & \arxivadded{3.95} & \arxivadded{90/90} \\
Kimi K3 & 82.69 & 90/90 & 294.6 & 90/90 & 7.34 & 90/90 \\
GLM-5.3 & 89.89 & 30/30 & 298.3 & 30/30 & 8.01 & 30/30 \\
\qwenagent{} & 67.35 & 30/30 & 349.0 & 30/30 & 8.38 & 30/30 \\
Grok 4.6 & 14.37 & 30/30 & 121.5 & 30/30 & 2.26 & 30/30 \\
\deepseekagent{} & 79.45 & 90/90 & 306.9 & 90/90 & 5.09 & 90/90 \\
\arxivadded{\deepseeknewagent{}} & \arxivadded{89.20} & \arxivadded{90/90} & \arxivadded{346.5} & \arxivadded{90/90} & \arxivadded{5.46} & \arxivadded{90/90} \\
\addlinespace
\multicolumn{7}{@{}l}{\trackmargin{construction}\textbf{Constructions}}\\*
GPT-6 Astra & 89.95 & 20/20 & 441.6 & 20/20 & 6.56 & 20/20 \\
\arxivadded{Kimi K3.1 Pro} & \arxivadded{228.70} & \arxivadded{59/60} & \arxivadded{550.7} & \arxivadded{59/60} & \arxivadded{10.06} & \arxivadded{60/60} \\
Kimi K3 & 59.50 & 60/60 & 297.6 & 60/60 & 8.66 & 60/60 \\
GLM-5.3 & 84.23 & 20/20 & 531.0 & 20/20 & 7.16 & 20/20 \\
\qwenagent{} & 81.15 & 20/20 & 931.8 & 20/20 & 10.02 & 20/20 \\
Grok 4.6 & 19.93 & 19/20 & 142.4 & 19/20 & 3.72 & 20/20 \\
\deepseekagent{} & 59.59 & 59/60 & 525.1 & 59/60 & 3.91 & 60/60 \\
\arxivadded{\deepseeknewagent{}} & \arxivadded{123.01} & \arxivadded{57/60} & \arxivadded{610.0} & \arxivadded{57/60} & \arxivadded{4.92} & \arxivadded{59/60} \\
\addlinespace
\multicolumn{7}{@{}l}{\trackmargin{open}\textbf{Open conjectures}}\\*
GPT-6 Astra & 60.36 & 30/30 & 468.1 & 30/30 & 4.58 & 30/30 \\
\arxivadded{Kimi K3.1 Pro} & \arxivadded{115.75} & \arxivadded{90/90} & \arxivadded{364.1} & \arxivadded{90/90} & \arxivadded{2.72} & \arxivadded{90/90} \\
Kimi K3 & 76.66 & 90/90 & 461.8 & 90/90 & 5.16 & 90/90 \\
GLM-5.3 & 93.83 & 30/30 & 447.9 & 30/30 & 3.92 & 30/30 \\
\qwenagent{} & 97.68 & 30/30 & 591.1 & 30/30 & 7.62 & 30/30 \\
Grok 4.6 & 13.47 & 30/30 & 111.6 & 30/30 & 0.96 & 30/30 \\
\deepseekagent{} & 53.34 & 90/90 & 427.1 & 90/90 & 1.65 & 90/90 \\
\arxivadded{\deepseeknewagent{}} & \arxivadded{61.18} & \arxivadded{90/90} & \arxivadded{585.9} & \arxivadded{90/90} & \arxivadded{1.80} & \arxivadded{90/90} \\
\end{longtable}
\endgroup

%% file: sections/appendix-e-results.tex
\section{Complete Results and Diagnostics}
\label{app:results}

\subsection{Task-level performance and repetitions}
\label{app:bare-matrix}

The following tables use the same experiment records as Table~\ref{tab:mainresults}. All scheduled bare-theorem attempts are included using their recorded checker outcomes. \arxivadded{Construction entries average the task-specific scores defined in Appendix~\ref{app:scoring-construction}; their evaluation conditions appear in Appendix~\ref{app:protocol}.}

\begin{figure}[p]
\setlength{\parskip}{0pt}
\setlength{\abovecaptionskip}{5pt}
\centering
\arxivfigure{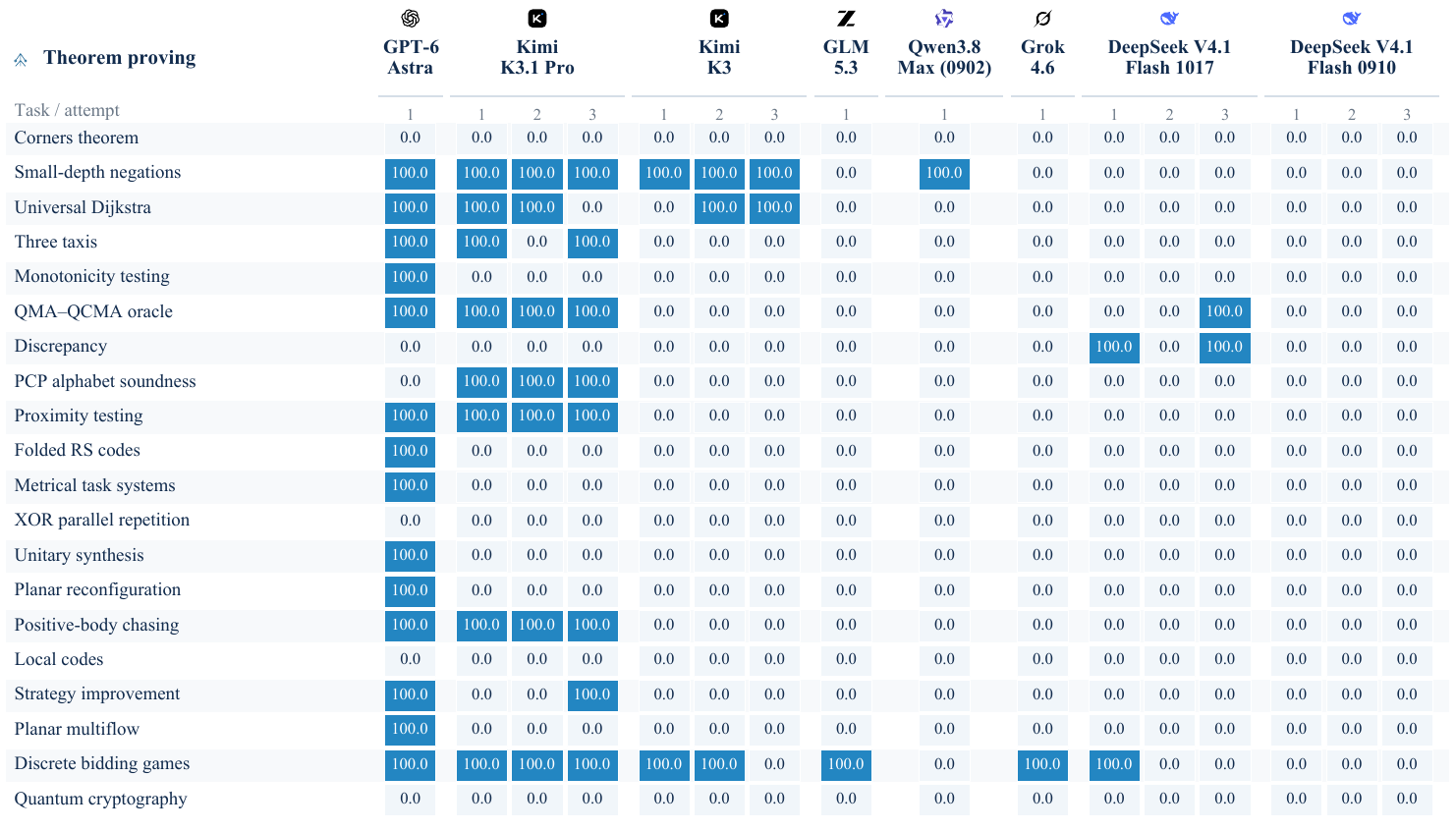}
\caption{\arxivadded{\textbf{Complete theorem proofs across individual attempts.} Each cell represents one scheduled attempt: 100.0\% denotes an accepted complete proof and 0.0\% denotes no accepted complete proof. The eight agents contribute 320 attempts on the same 20 theorems. Columns number independent attempts within each agent. Task indices and theorem names match Table~\ref{tab:theoremdetail}.}}
\label{fig:theoremattempts}
\end{figure}

\begin{figure}[p]
\setlength{\parskip}{0pt}
\setlength{\abovecaptionskip}{5pt}
\centering
\arxivfigure{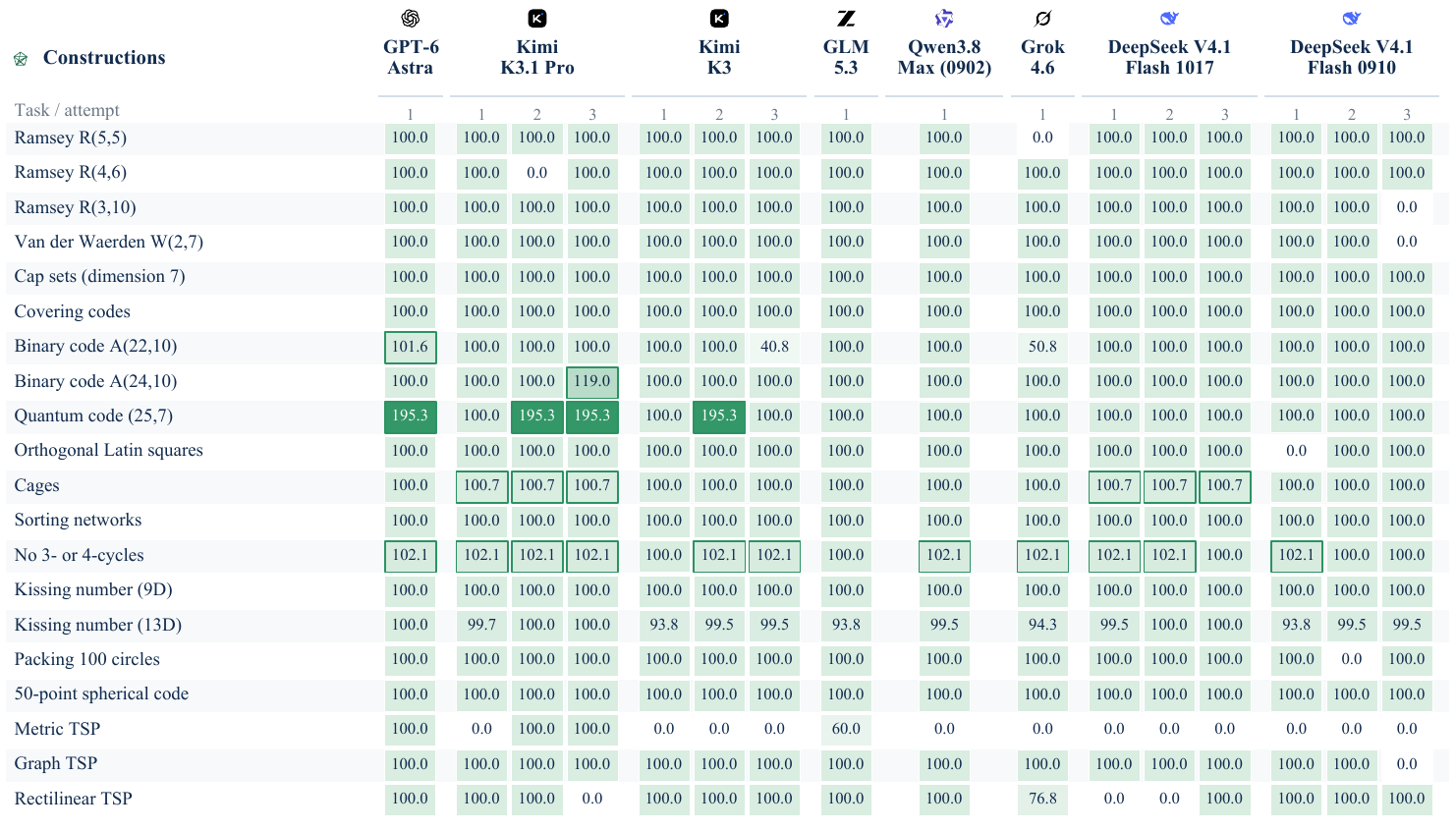}
\caption{\arxivadded{\textbf{Construction scores across individual attempts.} The matrix contains all 320 scheduled Full attempts across eight agents. Cells show percentage scores to one decimal; outlines identify unrounded scores above the reference of 100.0\%. Color follows a common score scale, while mathematical parameters and scoring functions differ by task. Evaluation conditions are documented in Appendix~\ref{app:protocol}.}}
\label{fig:constructionattempts}
\end{figure}

\arxivadded{The tables and Figures~\ref{fig:theoremattempts} and~\ref{fig:constructionattempts} include all eight Full agents. Kimi K3.1 Pro solves nine theorem tasks: all three attempts on six tasks, two on Dijkstra and three taxis, and one on strategy improvement. \deepseeknewagent{} solves discrepancy twice and discrete bidding and the QMA--QCMA oracle task once each.} For theorem proving, Kimi K3 completes the negations theorem in all three attempts, and universal Dijkstra and discrete bidding games in two attempts each. Its seven successful attempts therefore cover three distinct theorems. Grok 4.6 and GLM-5.3 each complete the discrete-bidding theorem, while \qwenagent{} completes the negations theorem. For constructions, the matrix distinguishes repeated attainment of the reference score from variation across attempts, including the occasional scores above \arxivadded{100.0\%}. \arxivadded{Appendix~\ref{app:cases} presents the achieved objects, mathematical parameters, and comparisons with their reference sources.}

{\small
\begin{arxivaddition}
\input{tables/full-task-results-arxiv}
\end{arxivaddition}
}

\noindent\arxivadded{\textbf{Sources of accepted conjecture submissions.} All six accepted attempts in Table~\ref{tab:full-open} adapt OpenAI's published counterexamples and Lean proofs for Ryser and MIRUP \citep{openaiMathRyser,openaiMathBinPacking}. Their trajectories and submitted files identify these sources and the adaptations to the benchmark's fixed definitions.}

\clearpage
\tracksubsection{solver}{Solver objectives and score sensitivity}
\label{app:solverdiagnostics}

\begin{figure}[p]
\setlength{\parskip}{0pt}
\setlength{\abovecaptionskip}{5pt}
\centering
\arxivfigure{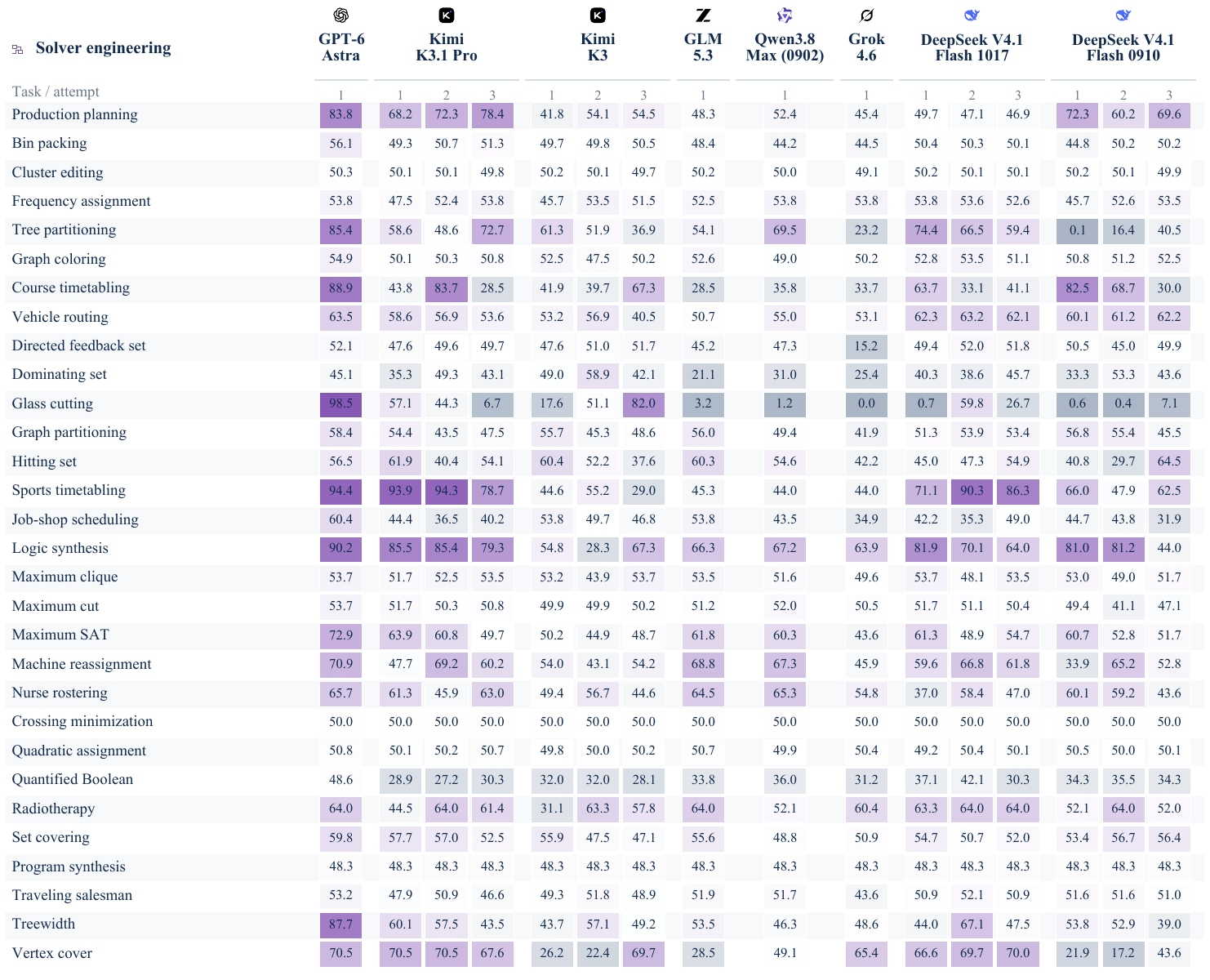}
\caption{\arxivadded{\textbf{Solver results across individual attempts.} All 480 Full attempts across eight agents are shown on the Kimi K3-relative scale; a task uses the same eligible instances for all models. Table~\ref{tab:solverratios} gives original unbounded scores for the initial six agents.}}
\label{fig:solverattempts}
\end{figure}

\arxivadded{Table~\ref{tab:solversummary} compares the two Solver metrics and failed outcomes. Table~\ref{tab:solverratios} gives original scores by attempt; Table~\ref{tab:solverrelative} and Figure~\ref{fig:solverattempts} give task means and individual attempts on the Kimi K3-relative scale.}

{\begin{arxivaddition}\input{tables/solver-summary}\end{arxivaddition}}

\paragraph{Coverage across an instance family.}
Table~\ref{tab:solvercoverage} separates producing valid optimization outputs from meeting a certificate-based target. For each optimization attempt, we compute the fraction of judged instances with accepted feasible solutions, then average attempts within tasks and tasks equally. An attempt contributes to the full-set count only if every instance has a valid result. Across the 27 optimization tasks, Astra's mean validity is \arxivadded{99.3\%, compared with 93.9--97.5\% for the other agents in Table~\ref{tab:solvercoverage}}; it returns valid solutions on every instance in 24 tasks. The three certificate-based tasks instead require the designated correctness or target certificate; their binary completion rates are reported separately. These statistics describe reliability across the instances of each problem, while objective values measure solution quality.

{\begin{arxivaddition}\input{tables/solver-family-coverage}\end{arxivaddition}}

\paragraph{Reference coverage and exclusions.}
The Kimi K3-relative comparison uses a common set of \arxivadded{1,838} judged instances for every model. The excluded 32 instances lack two valid numerical Kimi K3 outcomes, which are required to estimate the sample standard deviation in Appendix~\ref{app:scoring-solver}. On 23 maximum-satisfiability instances, all three attempts are rejected because the output lacks the required objective-value line. Among nine sports-timetabling instances, eight have no feasible Kimi K3 solution and one has only a single feasible solution; the other submissions violate hard constraints. These are recorded unsuccessful solutions, with no missing judging records. We exclude these instances from every model's relative score to maintain a common comparison set. Their outcomes remain in the raw records and failure counts, and the original task-score rules remain unchanged.

All other tasks have full relative-score coverage. The scale floor is active on \arxivadded{1,410} of the \arxivadded{1,838} included instances, so the comparison frequently uses the specified minimum scale rather than the empirical standard deviation alone. \arxivadded{Kimi K3's overall score is 48.7\%: failures receive 0.0\%, and the nonlinear transformation of repeated values need not average to 50.0\%.}

\clearpage
{\begin{arxivaddition}\input{tables/solver-attempt-ratios}\end{arxivaddition}}

Table~\ref{tab:solverscoreexamples} illustrates how raw instance outcomes enter the unbounded ratio score. On treewidth instance \texttt{he134}, Astra attains width 201 against a locally measured FlowCutter reference of 248. Because the baseline exceeds this reference by only one, the score is \arxivadded{4,800.0\%}. \arxivadded{The narrow baseline--reference difference amplifies the improvement over this locally measured reference.} On logic-synthesis instance \texttt{ex37}, \qwenagent{}\textquotesingle s submitted circuit passes validation but has a much larger objective than the baseline, producing a large negative contribution. These values are retained in the task mean.

{\begin{arxivaddition}\input{tables/solver-score-examples}\end{arxivaddition}}

\arxivadded{Extreme instance scores can strongly affect a task mean even when tasks have equal aggregate weight. The Kimi K3-relative score bounds their contributions, while raw objectives show the size and direction of each performance difference.}

\clearpage
\tracksubsection{theorem}{Proof-guidance scores}
\label{app:checkpoints}

\begin{figure}[htbp]
\centering
\arxivfigure{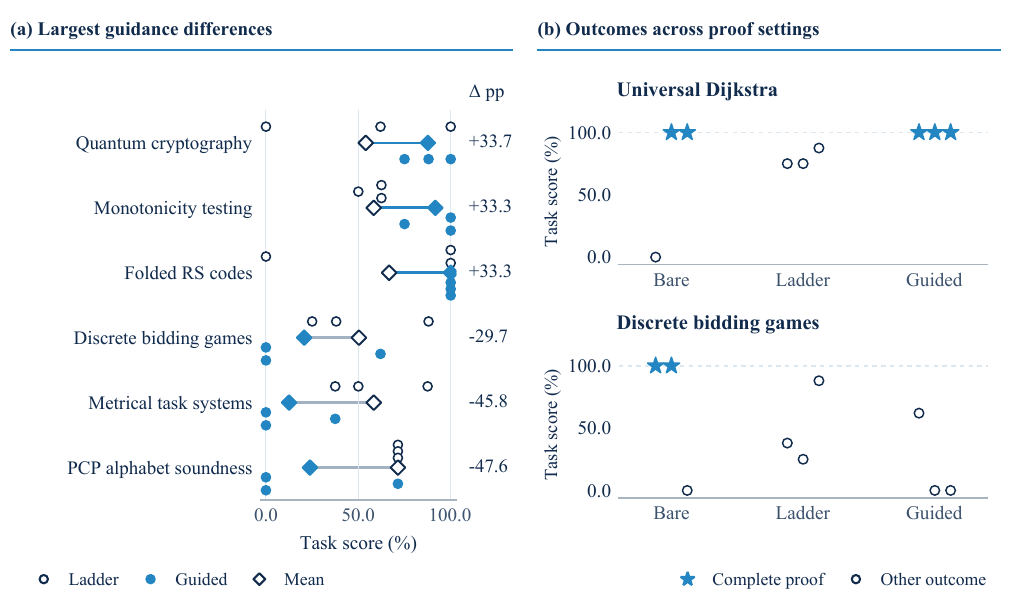}
\caption{\textbf{Kimi K3 performance across proof settings.} (a) The six largest absolute guided-minus-ladder mean differences among 20 theorems. Circles show individual attempts; diamonds mark means, with differences in percentage points. (b) Universal Dijkstra and discrete bidding games, which retains bare success with incomplete ladder/guided attempts. Each marker is an independent attempt; stars identify complete proofs. Bare receives binary credit, while ladder/guided also award partial checkpoint credit. Horizontal offsets separate the three attempts. Full results and checkpoint details appear in Appendix~\ref{app:checkpoints}.}
\label{fig:proofdiagnostics}
\end{figure}

Table~\ref{tab:proofsettings} compares complete-proof outcomes on the same 20 targets. From bare to ladder to guided, the numbers of theorems solved at least once increase from 3 to 4 to 7 for Kimi K3, and from 0 to 2 to 6 for \deepseekagent{}. Kimi K3 has equal per-attempt success under bare and ladder (7/60), then rises to 18/60 under guided; \deepseekagent{} has 0/60, 2/60, and 11/60, respectively. Each setting uses three attempts per theorem.

\input{tables/proof-settings}
\Needspace{12\baselineskip}
{\small\begin{arxivaddition}\input{tables/guidance-task-results}\end{arxivaddition}}

Table~\ref{tab:guidancedetail} reports task scores, including partial credit, on all 20 theorems. Kimi K3 averages \arxivadded{67.0\% under guided and 60.4\% under ladder, a difference of 6.6 percentage points}; \deepseekagent{} averages \arxivadded{57.8\% and 49.1\%, a difference of 8.7 points}. Guided has a higher mean on 14 theorems for each model; Kimi K3 has five decreases and one tie, while \deepseekagent{} has four decreases and two ties. Each theorem contributes equally after averaging its three attempts. Pooling all checkpoints across tasks would instead give more weight to theorems with more checkpoints. Appendix~\ref{app:guidancepopulation} describes the evaluation protocol; the figures below examine Kimi K3 in more detail.

Figure~\ref{fig:guidanceattempts} separates the three attempts within each level. For example, monotonicity testing improves from 62.5\%, 62.5\%, and \arxivadded{50.0\% under ladder to 100.0\%, 100.0\%, and 75.0\%} under guided. In universal Dijkstra, ladder scores are \arxivadded{75.0\%, 75.0\%, and 87.5\%}, while all three guided scores are \arxivadded{100.0\%}. \arxivadded{The individual-attempt plots show how these mean differences arise across the three independent runs in each setting.}

\arxivadded{Table~\ref{tab:proofsettings} reports complete-proof success separately from conditional checkpoint coverage, using the two verification modes defined in Appendix~\ref{app:scoring-lean}. The selected cases below show how these outcomes correspond to the submitted proofs.}

\paragraph{Bare success across proof settings.}
Table~\ref{tab:k3proofcases} lists all three theorems solved by Kimi K3 in at least one bare attempt. Discrete bidding games retains bare success with no full-score ladder or guided attempt. Dijkstra has three accepted guided proofs; both cases appear in Figure~\ref{fig:proofdiagnostics}b. Small-depth negations provides a contrasting case: bare succeeds three times, ladder reaches full credit once, and guided reaches it three times. \arxivadded{Values use the reported package scores, displayed to one decimal.}

\begin{table}[ht]
\centering
\caption{\arxivadded{Proof-guidance outcomes for the three theorems solved by Kimi K3 in the bare setting. Bare gives complete proofs / attempts; ladder and guided list task scores (\%). Attempts are independent across settings.}}
\label{tab:k3proofcases}
\small
\setlength{\tabcolsep}{5pt}
\begin{tabular}{lccc}
\toprule
Theorem & Bare & Ladder (\%) & Guided (\%) \\
\midrule
Universal Dijkstra & 2/3 & \arxivadded{75.0, 75.0, 87.5} & \arxivadded{100.0, 100.0, 100.0} \\
Discrete bidding games & 2/3 & \arxivadded{38.0, 25.0, 88.0} & \arxivadded{62.0, 0.0, 0.0} \\
Small-depth negations & 3/3 & \arxivadded{57.1, 85.7, 100.0} & \arxivadded{100.0, 100.0, 100.0} \\
\bottomrule
\end{tabular}
\end{table}

\paragraph{Statement and scoring checks.}
For Dijkstra, static comparison across the bare and scaffolded task versions confirms that the main statement and shared definitions agree after renaming the external interval-DAG bound and its bound variable; the scaffolded tasks additionally define the heap operations used by their checkpoints. For bidding games, the main statement and shared definitions agree, with additional helper declarations in the scaffolded tasks. The negations statements and shared definitions agree without renaming. These checks establish a common formal target for each case while retaining the differences in supplied context, prompts, and verifiers. In both selected tasks, the source verifiers award full credit to an accepted direct main proof as well as to an assembled proof. \arxivadded{Submissions can therefore earn full credit through either proof route.}

\begin{figure}[ht]
\centering
\includegraphics[width=\linewidth]{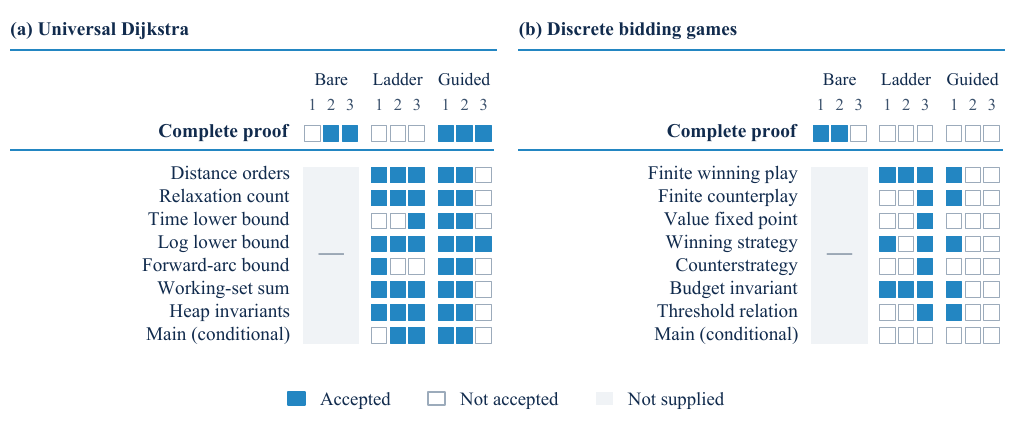}
\caption{\textbf{Complete-proof and checkpoint acceptance for the Kimi K3 cases.} The top row records complete proofs. Each lower row is a conditional checkpoint, including the main-theorem slot. Columns number independent attempts within each setting; gray denotes checkpoints absent from bare. Dijkstra's third guided submission has an accepted complete proof despite failures in seven isolated checkpoint checks. In bidding's third ladder attempt, all intermediate checkpoints pass while the main slot does not.}
\label{fig:k3proofcheckpoints}
\end{figure}

\begin{figure}[p]
\centering
\arxivfigure{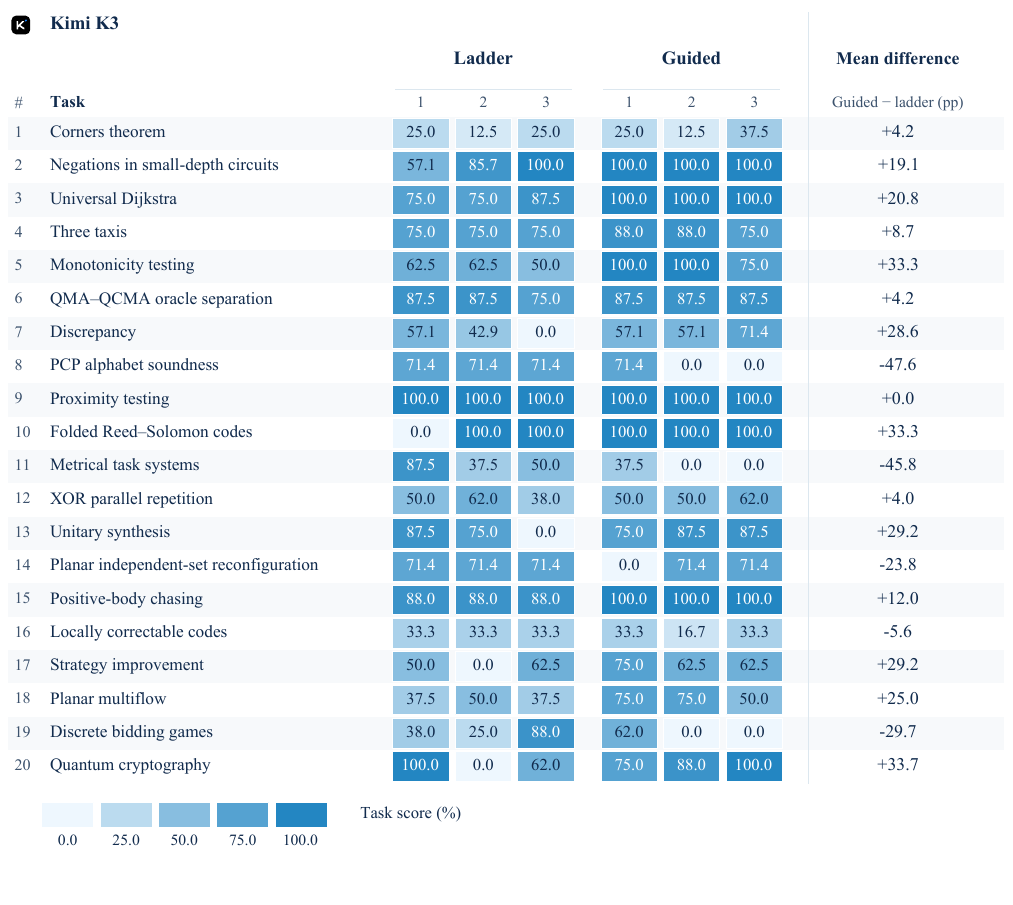}
\caption{\textbf{Kimi K3 task scores across individual attempts.} Each cell shows a package score as a percentage, \arxivadded{rounded to one decimal}. Mean differences use the unrounded scores and are given in percentage points. The 20 paired theorems contribute 60 attempts per setting. Scores combine partial checkpoint credit with full credit for accepted complete proofs; complete-proof outcomes are shown separately for selected cases in Figure~\ref{fig:k3proofcheckpoints}.}
\label{fig:guidanceattempts}
\end{figure}

\paragraph{Checkpoint evidence.}
Figure~\ref{fig:k3proofcheckpoints} separates complete-proof acceptance from isolated checkpoint acceptance. All three Dijkstra guided submissions have accepted complete proofs. The first two also pass all eight isolated checkpoints; the third passes only one because extracting the checkpoint bodies disrupts its Lean scopes. \arxivadded{The separate complete-proof check preserves full credit for this submission.} Bidding's second ladder attempt passes two checkpoints and receives a score of \arxivadded{25.0\%}, without an accepted complete proof. Its third ladder attempt passes all intermediate checkpoints but leaves the main-theorem checkpoint unproved. These distinctions separate completion of the target theorem from progress along the supplied proof structure.

\paragraph{Bidding-game case: a direct invariant proof.}
The target asks for a budget-agnostic winning strategy, whose action depends on the current vertex and tie-breaking advantage rather than the exact available budget. Theorem~27 of \citet{bidding} constructs this strategy by applying value iteration at a trimmed threshold budget. Its proof tracks the excess budget, which is nondecreasing and bounded, and partitions play into intervals on which this quantity stays constant. This gives the initial energy bound $M=(k+1)(W+H)$, where $k$ is the total integer budget, $W$ bounds absolute edge weights, and $H$ is the maximum energy threshold at the threshold budgets.

One successful bare submission uses the same strategy and bound but organizes the argument as a single inductive invariant. For configuration $c=(v,B)$, let $e(c)$ be the integer budget above the trimmed threshold, and let $u(c)$ be the value-iteration energy threshold at that trimmed budget. Its helper lemma proves, for every legal transition $c\to c'$ consistent with the strategy,
\begin{equation}
u(c)+w(v,v')+(W+H)\bigl(e(c')-e(c)\bigr)\geq u(c').
\end{equation}
Writing $S_n$ for the sum of the first $n$ edge weights, induction yields
\begin{equation}
M+S_n\geq u(c_n)+M-(W+H)e(c_n)-H\geq 0,
\end{equation}
where the last step uses $e(c_n)\leq k$ and $u(c_n)\geq0$. This establishes the energy guarantee directly for every play prefix. The submitted proof defines its own supporting lemmas rather than completing the supplied checkpoint sequence. Independent rechecks accept both successful bare submissions with only Lean's standard axioms. \arxivadded{The submission reproduces the source theorem\textquotesingle s strategy and energy bound through a single inductive invariant, providing an alternative organization of the formal proof.}

\paragraph{Seven accepted checkpoints with the target unfinished.}
The third ladder attempt provides a concrete comparison. Its final judge log accepts all seven intermediate checkpoints, including the value-iteration fixed point, the budget invariant, and the threshold-average property. The main theorem remains a \texttt{sorry} placeholder, and both complete-proof checks fail; \arxivadded{the exact checkpoint fraction $7/8$ corresponds to 87.5\% (the historical score export rounds this to 88.0\% in Table~\ref{tab:k3proofcases})}. Near the end of the 12-hour budget, the last build still reports the unfinished declaration and the final response records the target as incomplete. Thus this attempt's partial credit comes from intermediate obligations that were accepted conditionally, while the final strategy-existence argument was not completed. \arxivadded{Against the same formal target, the successful bare submission completes its invariant argument, while this independent ladder attempt leaves the final strategy-existence obligation open.} In the second guided attempt, the model proposed the auxiliary lemma \texttt{muN\_kStar\_le\_xN}, then identified a counterexample involving tie-breaking. The lemma was absent from the supplied task; the attempt exhausted its budget while revising the proof. A third guided attempt worked in auxiliary Lean files but left the designated submission unchanged, yielding no accepted checkpoints. This distinguishes unfinished mathematical obligations from failure to integrate working files into the submitted artifact.

\clearpage

%% file: tables/full-task-results-arxiv.tex
\begingroup\footnotesize\setlength{\tabcolsep}{2.2pt}
\begin{longtable}{@{}p{0.27\linewidth}rrrrrrrr@{}}
\caption{Theorem-proving results on TCSExam-Full. Entries give complete proofs / attempts.}\label{tab:theoremdetail} \\
\toprule
\tablehead[l]{Task} & \tablehead{GPT-6\\Astra} & \tablehead{Kimi\\K3.1 Pro} & \tablehead{Kimi\\K3} & \tablehead{GLM-5.3} & \arxivadded{\tablehead{Qwen3.8\\Max (0902)}} & \tablehead{Grok\\4.6} & \tablehead{DeepSeek\\V4.1 Flash\\1017} & \arxivadded{\tablehead{DeepSeek\\V4.1 Flash\\0910}} \\
\midrule\endfirsthead
\toprule
\tablehead[l]{Task} & \tablehead{GPT-6\\Astra} & \tablehead{Kimi\\K3.1 Pro} & \tablehead{Kimi\\K3} & \tablehead{GLM-5.3} & \arxivadded{\tablehead{Qwen3.8\\Max (0902)}} & \tablehead{Grok\\4.6} & \tablehead{DeepSeek\\V4.1 Flash\\1017} & \arxivadded{\tablehead{DeepSeek\\V4.1 Flash\\0910}} \\
\midrule\endhead
1. Corners theorem & 0/1 & 0/3 & 0/3 & 0/1 & 0/1 & 0/1 & 0/3 & 0/3 \\
2. Small-depth negations & 1/1 & 3/3 & 3/3 & 0/1 & 1/1 & 0/1 & 0/3 & 0/3 \\
3. Universal Dijkstra & 1/1 & 2/3 & 2/3 & 0/1 & 0/1 & 0/1 & 0/3 & 0/3 \\
4. Three taxis & 1/1 & 2/3 & 0/3 & 0/1 & 0/1 & 0/1 & 0/3 & 0/3 \\
5. Monotonicity testing & 1/1 & 0/3 & 0/3 & 0/1 & 0/1 & 0/1 & 0/3 & 0/3 \\
6. QMA--QCMA separation & 1/1 & 3/3 & 0/3 & 0/1 & 0/1 & 0/1 & 1/3 & 0/3 \\
7. Discrepancy & 0/1 & 0/3 & 0/3 & 0/1 & 0/1 & 0/1 & 2/3 & 0/3 \\
8. PCP alphabet soundness & 0/1 & 3/3 & 0/3 & 0/1 & 0/1 & 0/1 & 0/3 & 0/3 \\
9. Proximity testing & 1/1 & 3/3 & 0/3 & 0/1 & 0/1 & 0/1 & 0/3 & 0/3 \\
10. Folded RS codes & 1/1 & 0/3 & 0/3 & 0/1 & 0/1 & 0/1 & 0/3 & 0/3 \\
11. Metrical task systems & 1/1 & 0/3 & 0/3 & 0/1 & 0/1 & 0/1 & 0/3 & 0/3 \\
12. XOR parallel repetition & 0/1 & 0/3 & 0/3 & 0/1 & 0/1 & 0/1 & 0/3 & 0/3 \\
13. Unitary synthesis & 1/1 & 0/3 & 0/3 & 0/1 & 0/1 & 0/1 & 0/3 & 0/3 \\
14. Planar reconfiguration & 1/1 & 0/3 & 0/3 & 0/1 & 0/1 & 0/1 & 0/3 & 0/3 \\
15. Positive-body chasing & 1/1 & 3/3 & 0/3 & 0/1 & 0/1 & 0/1 & 0/3 & 0/3 \\
16. Locally correctable codes & 0/1 & 0/3 & 0/3 & 0/1 & 0/1 & 0/1 & 0/3 & 0/3 \\
17. Strategy improvement & 1/1 & 1/3 & 0/3 & 0/1 & 0/1 & 0/1 & 0/3 & 0/3 \\
18. Planar multiflow & 1/1 & 0/3 & 0/3 & 0/1 & 0/1 & 0/1 & 0/3 & 0/3 \\
19. Discrete bidding games & 1/1 & 3/3 & 2/3 & 1/1 & 0/1 & 1/1 & 1/3 & 0/3 \\
20. Quantum cryptography & 0/1 & 0/3 & 0/3 & 0/1 & 0/1 & 0/1 & 0/3 & 0/3 \\
\bottomrule\end{longtable}
\begin{longtable}{@{}p{0.27\linewidth}rrrrrrrr@{}}
\caption{Solver results on TCSExam-Full. Entries are mean Kimi K3-relative scores (\%).}\label{tab:solverrelative} \\
\toprule
\tablehead[l]{Task} & \tablehead{GPT-6\\Astra} & \tablehead{Kimi\\K3.1 Pro} & \tablehead{Kimi\\K3} & \tablehead{GLM-5.3} & \arxivadded{\tablehead{Qwen3.8\\Max (0902)}} & \tablehead{Grok\\4.6} & \tablehead{DeepSeek\\V4.1 Flash\\1017} & \arxivadded{\tablehead{DeepSeek\\V4.1 Flash\\0910}} \\
\midrule\endfirsthead
\toprule
\tablehead[l]{Task} & \tablehead{GPT-6\\Astra} & \tablehead{Kimi\\K3.1 Pro} & \tablehead{Kimi\\K3} & \tablehead{GLM-5.3} & \arxivadded{\tablehead{Qwen3.8\\Max (0902)}} & \tablehead{Grok\\4.6} & \tablehead{DeepSeek\\V4.1 Flash\\1017} & \arxivadded{\tablehead{DeepSeek\\V4.1 Flash\\0910}} \\
\midrule\endhead
Production planning & 83.8 & 73.0 & 50.1 & 48.3 & 52.4 & 45.4 & 47.9 & 67.3 \\
Bin packing & 56.1 & 50.4 & 50.0 & 48.4 & 44.2 & 44.5 & 50.3 & 48.4 \\
Cluster editing & 50.3 & 50.0 & 50.0 & 50.2 & 50.0 & 49.1 & 50.2 & 50.1 \\
Frequency assignment & 53.8 & 51.2 & 50.2 & 52.5 & 53.8 & 53.8 & 53.4 & 50.6 \\
Tree partitioning & 85.4 & 60.0 & 50.0 & 54.1 & 69.5 & 23.2 & 66.8 & 19.0 \\
Graph coloring & 54.9 & 50.4 & 50.1 & 52.6 & 49.0 & 50.2 & 52.5 & 51.5 \\
Course timetabling & 88.9 & 52.0 & 49.6 & 28.5 & 35.8 & 33.7 & 46.0 & 60.4 \\
Vehicle routing & 63.5 & 56.4 & 50.2 & 50.7 & 55.0 & 53.1 & 62.5 & 61.2 \\
Directed feedback set & 52.1 & 49.0 & 50.1 & 45.2 & 47.3 & 15.2 & 51.0 & 48.4 \\
Dominating set & 45.1 & 42.5 & 50.0 & 21.1 & 31.0 & 25.4 & 41.5 & 43.4 \\
Glass cutting & 98.5 & 36.0 & 50.2 & 3.2 & 1.2 & 0.0 & 29.1 & 2.7 \\
Graph partitioning & 58.4 & 48.4 & 49.9 & 56.0 & 49.4 & 41.9 & 52.9 & 52.6 \\
Hitting set & 56.5 & 52.2 & 50.0 & 60.3 & 54.6 & 42.2 & 49.1 & 45.0 \\
Sports timetabling & 94.4 & 89.0 & 42.9 & 45.3 & 44.0 & 44.0 & 82.5 & 58.8 \\
Job-shop scheduling & 60.4 & 40.4 & 50.1 & 53.8 & 43.5 & 34.9 & 42.1 & 40.1 \\
Logic synthesis & 90.2 & 83.4 & 50.2 & 66.3 & 67.2 & 63.9 & 72.0 & 68.7 \\
Maximum clique & 53.7 & 52.6 & 50.3 & 53.5 & 51.6 & 49.6 & 51.7 & 51.2 \\
Maximum cut & 53.7 & 51.0 & 50.0 & 51.2 & 52.0 & 50.5 & 51.1 & 45.9 \\
Maximum SAT & 72.9 & 58.1 & 47.9 & 61.8 & 60.3 & 43.6 & 55.0 & 55.1 \\
Machine reassignment & 70.9 & 59.0 & 50.4 & 68.8 & 67.3 & 45.9 & 62.7 & 50.6 \\
Nurse rostering & 65.7 & 56.7 & 50.2 & 64.5 & 65.3 & 54.8 & 47.5 & 54.3 \\
Crossing minimization & 50.0 & 50.0 & 50.0 & 50.0 & 50.0 & 50.0 & 50.0 & 50.0 \\
Quadratic assignment & 50.8 & 50.3 & 50.0 & 50.7 & 49.9 & 50.4 & 49.9 & 50.2 \\
Quantified Boolean & 48.6 & 28.8 & 30.7 & 33.8 & 36.0 & 31.2 & 36.5 & 34.7 \\
Radiotherapy & 64.0 & 56.7 & 50.7 & 64.0 & 52.1 & 60.4 & 63.8 & 56.0 \\
Set covering & 59.8 & 55.7 & 50.2 & 55.6 & 48.8 & 50.9 & 52.5 & 55.5 \\
Program synthesis & 48.3 & 48.3 & 48.3 & 48.3 & 48.3 & 48.3 & 48.3 & 48.3 \\
Traveling salesman & 53.2 & 48.5 & 50.0 & 51.9 & 51.7 & 43.6 & 51.3 & 51.4 \\
Treewidth & 87.7 & 53.7 & 50.0 & 53.5 & 46.3 & 48.6 & 52.9 & 48.6 \\
Vertex cover & 70.5 & 69.5 & 39.4 & 28.5 & 49.1 & 65.4 & 68.8 & 27.6 \\
\bottomrule\end{longtable}
\begin{longtable}{@{}p{0.27\linewidth}rrrrrrrr@{}}
\caption{Construction results on TCSExam-Full. Entries are mean task scores (\%).}\label{tab:constructiondetail} \\
\toprule
\tablehead[l]{Task} & \tablehead{GPT-6\\Astra} & \tablehead{Kimi\\K3.1 Pro} & \tablehead{Kimi\\K3} & \tablehead{GLM-5.3} & \arxivadded{\tablehead{Qwen3.8\\Max (0902)}} & \tablehead{Grok\\4.6} & \tablehead{DeepSeek\\V4.1 Flash\\1017} & \arxivadded{\tablehead{DeepSeek\\V4.1 Flash\\0910}} \\
\midrule\endfirsthead
\toprule
\tablehead[l]{Task} & \tablehead{GPT-6\\Astra} & \tablehead{Kimi\\K3.1 Pro} & \tablehead{Kimi\\K3} & \tablehead{GLM-5.3} & \arxivadded{\tablehead{Qwen3.8\\Max (0902)}} & \tablehead{Grok\\4.6} & \tablehead{DeepSeek\\V4.1 Flash\\1017} & \arxivadded{\tablehead{DeepSeek\\V4.1 Flash\\0910}} \\
\midrule\endhead
Ramsey R(5,5) & 100.0 & 100.0 & 100.0 & 100.0 & 100.0 & 0.0 & 100.0 & 100.0 \\
Ramsey R(4,6) & 100.0 & 66.7 & 100.0 & 100.0 & 100.0 & 100.0 & 100.0 & 100.0 \\
Ramsey R(3,10) & 100.0 & 100.0 & 100.0 & 100.0 & 100.0 & 100.0 & 100.0 & 66.7 \\
Van der Waerden W(2,7) & 100.0 & 100.0 & 100.0 & 100.0 & 100.0 & 100.0 & 100.0 & 66.7 \\
Cap sets (dimension 7) & 100.0 & 100.0 & 100.0 & 100.0 & 100.0 & 100.0 & 100.0 & 100.0 \\
Covering codes & 100.0 & 100.0 & 100.0 & 100.0 & 100.0 & 100.0 & 100.0 & 100.0 \\
Binary code A(22,10) & 101.6 & 100.0 & 80.3 & 100.0 & 100.0 & 50.8 & 100.0 & 100.0 \\
Binary code A(24,10) & 100.0 & 106.3 & 100.0 & 100.0 & 100.0 & 100.0 & 100.0 & 100.0 \\
Quantum code (25,7) & 195.3 & 163.5 & 131.8 & 100.0 & 100.0 & 100.0 & 100.0 & 100.0 \\
Orthogonal Latin squares & 100.0 & 100.0 & 100.0 & 100.0 & 100.0 & 100.0 & 100.0 & 66.7 \\
Cages & 100.0 & 100.7 & 100.0 & 100.0 & 100.0 & 100.0 & 100.7 & 100.0 \\
Sorting networks & 100.0 & 100.0 & 100.0 & 100.0 & 100.0 & 100.0 & 100.0 & 100.0 \\
No 3- or 4-cycles & 102.1 & 102.1 & 101.4 & 100.0 & 102.1 & 102.1 & 101.4 & 100.7 \\
Kissing number (9D) & 100.0 & 100.0 & 100.0 & 100.0 & 100.0 & 100.0 & 100.0 & 100.0 \\
Kissing number (13D) & 100.0 & 99.9 & 97.6 & 93.8 & 99.5 & 94.3 & 99.8 & 97.6 \\
Packing 100 circles & 100.0 & 100.0 & 100.0 & 100.0 & 100.0 & 100.0 & 100.0 & 66.7 \\
50-point spherical code & 100.0 & 100.0 & 100.0 & 100.0 & 100.0 & 100.0 & 100.0 & 100.0 \\
Metric TSP & 100.0 & 66.7 & 0.0 & 60.0 & 0.0 & 0.0 & 0.0 & 0.0 \\
Graph TSP & 100.0 & 100.0 & 100.0 & 100.0 & 100.0 & 100.0 & 100.0 & 66.7 \\
Rectilinear TSP & 100.0 & 66.7 & 100.0 & 100.0 & 100.0 & 76.8 & 33.3 & 100.0 \\
\bottomrule\end{longtable}
\begin{longtable}{@{}p{0.27\linewidth}rrrrrrrr@{}}
\caption{Open-conjecture results on TCSExam-Full. Accepted proofs or refutations / attempts.}\label{tab:full-open} \\
\toprule
\tablehead[l]{Task} & \tablehead{GPT-6\\Astra} & \tablehead{Kimi\\K3.1 Pro} & \tablehead{Kimi\\K3} & \tablehead{GLM-5.3} & \arxivadded{\tablehead{Qwen3.8\\Max (0902)}} & \tablehead{Grok\\4.6} & \tablehead{DeepSeek\\V4.1 Flash\\1017} & \arxivadded{\tablehead{DeepSeek\\V4.1 Flash\\0910}} \\
\midrule\endfirsthead
\toprule
\tablehead[l]{Task} & \tablehead{GPT-6\\Astra} & \tablehead{Kimi\\K3.1 Pro} & \tablehead{Kimi\\K3} & \tablehead{GLM-5.3} & \arxivadded{\tablehead{Qwen3.8\\Max (0902)}} & \tablehead{Grok\\4.6} & \tablehead{DeepSeek\\V4.1 Flash\\1017} & \arxivadded{\tablehead{DeepSeek\\V4.1 Flash\\0910}} \\
\midrule\endhead
Aaronson--Ambainis & 0/1 & 0/3 & 0/3 & 0/1 & 0/1 & 0/1 & 0/3 & 0/3 \\
Alon--Jaeger--Tarsi & 0/1 & 0/3 & 0/3 & 0/1 & 0/1 & 0/1 & 0/3 & 0/3 \\
Bin packing MIRUP & 0/1 & 0/3 & 0/3 & 0/1 & 0/1 & 0/1 & 3/3 & 0/3 \\
Caccetta--Haggkvist & 0/1 & 0/3 & 0/3 & 0/1 & 0/1 & 0/1 & 0/3 & 0/3 \\
Deterministic vs. randomized queries & 0/1 & 0/3 & 0/3 & 0/1 & 0/1 & 0/1 & 0/3 & 0/3 \\
Evasiveness & 0/1 & 0/3 & 0/3 & 0/1 & 0/1 & 0/1 & 0/3 & 0/3 \\
Explicit Ramsey graphs & 0/1 & 0/3 & 0/3 & 0/1 & 0/1 & 0/1 & 0/3 & 0/3 \\
Explicit rigid matrices & 0/1 & 0/3 & 0/3 & 0/1 & 0/1 & 0/1 & 0/3 & 0/3 \\
Fourier entropy--influence & 0/1 & 0/3 & 0/3 & 0/1 & 0/1 & 0/1 & 0/3 & 0/3 \\
Frege proof length & 0/1 & 0/3 & 0/3 & 0/1 & 0/1 & 0/1 & 0/3 & 0/3 \\
Frege vs. extended Frege & 0/1 & 0/3 & 0/3 & 0/1 & 0/1 & 0/1 & 0/3 & 0/3 \\
KRW composition & 0/1 & 0/3 & 0/3 & 0/1 & 0/1 & 0/1 & 0/3 & 0/3 \\
Log-rank & 0/1 & 0/3 & 0/3 & 0/1 & 0/1 & 0/1 & 0/3 & 0/3 \\
Lovasz Hamiltonicity & 0/1 & 0/3 & 0/3 & 0/1 & 0/1 & 0/1 & 0/3 & 0/3 \\
Mansour & 0/1 & 0/3 & 0/3 & 0/1 & 0/1 & 0/1 & 0/3 & 0/3 \\
MDS codes (non-prime) & 0/1 & 0/3 & 0/3 & 0/1 & 0/1 & 0/1 & 0/3 & 0/3 \\
NEXP vs. $\mathrm{TC}^0$ & 0/1 & 0/3 & 0/3 & 0/1 & 0/1 & 0/1 & 0/3 & 0/3 \\
Branching programs & 0/1 & 0/3 & 0/3 & 0/1 & 0/1 & 0/1 & 0/3 & 0/3 \\
NP vs. $\mathrm{ACC}^0$ & 0/1 & 0/3 & 0/3 & 0/1 & 0/1 & 0/1 & 0/3 & 0/3 \\
Matrix multiplication exponent & 0/1 & 0/3 & 0/3 & 0/1 & 0/1 & 0/1 & 0/3 & 0/3 \\
Online bin packing & 0/1 & 0/3 & 0/3 & 0/1 & 0/1 & 0/1 & 0/3 & 0/3 \\
Online matching & 0/1 & 0/3 & 0/3 & 0/1 & 0/1 & 0/1 & 0/3 & 0/3 \\
Permanent vs. determinant & 0/1 & 0/3 & 0/3 & 0/1 & 0/1 & 0/1 & 0/3 & 0/3 \\
Integer PFR & 0/1 & 0/3 & 0/3 & 0/1 & 0/1 & 0/1 & 0/3 & 0/3 \\
Planar GNRS & 0/1 & 0/3 & 0/3 & 0/1 & 0/1 & 0/1 & 0/3 & 0/3 \\
Polar scaling exponent & 0/1 & 0/3 & 0/3 & 0/1 & 0/1 & 0/1 & 0/3 & 0/3 \\
Ryser & 0/1 & 1/3 & 0/3 & 0/1 & 0/1 & 0/1 & 2/3 & 0/3 \\
Total coloring & 0/1 & 0/3 & 0/3 & 0/1 & 0/1 & 0/1 & 0/3 & 0/3 \\
Tuza & 0/1 & 0/3 & 0/3 & 0/1 & 0/1 & 0/1 & 0/3 & 0/3 \\
XOR log-rank & 0/1 & 0/3 & 0/3 & 0/1 & 0/1 & 0/1 & 0/3 & 0/3 \\
\bottomrule\end{longtable}
\endgroup

%% file: tables/solver-summary.tex
\begin{table}[ht]
\ifmarkarxivchanges\color{red}\fi
\centering\small
\caption{Solver scores and unsuccessful evaluations. Task scores use unbounded difference ratios or solve rates. Failed evaluations include rejected solutions, crashes, and timeouts; counts cover all judged instances.}
\label{tab:solversummary}
\setlength{\tabcolsep}{4pt}
\begin{tabular*}{\linewidth}{@{\extracolsep{\fill}}lrrr@{}}
\toprule
Model & Relative (\%) & Task score (\%) & Failed / judged \\ \midrule
GPT-6 Astra & 64.7 & 107.1 & 44/1870 \\
Kimi K3 & 48.7 & 78.1 & 313/5610 \\
GLM-5.3 & 49.1 & 81.1 & 108/1870 \\
\qwenagent{} & 49.2 & 64.7 & 86/1870 \\
Grok 4.6 & 43.8 & 74.6 & 227/1870 \\
\deepseekagent{} & 48.3 & 77.9 & 347/5610 \\
\bottomrule\end{tabular*}
\end{table}

%% file: tables/solver-family-coverage.tex
\begin{table}[ht]
\ifmarkarxivchanges\color{red}\fi
\centering
\caption{Validity and instance coverage of submitted solvers. Validity gives equal weight to each of the 27 optimization tasks; Full sets count attempts with valid results on every instance. Certified averages completion rates for quantified Boolean formulas, syntax-guided synthesis, and vertex-cover certificates.}
\label{tab:solvercoverage}
\small
\setlength{\tabcolsep}{4pt}
\begin{tabular*}{\linewidth}{@{\extracolsep{\fill}}lrrr@{}}
\toprule
Model & Validity (\%) & Full sets / attempts & Certified (\%) \\ \midrule
GPT-6 Astra & 99.3 & 24/27 & 90.1 \\
Kimi K3 & 97.2 & 75/81 & 72.1 \\
GLM-5.3 & 97.5 & 25/27 & 68.8 \\
\qwenagent{} & 97.2 & 25/27 & 77.7 \\
Grok 4.6 & 93.9 & 23/27 & 81.6 \\
\deepseekagent{} & 97.2 & 74/81 & 68.2 \\
\bottomrule
\end{tabular*}
\end{table}

%% file: tables/solver-attempt-ratios.tex
\begingroup\footnotesize
\setlength{\tabcolsep}{2pt}
\begin{longtable}{@{}>{\raggedright\arraybackslash}p{0.265\linewidth}rrrrrrrrrr@{}}
\caption{Per-attempt Solver task scores (\%). Optimization tasks use unbounded difference ratios; decision tasks use solve rates.}\label{tab:solverratios}\\
\toprule
\tablehead[l]{Task} & \tablehead{Astra} & \multicolumn{3}{c}{\tablehead{Kimi K3}} & \tablehead{GLM} & \arxivadded{\tablehead{Qwen3.8\\Max (0902)}} & \tablehead{Grok} & \multicolumn{3}{c}{\arxivadded{\tablehead[l]{DeepSeek V4.1\\Flash 0910}}} \\
 & 1 & 1 & 2 & 3 & 1 & 1 & 1 & 1 & 2 & 3 \\ \midrule\endfirsthead
\toprule \tablehead[l]{Task} & \tablehead{Astra} & \multicolumn{3}{c}{\tablehead{Kimi K3}} & \tablehead{GLM} & \arxivadded{\tablehead{Qwen3.8\\Max (0902)}} & \tablehead{Grok} & \multicolumn{3}{c}{\arxivadded{\tablehead[l]{DeepSeek V4.1\\Flash 0910}}} \\
 & 1 & 1 & 2 & 3 & 1 & 1 & 1 & 1 & 2 & 3 \\ \midrule\endhead
\bottomrule\endfoot
AtCoder AHC058 & 100.1 & 96.3 & 97.3 & 97.2 & 96.7 & 97.1 & 95.9 & 98.8 & 97.8 & 98.5 \\
Bin packing & 100.0 & 78.8 & 79.1 & 78.8 & 73.5 & 68.6 & 69.0 & 69.1 & 78.2 & 77.9 \\
Cluster editing & 100.4 & 100.4 & 100.3 & 100.2 & 100.4 & 100.2 & 100.0 & 100.3 & 100.3 & 100.3 \\
Frequency assignment & 100.3 & 99.1 & 100.3 & 100.2 & 100.1 & 100.3 & 100.3 & 100.0 & 100.3 & 100.3 \\
AtCoder AHC041 & 101.3 & 99.6 & 99.1 & 98.2 & 99.2 & 100.1 & 97.3 & 92.1 & 96.8 & 98.4 \\
Graph coloring & 109.7 & 93.4 & 81.2 & 85.4 & 94.0 & 82.7 & 84.7 & 86.2 & 90.9 & 94.4 \\
Course timetabling & 91.9 & 47.7 & 41.3 & 62.2 & 35.0 & 44.8 & 31.3 & 75.6 & 64.7 & -2.9 \\
Capacitated vehicle routing & 97.5 & 82.5 & 87.4 & 64.0 & 76.1 & 86.0 & 84.1 & 91.6 & 95.6 & 96.4 \\
Directed feedback vertex set & 100.0 & 99.0 & 99.3 & 99.6 & 97.6 & 96.1 & 32.6 & 99.6 & 97.4 & 98.0 \\
Dominating set & 87.8 & 90.4 & 97.4 & 83.8 & 59.2 & 73.6 & 66.6 & 75.8 & 91.3 & 85.8 \\
Glass cutting & 100.6 & 89.0 & 92.7 & 95.6 & 84.4 & 75.8 & 62.1 & 78.7 & 78.2 & 82.9 \\
Graph partitioning & 100.5 & 97.2 & 91.6 & 94.5 & 98.3 & 73.0 & 84.9 & 97.9 & 96.9 & 92.8 \\
Hitting set & 92.4 & 95.1 & 85.6 & 74.7 & 95.8 & 87.8 & 76.5 & 75.4 & 67.6 & 100.7 \\
Sports timetabling & 92.2 & 40.7 & 35.9 & 35.7 & 38.5 & 63.8 & 38.1 & 41.9 & 64.6 & -10.1 \\
Job-shop scheduling & 100.1 & 97.9 & 96.4 & 96.0 & 98.0 & 95.2 & 91.3 & 95.1 & 95.1 & 87.7 \\
Logic synthesis & 98.2 & 60.1 & 49.3 & 76.0 & 72.6 & -386.5 & 70.4 & 86.0 & 85.1 & 53.2 \\
Maximum clique & 100.0 & 100.0 & 77.4 & 100.0 & 100.0 & 98.1 & 90.6 & 100.0 & 84.2 & 92.5 \\
Maximum cut & 102.4 & 96.3 & 96.5 & 96.1 & 97.3 & 98.8 & 96.9 & 93.4 & 80.5 & 81.4 \\
Maximum satisfiability & 79.6 & 21.6 & -10.1 & 21.5 & 39.4 & 33.3 & 13.7 & 30.5 & 6.9 & 20.9 \\
Machine reassignment & 112.1 & 80.1 & 78.2 & 105.8 & 103.0 & 102.6 & 95.2 & 57.4 & 61.2 & 68.5 \\
Nurse rostering & 99.7 & 98.3 & 98.7 & 97.9 & 99.5 & 99.5 & 98.5 & 99.1 & 98.9 & 97.5 \\
One-sided crossing minimization & 100.0 & 99.9 & 98.0 & 99.9 & 100.0 & 100.0 & 92.5 & 100.0 & 99.7 & 96.1 \\
Quadratic assignment & 99.1 & 95.3 & 95.9 & 95.8 & 98.3 & 95.3 & 98.0 & 97.8 & 96.4 & 96.1 \\
Quantified Boolean formulas & 66.7 & 33.3 & 33.3 & 12.5 & 37.5 & 41.7 & 29.2 & 37.5 & 41.7 & 37.5 \\
Radiation therapy & 101.8 & 100.7 & 101.8 & 101.6 & 101.8 & 101.4 & 101.4 & 101.4 & 101.8 & 97.1 \\
Set cover & 99.8 & 96.8 & 92.8 & 89.8 & 96.7 & 89.8 & 95.2 & 94.3 & 97.6 & 96.9 \\
Syntax-guided synthesis & 100.0 & 100.0 & 100.0 & 100.0 & 100.0 & 100.0 & 100.0 & 100.0 & 100.0 & 100.0 \\
Traveling salesperson & 99.3 & 46.5 & 81.1 & -25.1 & 77.7 & 28.7 & -45.3 & 70.6 & 80.2 & 16.4 \\
Treewidth & 379.2 & 15.8 & 15.3 & -5.3 & 31.2 & 27.9 & 99.3 & 0.2 & 32.7 & 42.5 \\
Vertex-cover certificates & 100.0 & 29.2 & 22.2 & 98.6 & 30.6 & 65.3 & 88.9 & 18.1 & 8.3 & 55.6 \\
\end{longtable}
\endgroup

%% file: tables/solver-score-examples.tex
\begin{table}[ht]
\ifmarkarxivchanges\color{red}\fi
\centering\small
\caption{Raw objectives and ratio scores for two minimization instances. $B$: baseline; $K$: task reference; $A$: attained objective.}
\label{tab:solverscoreexamples}
\setlength{\tabcolsep}{3pt}
\begin{tabular*}{\linewidth}{@{\extracolsep{\fill}}lllrrrr@{}}
\toprule
Task & Model & Instance & $B$ & $K$ & $A$ & Score (\%) \\
\midrule
Treewidth & GPT-6 Astra & he134 & 249 & 248 & 201 & 4800.0 \\
Logic synthesis & \qwenagent{} & ex37 & 698 & 106 & 273\,020 & -46000.3 \\
\bottomrule
\end{tabular*}
\end{table}

%% file: tables/proof-settings.tex
\begin{table}[htbp]
\centering
\caption{\arxivadded{Complete proofs under bare, ladder, and guided settings. Each agent has three attempts on each of 20 theorems. A theorem is solved if at least one attempt proves it completely.}}
\label{tab:proofsettings}
\setlength{\tabcolsep}{4pt}
\begin{tabular}{llrrr}
\toprule
Model & Outcome & Bare & Ladder & Guided \\
\midrule
Kimi K3 & Successful attempts & 7/60 (11.7\%) & 7/60 (11.7\%) & 18/60 (30.0\%) \\
 & Theorems solved & 3/20 & 4/20 & 7/20 \\
\deepseekagent{} & Successful attempts & 0/60 (0.0\%) & 2/60 (3.3\%) & 11/60 (18.3\%) \\
 & Theorems solved & 0/20 & 2/20 & 6/20 \\
\bottomrule
\end{tabular}
\end{table}

%% file: tables/guidance-task-results.tex
\begin{longtable}{@{}rp{0.32\linewidth}rrrrrr@{}}
\caption{Proof-guidance scores on 20 theorems. Scores are percentages averaged over three attempts; $\Delta$ is guided minus ladder in percentage points.}\label{tab:guidancedetail}\\
\toprule
 & & \multicolumn{3}{c}{Kimi K3} & \multicolumn{3}{c}{\deepseekagent{}} \\
\cmidrule(lr){3-5}\cmidrule(l){6-8}
\# & Theorem & Guided & Ladder & $\Delta$ & Guided & Ladder & $\Delta$ \\ \midrule
\endfirsthead
\toprule
 & & \multicolumn{3}{c}{Kimi K3} & \multicolumn{3}{c}{\deepseekagent{}} \\
\cmidrule(lr){3-5}\cmidrule(l){6-8}
\# & Theorem & Guided & Ladder & $\Delta$ & Guided & Ladder & $\Delta$ \\ \midrule
\endhead
1 & Corners theorem & 25.0 & 20.8 & +4.2 & 25.0 & 25.0 & +0.0 \\
2 & Negations in small-depth circuits & 100.0 & 81.0 & +19.1 & 61.9 & 57.1 & +4.7 \\
3 & Universal Dijkstra & 100.0 & 79.2 & +20.8 & 41.7 & 41.8 & -0.2 \\
4 & Three taxis & 83.7 & 75.0 & +8.7 & 79.2 & 66.7 & +12.5 \\
5 & Monotonicity testing & 91.7 & 58.3 & +33.3 & 70.7 & 62.5 & +8.2 \\
6 & QMA--QCMA oracle separation & 87.5 & 83.3 & +4.2 & 58.3 & 70.8 & -12.5 \\
7 & Discrepancy & 61.9 & 33.3 & +28.6 & 33.3 & 28.6 & +4.8 \\
8 & PCP alphabet soundness & 23.8 & 71.4 & -47.6 & 76.2 & 66.7 & +9.5 \\
9 & Proximity testing & 100.0 & 100.0 & +0.0 & 100.0 & 83.3 & +16.7 \\
10 & Folded Reed--Solomon codes & 100.0 & 66.7 & +33.3 & 91.7 & 91.7 & +0.0 \\
11 & Metrical task systems & 12.5 & 58.3 & -45.8 & 41.7 & 33.3 & +8.3 \\
12 & XOR parallel repetition & 54.0 & 50.0 & +4.0 & 12.5 & 20.8 & -8.3 \\
13 & Unitary synthesis & 83.3 & 54.2 & +29.2 & 71.0 & 66.7 & +4.3 \\
14 & Planar independent-set reconfiguration & 47.6 & 71.4 & -23.8 & 71.4 & 76.2 & -4.8 \\
15 & Positive-body chasing & 100.0 & 88.0 & +12.0 & 100.0 & 54.2 & +45.8 \\
16 & Locally correctable codes & 27.8 & 33.3 & -5.6 & 22.2 & 16.7 & +5.6 \\
17 & Strategy improvement & 66.7 & 37.5 & +29.2 & 45.8 & 25.0 & +20.8 \\
18 & Planar multiflow & 66.7 & 41.7 & +25.0 & 41.7 & 29.2 & +12.5 \\
19 & Discrete bidding games & 20.7 & 50.3 & -29.7 & 37.5 & 20.8 & +16.7 \\
20 & Quantum cryptography & 87.7 & 54.0 & +33.7 & 75.0 & 45.8 & +29.2 \\
\midrule & Mean & 67.0 & 60.4 & +6.6 & 57.8 & 49.1 & +8.7 \\
\bottomrule\end{longtable}

%% file: sections/appendix-f-cases.tex
\section{Submitted Outputs and Case Studies}
\label{app:cases}

The selected cases connect benchmark scores to mathematical parameters and proof obligations. Solver cases illustrate improvements on named instances; construction cases identify attained objects and task references; open-conjecture cases examine verified special cases and remaining proof obligations. Each is interpreted under its originating task version.

\tracksubsection{solver}{Logic-synthesis solvers and public competition references}
\label{app:solvercases}

The solver examples use three logic-synthesis submissions from the designated evaluation. GPT-6 Astra reduces the AND-gate count from 138 to 108 on instance \texttt{ex59}; Kimi K3 reduces it from 19 to 18 on \texttt{ex30}; Grok 4.6 reduces it from 51 to 50 on \texttt{ex68}. The corresponding reductions are \arxivadded{21.7\%, 5.3\%, and 2.0\%}. Each reference is the minimum count across the four teams' publicly released AIG-track circuits in the IWLS 2023 repository \citep{iwls2023}. We downloaded all twelve reference circuits at commit \texttt{7dbfd510e259} and read their AND counts from the AIGER headers. Google's submission attains all three minima, with the other teams tying on \texttt{ex30}. The released files and their checksums accompany the case data.

Each example comes from the final evaluation of the submitted program; its objective and validity were checked against the evaluation records. The evaluation instances permute and negate inputs and outputs of the contest functions. These transformations preserve the AND-gate count because inversion is represented on edges. The examples illustrate individual outcomes, while task averages include all scored instances.

\arxivadded{These comparisons use the publicly released team circuits and TCSExam\textquotesingle s execution budgets (Appendix~\ref{app:solvertasks}). The contest\textquotesingle s aggregate results additionally incorporate unreleased organizer circuits. The three examples therefore report improvements against the specified public circuit references.}

\tracksubsection{construction}{Graph and quantum-code constructions}

The graph task seeks dense 124-vertex graphs without triangles or 4-cycles. \deepseekagent{}'s 630 edges consist of a published 123-vertex, 621-edge seed and nine new edges to an added vertex; the selected neighbors have pairwise seed distance at least three. The seed size and 629-edge reference at 124 vertices appear in Table~1 of \citet{girthbounds}, version 1. The quantum-code record gives $[[25,7,6]]$ binary stabilizer codes versus distance 5 in the designated table entry \citep{codetables25}. These comparisons use the designated task references, with the source versions stated explicitly.

\begin{figure}[t]
\centering
\includegraphics[width=\linewidth]{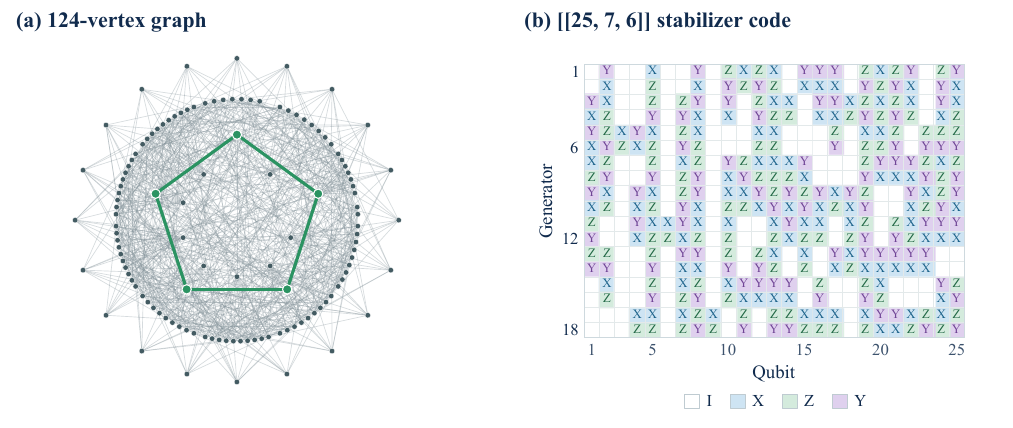}
\caption{\textbf{Two constructions submitted by GPT-6 Astra.} (a) A 124-vertex, 630-edge graph without 3- or 4-cycles, with a shortest cycle highlighted. (b) The 18 submitted stabilizer generators of a $[[25,7,6]]$ quantum code. Both panels use Astra's actual submissions; they attain the same parameters as Kimi K3's constructions in Figure~\ref{fig:researchoutputs}.}
\label{fig:astraconstructions}
\end{figure}

\paragraph{Verification of the displayed constructions.}
Figure~\ref{fig:researchoutputs} uses Kimi K3's graph and quantum code, extracted directly from its submitted formal certificates. Figure~\ref{fig:astraconstructions} shows Astra's corresponding submissions, which attain the same parameters. Independent finite checks confirm that each graph has 124 vertices, 630 unique edges, no triangles, and at most one common neighbor for every vertex pair, excluding 4-cycles. The highlighted cycles are $0,21,32,11,121,0$ for Kimi K3 and $0,2,21,46,18,0$ for Astra, establishing girth exactly 5. Each drawing shows every vertex and edge, with positions chosen for readability.

For each quantum code, the 18 submitted pairs of binary masks are converted directly to Pauli letters, with qubits 1--25 corresponding to bits 0--24. Each matrix retains the submitted row order. Independent checks confirm binary rank 18 and pairwise commutation, yielding $25-18=7$ logical qubits. The archived verifiers certify distance at least 6; independent distance checks corroborate this value. The edge lists, generator masks, source checksums, and drawing checks accompany the two figure data files. \arxivadded{Astra\textquotesingle s generator masks are also available in the \href{https://github.com/MoonshotAI/TCSExam/blob/e0ff9674c1cb58449b950d3c562aee8bce0aa4c1/docs/paper/quantum-code.json}{public case data}.}

\paragraph{A binary-code upper bound.}
Astra's submission proves $A(22,10)\leq82$. Here $A(22,10)$ is the largest possible number of binary words of length 22 with pairwise Hamming distance at least 10. The published semidefinite-programming bound is 84 \citep{gijswijt2012}, also listed in Brouwer's table when checked on September 26, 2026 \citep{brouwerbinary}. The verified upper-bound score is \arxivadded{101.6\%}. The same submission supplies a 64-word code, which gives a separate lower bound. The construction verification protocol is described in Appendix~\ref{app:constructionprotocol}.

\tracksubsection{open}{A verified nine-dimensional Alon--Jaeger--Tarsi result}
\label{app:ajt}

\paragraph{Statement and relation to prior work.}
The Alon--Jaeger--Tarsi conjecture asks whether every invertible matrix over a field with at least four elements admits a vector whose coordinates and image coordinates are all nonzero. Astra's Lean proof establishes
\begin{equation}
\label{eq:ajt-special-case}
  \forall p\geq5\text{ prime},\quad \forall n\leq9,\quad
  \forall A\in\operatorname{GL}_n(\mathbb{F}_p),\quad
  \exists x\in(\mathbb{F}_p\setminus\{0\})^n:\ Ax\in(\mathbb{F}_p\setminus\{0\})^n.
\end{equation}
The contribution relative to the literature reviewed here is the case $p=5$, $n=9$. Yu's dimension guarantee $n<2^{p-2}$ gives $n\leq7$ over $\mathbb{F}_5$ \citep{yu1999permanent}; this guarantee is also stated in the introduction of \citet{nagypach2025ajt}. Version~1 of \citet{zeng2026ajt}, posted August 18, 2026, reports $n\leq2^{p-2}$, including $n=8$ over $\mathbb{F}_5$. The large-field result of \citet{nagypach2025ajt} applies when $61<|\mathbb{F}|\neq79$ and does not cover $\mathbb{F}_5$. The nine-dimensional result goes beyond these ranges. The comparison uses the publicly available versions checked on September 26, 2026. For $p\geq7$, the dimensions in Equation~\ref{eq:ajt-special-case} already fall within Yu's guarantee.

\paragraph{From a minimal counterexample to counting constraints.}
The proof first establishes smaller dimensions and then assumes a minimal counterexample $A$ of dimension nine over $\mathbb{F}_5$. Let $\operatorname{wt}(x)$ denote the number of nonzero coordinates of $x$. Define
\[
  W_{ij}=\bigl|\{x:\operatorname{wt}(x)=i,\ \operatorname{wt}(Ax)=j\}\bigr|,
  \qquad
  V_{ij}=\bigl|\{x:\operatorname{wt}(x)=j,\ \operatorname{wt}(A^{\mathsf T}x)=i\}\bigr|.
\]
These two $10\times10$ arrays have nonnegative integer entries. Invertibility fixes their row and column sums; a finite Fourier identity relates the two arrays. Minimality supplies further vanishing conditions, including $W_{22}=W_{23}=W_{32}=0$.

Additional inequalities come from pairs of columns $u,v$. Each coordinate pair $(u_i,v_i)$ belongs to one of seven classes: both zero, only the first nonzero, only the second nonzero, or one of the four nonzero cancellation ratios $u_i+t v_i=0$. Figure~\ref{fig:ajtcertificate}(a) displays all 25 coordinate pairs. Their seven class counts $h_0,\ldots,h_6$ sum to nine, leaving $\binom{15}{6}=5{,}005$ possible count vectors. The submitted proof checks the relevant inequalities over these finite count vectors and sums them across column pairs. Applying the argument to $A$, $A^{-1}$, $A^{\mathsf T}$, and $(A^{\mathsf T})^{-1}$ yields four nonnegative linear quantities $L_1,\ldots,L_4$ in $W$ and $V$. These finite checks concern coordinate counts, not enumeration of all invertible nine-dimensional matrices.

\paragraph{The submitted contradiction certificate.}
Collect the 200 entries of $W,V$ and the four quantities $L_o$ in a vector $z\geq0$. The identities above give an integer linear system $Ez=b$. The submitted certificate contains integer multipliers $\lambda$ satisfying
\begin{equation}
\label{eq:ajt-certificate}
  \lambda^{\mathsf T}E\geq0,\qquad \lambda^{\mathsf T}b<0.
\end{equation}
Consequently, $0\leq\lambda^{\mathsf T}Ez=\lambda^{\mathsf T}b<0$, a contradiction. Of the 197 equations in the source certificate, 143 have nonzero multipliers; the Lean declaration retains precisely these 143 equations. Independent integer recomputation matches each retained equation, multiplier, and right-hand side to the Lean source and verifies Equation~\ref{eq:ajt-certificate}. All 204 resulting coefficients are nonnegative: 93 are positive and 111 are zero. Figure~\ref{fig:ajtcertificate}(b) shows their exact support, grouping the coefficients on $W$, $V$, and the four pair terms. For its displayed identity, we divide the certificate by $|\lambda^{\mathsf T}b|$ and write $c^{\mathsf T}=\lambda^{\mathsf T}E/|\lambda^{\mathsf T}b|$, giving $0\leq c^{\mathsf T}z=-1$.

\paragraph{Independent formal verification.}
The 71-module source project was rebuilt in a clean sandbox using Lean 4.33.0-rc1 and \arxivadded{public mathlib revision \texttt{5b3f719b}}, without reusing model-generated compiled files. The clean build used a 30\,GB memory limit. The three declarations establishing the result over $\mathbb{F}_5$, its prime-field extension, and the minimum counterexample dimension depend only on \texttt{propext}, \texttt{Classical.choice}, and \texttt{Quot.sound}. Their dependency audits contain no \texttt{sorryAx} or additional axioms and do not depend on the unresolved conjecture slots. The definition of the required nowhere-zero vector agrees with the original task. Separate integer checks reproduce the certificate and the inequalities over the 5,005 class-count vectors.

\input{sections/appendix-f-additional-results}

\tracksubsection{open}{An unresolved Paley graph conjecture attempt}
\label{app:openanalysis}

\begin{figure}[t]
\setlength{\parskip}{0pt}
\setlength{\abovecaptionskip}{5pt}
\centering
\arxivfigure{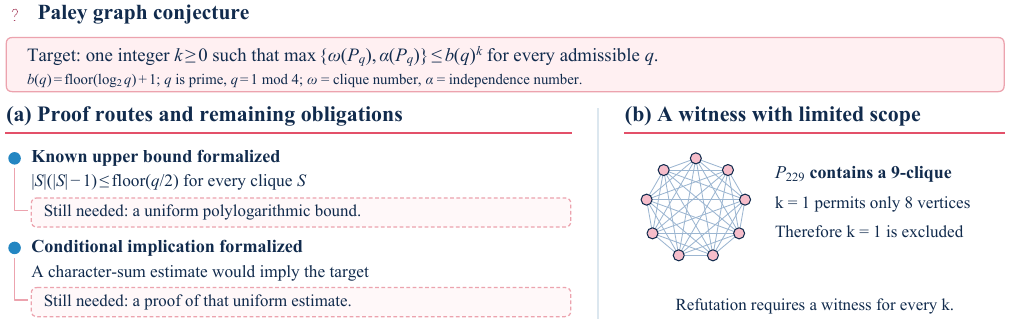}
\caption{\textbf{An unresolved Paley graph attempt.} (a) Auxiliary statements and remaining obligations. (b) A witness excluding $k=1$ only. All 36 edges of the nine-vertex clique are independently verified.}
\label{fig:opendiagnostics}
\end{figure}

Figure~\ref{fig:opendiagnostics} examines one GPT-6 Astra attempt on the Paley graph conjecture from the aggregate evaluation. In the Paley graph $P_q$, vertices are residues modulo a prime $q\equiv1\pmod4$; distinct vertices are adjacent when their difference is a square modulo $q$. Write $b(q)=\lfloor\log_2q\rfloor+1$. The target asks whether some fixed nonnegative integer $k$ bounds both the clique and independence numbers by $b(q)^k$ for every admissible prime.

\paragraph{A concrete witness.} The submitted declaration \texttt{paley229\_nine\_clique} gives the vertex set
\[
\{0,1,57,71,135,154,203,215,218\}\subseteq\mathbb{Z}/229\mathbb{Z}.
\]
Its 36 unordered pairs all have square differences. \arxivadded{Independent computation confirms this finite property; Lean dependency checks verify this declaration and \texttt{exponent\_one\_fails} using only the permitted standard axioms.} \arxivadded{Since $b(229)=8$, this finite witness establishes that any uniform exponent satisfying the target must be at least two.}

\paragraph{Proof routes and missing obligations.} The submitted polynomial-method development contains \texttt{hanson\_petridis\_bound}, stating $|S|(|S|-1)\leq\lfloor q/2\rfloor$ for any clique $S$. This reproduces the square-root-scale bound of \citet{hansonpetridis}. \arxivadded{The resulting upper bound grows on the order of $\sqrt q$, while the target requires a bound polynomial in $\log q$. Closing this difference is the remaining step along this proof route.}

The character-sum development proves a conditional implication. Let $\chi_q$ be the quadratic character, equal to 0 at zero, 1 at nonzero squares, and $-1$ otherwise. For a fixed $k$, the required hypothesis is
\[
  2\sum_{x,y\in S}\chi_q(x-y)\leq |S|^2
  \quad\text{whenever } |S|\geq b(q)^k+3,
\]
uniformly over admissible primes $q$ and sets $S$. \arxivadded{The submission derives the target with exponent $k+2$ conditional on this uniform estimate, which remains an open proof obligation.} The accompanying research notes record this missing obligation alongside unsuccessful attempts to apply the polynomial bound and an interval-clique construction.

\paragraph{Verification and scope.} The clique witness is checked both in Lean and by independent finite computation. The auxiliary statements have only the permitted standard axioms as dependencies. \arxivadded{The attempt receives a complete-resolution score of zero. Figure~\ref{fig:opendiagnostics} identifies the verified finite witness, conditional implication, and remaining uniform-bound obligation.}

\subsection{Evidence for reference improvements and novelty claims}
\label{app:witness}

Artifact validity and novelty are recorded separately. The archive retains the original output, its checksum, task version, verifier and toolchain, verification log, recomputed value or accepted statement, and evaluation resources. A record comparison additionally includes the original source, date and location, prior value, parameter restrictions, and applicable resource conditions. A floating-point difference within the declared tolerance is not counted as an improvement.

\arxivadded{Case studies identify their comparison explicitly: a task reference, a public competition result, or a published mathematical bound. Claims of new records include a source comparison matching the problem parameters and applicable resource conditions. A claimed resolution of an open conjecture additionally undergoes renewed checking of the formal statement against the source conjecture.}

%% file: sections/appendix-f-additional-results.tex
\tracksubsection{open}{Additional verified results from open-conjecture attempts}
\label{app:additionalopen}

\arxivadded{We independently verify four further mathematical results from Astra\textquotesingle s open-conjecture attempts: explicit inequalities, an exact rigidity formula, and finite-instance matching bounds. Clean source builds and axiom audits accept these statements using Lean\textquotesingle s standard axioms, independently of the unresolved conjecture slots. The statements and proof methods are given below.}

\paragraph{A six-variable Fourier entropy--influence bound.}
For every Boolean function $f:\{-1,1\}^{6}\to\{-1,1\}$, the submitted proof establishes
\[
 H(\widehat f^{\,2})\leq (7\ln 2)\,I(f).
\]
Here $H$ is the Shannon entropy of the squared Fourier coefficients, using natural logarithms, and $I(f)=\sum_i\Pr[f(X)\ne f(X^{\oplus i})]$ is the total influence under uniform $X$. Equivalently, the constant is 7 when entropy is measured in bits. The proof reduces the bound to finite integer inequalities using the discrete values of Walsh coefficients, congruence constraints, and Parseval's identity. It covers all six-variable Boolean functions without enumerating their $2^{64}$ truth tables.

\paragraph{Exact Hilbert-matrix rigidity at high target ranks.}
Let $H_n$ be the rational Hilbert matrix with entries $(H_n)_{ij}=1/(i+j+1)$ for $0\leq i,j<n$, and let $R_{H_n}(r)$ be the minimum number of entries that must change to obtain rank at most $r$. The submitted proof establishes the exact formula
\[
 R_{H_n}(r)=2(n-r)-1,
 \qquad \lfloor n/2\rfloor\leq r<n.
\]
The upper bound constructs a rational matrix of rank exactly $r$ differing in exactly $2(n-r)-1$ entries. The lower bound combines a row-and-column covering argument with nonsingularity of Hilbert submatrices. \arxivadded{This formula characterizes the high-target-rank regime; the task\textquotesingle s conjectured lower bound concerns lower target ranks.}

\paragraph{An explicit influence bound for quadratic polynomials.}
For any real polynomial $p$ of total degree at most two satisfying $0\leq p(x)\leq1$ on $\{-1,1\}^n$ and $\operatorname{Var}(p)>0$, the submitted proof finds a coordinate $i$ with
\[
 \operatorname{Inf}_i(p)\geq\frac{\operatorname{Var}(p)^2}{16},
 \qquad
 \operatorname{Inf}_i(p)=\mathbb E\bigl[(p(X)-p(X^{\oplus i}))^2\bigr].
\]
It also derives $\operatorname{Inf}_i(p)\geq(\varepsilon/2)^3$ whenever $\operatorname{Var}(p)\geq\varepsilon>0$. The bound is independent of the number of variables.

\paragraph{Finite-instance bounds for random-order online matching.}
The formal model reveals each online vertex's neighbors upon its arrival in a uniformly random permutation. Algorithms may adapt to the observed history and use an arbitrary finite number of independent uniform random bits. For graphs with maximum matching size at most two, an explicit algorithm guarantees a ratio of $7/8$, and a two-by-two instance family gives a matching upper bound. On graphs with three vertices on each side, another algorithm attains $5/6$, again with a matching upper bound.

For every algorithm in this formal model, a five-by-five graph $G$ with $\operatorname{OPT}(G)=5$ satisfies
\[
 \frac{\mathbb E[\operatorname{ALG}(G)]}{\operatorname{OPT}(G)}
 \leq \frac{6788069}{8329500}\approx0.814943154.
\]
The proof uses a weighted family of 570 graphs and their online-vertex relabelings. A certificate bounds the weighted expected matching size by $27{,}152{,}276$ at total weight $6{,}663{,}600$, yielding the displayed ratio after division by five. Padding with isolated vertices extends the obstruction to larger vertex counts while preserving $\operatorname{OPT}=5$. \arxivadded{The bound applies to this algorithm class under a strict multiplicative guarantee with zero additive error.}

%% file: sections/appendix-g-maintenance.tex
\section{Extension and Maintenance}
\label{app:maintenance}

\subsection{Adding a task}

An extension specifies the research question and source, submission type, accepted evidence, scoring rule, and execution resources. It then follows the development checks in Appendix~\ref{app:fidelity}: source correspondence, executable validation, negative tests, trial evaluation, and expert review. The task manifest records its field and parameter coverage so that extensions can increase both breadth and depth without changing the meaning of existing scores.

New problems use the corresponding output interface: a formal proof, a reusable solver, an explicit object with verified properties, or a proof or refutation of a conjecture. A task combining several methods still specifies exactly what must be submitted and what the checker establishes. This permits research workflows to vary while the evaluated output remains explicit.

\subsection{Versioning and corrections}

Statements, environments, scorers, instance sets, and references remain fixed within an evaluation version. Corrections and additional tasks receive new versions with dated explanations. Reports identify which results are affected and whether re-evaluation is necessary. Source corrections can update a reference value in a later release with provenance; a model improvement is retained as an outcome rather than silently changing the comparison used for that evaluation.

Retired tasks remain available with their historical version and retirement reason. Comparisons across releases report the common task set or explicitly identify the changed collection. A maintained benchmark can thereby grow without rewriting earlier results.

\subsection{Recording research outputs}

The project maintains an artifact index linking results to the statements or objects checked, their originating task versions, and verification records. Research status distinguishes reproduction, validated improvement relative to a specified reference, and a novelty claim supported by source review. Verified special cases and reductions can be indexed alongside the complete-resolution score, preserving their mathematical content without changing that score retrospectively.

%% file: sections/appendix-h-lite.tex
\clearpage
\begin{arxivaddition}
\section{Lite Selection, Results, and Full Comparison}
\label{app:lite}

\subsection{Task selection}

We selected five tasks per track, balancing subject coverage, difficulty, and evaluation cost. Using the original six agents' results, we compared candidate subsets by how well they preserved pairwise agent orderings and approximated Full track scores, then reviewed their mathematical and computational content to finalize the list. Selected tasks retain their original scoring rules and receive equal weight within each track. The list was fixed before evaluating Kimi K3.1 Pro and \deepseeknewagent{}. The Index in Equation~\ref{eq:index} is a later reporting summary, applied with the same weights to both suites.

\begin{longtable}{@{}p{0.16\linewidth}p{0.38\linewidth}p{0.39\linewidth}@{}}
\caption{Subject coverage of the 20 TCSExam-Lite tasks.}\label{tab:litetasks}\\
\toprule Track & Task & Subject \\
\midrule\endfirsthead
\toprule Track & Task & Subject \\
\midrule\endhead
\trackmargin{theorem}Theorem & Corners theorem & Additive combinatorics \\
& Discrete bidding games & Games and algorithms \\
& Folded Reed--Solomon codes & Coding theory \\
& Small-depth negations & Circuit complexity \\
& Quantum cryptography & Quantum computation \\
\midrule
\trackmargin{solver}Solver & Bin packing & Packing and integer optimization \\
& Course timetabling & Scheduling \\
& Maximum cut & Graph optimization \\
& Machine reassignment & Resource assignment \\
& Quantified Boolean formulas & Certified logical reasoning \\
\midrule
\trackmargin{construction}Construction & Binary code $A(24,10)$ & Coding theory \\
& Graphs without 3- or 4-cycles & Extremal graph theory \\
& Kissing number in dimension 13 & Discrete geometry \\
& Ramsey $R(4,6)$ & Ramsey theory \\
& Metric TSP integrality gap & Approximation algorithms \\
\midrule
\trackmargin{open}Open conjecture & Evasiveness & Graph query complexity \\
& Frege proof length & Proof complexity \\
& Log-rank & Communication complexity \\
& Random-order online matching & Online algorithms \\
& Constant-width branching programs & Pseudorandomness \\
\bottomrule
\end{longtable}

The theorem subset includes both tasks solved by several agents and tasks with no accepted proof in the original cohort. Solver tasks span several types of input and certificate. Construction tasks cover discrete and geometric objects as well as an approximation bound. The open-conjecture subset gives coverage across five research directions. These choices retain variation within a five-task track while reducing the total number of task attempts by 80\% relative to Full.

\subsection{Result provenance and agent configurations}
\label{app:lite-provenance}

Full contains 1,600 task attempts: 1,000 from the original six agents and 300 each from Kimi K3.1 Pro and \deepseeknewagent{}. Lite contains 340 entries across nine agents. Of these, 240 are the 15 shared non-Solver tasks across the eight common agents, reusing the same submissions in both suite summaries. The remaining 100 entries comprise 80 separate Lite Solver attempts and the 20 Opus attempts.

Kimi K3.1 Pro and DeepSeek 1017 each contribute three repetitions. For each agent, Full combines 45 shared non-Solver attempts with three repetitions on the remaining 85 tasks, including all 30 Online Solver problems. The 15 Offline Solver attempts contribute only to Lite. Every repetition contributes to the task means.

Opus uses Claude Code 2.1.283 with the one-million-token context option and maximum reasoning effort. Table~\ref{tab:systems} gives the configurations of the other agents.

All shared non-Solver scores are equal between the two suite records. Solver comparison uses the same fixed Kimi K3 reference, the original task score definitions, and each suite's recorded development protocol.

\subsection{Ranking comparison}
\label{app:lite-ranking}

We compare the eight agents with both Full and Lite results. Spearman's rank correlation is $\rho=0.93$, and 26 of the 28 unordered pairs have the same Index order. The two reversals are DeepSeek 1017 versus GLM-5.3 and \qwenagent{}: DeepSeek 1017 is fourth among the shared agents on Full and sixth on Lite. Astra, Kimi K3.1 Pro, and Kimi K3 retain the top three positions. Excluding DeepSeek 1017, the other seven agents retain their full pairwise order. The Lite task list was selected using the original six agents and fixed before either Kimi K3.1 Pro or \deepseeknewagent{} was evaluated. Opus is the ninth agent on the Lite leaderboard.

\input{tables/full-lite-rank-agreement}

Theorem rankings reverse DeepSeek 1017 against GLM-5.3, \qwenagent{}, and Grok 4.6. Solver reverses six pairs: Kimi K3.1 Pro versus \qwenagent{}, DeepSeek 1017, and DeepSeek 0910; Kimi K3 versus GLM-5.3 and DeepSeek 0910; and GLM-5.3 versus DeepSeek 0910. Construction reverses Kimi K3.1 Pro versus GLM-5.3, and Kimi K3 versus \qwenagent{} and DeepSeek 1017. All Lite open-conjecture scores are zero, whereas Full has accepted refutations on two tasks outside the subset. These results quantify the aggregate agreement and the capability-specific differences under the two suite protocols.

\subsection{Complete Lite task results}

The following tables retain every selected task. Theorem and open-conjecture cells give successes over attempts; Solver and construction scores are percentages rounded to one decimal. Agent labels follow Tables~\ref{tab:mainresults} and~\ref{tab:literesults}.

\Needspace{14\baselineskip}
\input{tables/lite-task-results-arxiv}

\begin{figure}[p]
\setlength{\parskip}{0pt}
\setlength{\abovecaptionskip}{5pt}
\centering
\arxivfigure{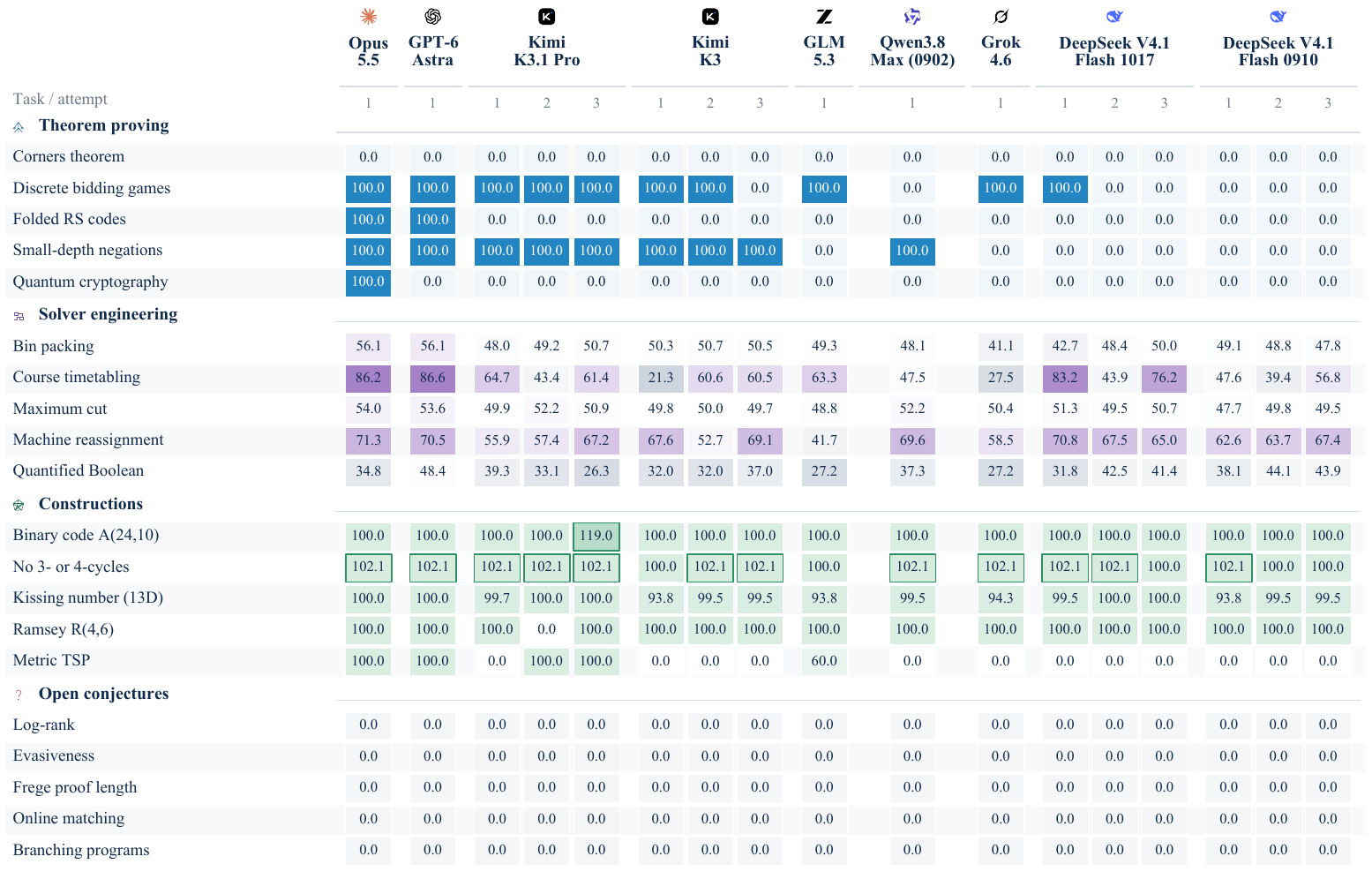}
\caption{\arxivadded{\textbf{All 340 Lite task attempts across nine agents.} Columns number attempts within each agent. All scores are percentages rounded to one decimal. Theorem and open-conjecture outcomes are 0.0 or 100.0; construction outlines mark unrounded scores above the 100.0 reference.}}
\label{fig:liteattempts}
\end{figure}
\end{arxivaddition}

%% file: tables/full-lite-rank-agreement.tex
\begin{table}[ht]\centering\small\begin{arxivaddition}
\caption{Full--Lite ranking agreement across 8 shared agents. Each row compares 28 agent pairs using unrounded scores.}
\label{tab:rankagreement}
\begin{tabular}{lrrrr}\toprule
Metric & Same strict order & Reversed & Tied in both & Tied in one \\
\midrule
Index & 26 & 2 & 0 & 0 \\
Theorem proving & 22 & 3 & 3 & 0 \\
Solver engineering & 22 & 6 & 0 & 0 \\
Constructions & 25 & 3 & 0 & 0 \\
Open conjectures & 0 & 0 & 15 & 13 \\
\bottomrule\end{tabular}\end{arxivaddition}\end{table}

%% file: tables/lite-task-results-arxiv.tex
\begingroup\footnotesize\setlength{\tabcolsep}{2.2pt}
\begin{longtable}{@{}>{\raggedright\arraybackslash}p{0.20\linewidth}rrrrrrrrr@{}}
\caption{Theorem-proving results on TCSExam-Lite. Entries give complete proofs / attempts.}\label{tab:lite-theorem} \\
\toprule
\tablehead[l]{Task} & \tablehead{Opus\\5.5} & \tablehead{GPT-6\\Astra} & \tablehead{Kimi\\K3.1 Pro} & \tablehead{Kimi\\K3} & \tablehead{GLM-5.3} & \arxivadded{\tablehead{Qwen3.8\\Max (0902)}} & \tablehead{Grok\\4.6} & \tablehead{DeepSeek\\V4.1 Flash\\1017} & \arxivadded{\tablehead{DeepSeek\\V4.1 Flash\\0910}} \\
\midrule\endfirsthead
\toprule
\tablehead[l]{Task} & \tablehead{Opus\\5.5} & \tablehead{GPT-6\\Astra} & \tablehead{Kimi\\K3.1 Pro} & \tablehead{Kimi\\K3} & \tablehead{GLM-5.3} & \arxivadded{\tablehead{Qwen3.8\\Max (0902)}} & \tablehead{Grok\\4.6} & \tablehead{DeepSeek\\V4.1 Flash\\1017} & \arxivadded{\tablehead{DeepSeek\\V4.1 Flash\\0910}} \\
\midrule\endhead
Corners theorem & 0/1 & 0/1 & 0/3 & 0/3 & 0/1 & 0/1 & 0/1 & 0/3 & 0/3 \\
Discrete bidding games & 1/1 & 1/1 & 3/3 & 2/3 & 1/1 & 0/1 & 1/1 & 1/3 & 0/3 \\
Folded RS codes & 1/1 & 1/1 & 0/3 & 0/3 & 0/1 & 0/1 & 0/1 & 0/3 & 0/3 \\
Small-depth negations & 1/1 & 1/1 & 3/3 & 3/3 & 0/1 & 1/1 & 0/1 & 0/3 & 0/3 \\
Quantum cryptography & 1/1 & 0/1 & 0/3 & 0/3 & 0/1 & 0/1 & 0/1 & 0/3 & 0/3 \\
\bottomrule\end{longtable}
\begin{longtable}{@{}>{\raggedright\arraybackslash}p{0.20\linewidth}rrrrrrrrr@{}}
\caption{Solver results on TCSExam-Lite. Entries are mean Kimi K3-relative scores (\%).}\label{tab:lite-solver} \\
\toprule
\tablehead[l]{Task} & \tablehead{Opus\\5.5} & \tablehead{GPT-6\\Astra} & \tablehead{Kimi\\K3.1 Pro} & \tablehead{Kimi\\K3} & \tablehead{GLM-5.3} & \arxivadded{\tablehead{Qwen3.8\\Max (0902)}} & \tablehead{Grok\\4.6} & \tablehead{DeepSeek\\V4.1 Flash\\1017} & \arxivadded{\tablehead{DeepSeek\\V4.1 Flash\\0910}} \\
\midrule\endfirsthead
\toprule
\tablehead[l]{Task} & \tablehead{Opus\\5.5} & \tablehead{GPT-6\\Astra} & \tablehead{Kimi\\K3.1 Pro} & \tablehead{Kimi\\K3} & \tablehead{GLM-5.3} & \arxivadded{\tablehead{Qwen3.8\\Max (0902)}} & \tablehead{Grok\\4.6} & \tablehead{DeepSeek\\V4.1 Flash\\1017} & \arxivadded{\tablehead{DeepSeek\\V4.1 Flash\\0910}} \\
\midrule\endhead
Bin packing & 56.1 & 56.1 & 49.3 & 50.5 & 49.3 & 48.1 & 41.1 & 47.0 & 48.5 \\
Course timetabling & 86.2 & 86.6 & 56.5 & 47.5 & 63.3 & 47.5 & 27.5 & 67.8 & 47.9 \\
Maximum cut & 54.0 & 53.6 & 51.0 & 49.8 & 48.8 & 52.2 & 50.4 & 50.5 & 49.0 \\
Machine reassignment & 71.3 & 70.5 & 60.2 & 63.1 & 41.7 & 69.6 & 58.5 & 67.8 & 64.6 \\
Quantified Boolean & 34.8 & 48.4 & 32.9 & 33.7 & 27.2 & 37.3 & 27.2 & 38.6 & 42.0 \\
\bottomrule\end{longtable}
\begin{longtable}{@{}>{\raggedright\arraybackslash}p{0.20\linewidth}rrrrrrrrr@{}}
\caption{Construction results on TCSExam-Lite. Entries are mean task scores (\%).}\label{tab:lite-construction} \\
\toprule
\tablehead[l]{Task} & \tablehead{Opus\\5.5} & \tablehead{GPT-6\\Astra} & \tablehead{Kimi\\K3.1 Pro} & \tablehead{Kimi\\K3} & \tablehead{GLM-5.3} & \arxivadded{\tablehead{Qwen3.8\\Max (0902)}} & \tablehead{Grok\\4.6} & \tablehead{DeepSeek\\V4.1 Flash\\1017} & \arxivadded{\tablehead{DeepSeek\\V4.1 Flash\\0910}} \\
\midrule\endfirsthead
\toprule
\tablehead[l]{Task} & \tablehead{Opus\\5.5} & \tablehead{GPT-6\\Astra} & \tablehead{Kimi\\K3.1 Pro} & \tablehead{Kimi\\K3} & \tablehead{GLM-5.3} & \arxivadded{\tablehead{Qwen3.8\\Max (0902)}} & \tablehead{Grok\\4.6} & \tablehead{DeepSeek\\V4.1 Flash\\1017} & \arxivadded{\tablehead{DeepSeek\\V4.1 Flash\\0910}} \\
\midrule\endhead
Binary code A(24,10) & 100.0 & 100.0 & 106.3 & 100.0 & 100.0 & 100.0 & 100.0 & 100.0 & 100.0 \\
No 3- or 4-cycles & 102.1 & 102.1 & 102.1 & 101.4 & 100.0 & 102.1 & 102.1 & 101.4 & 100.7 \\
Kissing number (13D) & 100.0 & 100.0 & 99.9 & 97.6 & 93.8 & 99.5 & 94.3 & 99.8 & 97.6 \\
Ramsey R(4,6) & 100.0 & 100.0 & 66.7 & 100.0 & 100.0 & 100.0 & 100.0 & 100.0 & 100.0 \\
Metric TSP & 100.0 & 100.0 & 66.7 & 0.0 & 60.0 & 0.0 & 0.0 & 0.0 & 0.0 \\
\bottomrule\end{longtable}
\begin{longtable}{@{}>{\raggedright\arraybackslash}p{0.20\linewidth}rrrrrrrrr@{}}
\caption{Open-conjecture results on TCSExam-Lite. Accepted proofs or refutations / attempts.}\label{tab:lite-open} \\
\toprule
\tablehead[l]{Task} & \tablehead{Opus\\5.5} & \tablehead{GPT-6\\Astra} & \tablehead{Kimi\\K3.1 Pro} & \tablehead{Kimi\\K3} & \tablehead{GLM-5.3} & \arxivadded{\tablehead{Qwen3.8\\Max (0902)}} & \tablehead{Grok\\4.6} & \tablehead{DeepSeek\\V4.1 Flash\\1017} & \arxivadded{\tablehead{DeepSeek\\V4.1 Flash\\0910}} \\
\midrule\endfirsthead
\toprule
\tablehead[l]{Task} & \tablehead{Opus\\5.5} & \tablehead{GPT-6\\Astra} & \tablehead{Kimi\\K3.1 Pro} & \tablehead{Kimi\\K3} & \tablehead{GLM-5.3} & \arxivadded{\tablehead{Qwen3.8\\Max (0902)}} & \tablehead{Grok\\4.6} & \tablehead{DeepSeek\\V4.1 Flash\\1017} & \arxivadded{\tablehead{DeepSeek\\V4.1 Flash\\0910}} \\
\midrule\endhead
Log-rank & 0/1 & 0/1 & 0/3 & 0/3 & 0/1 & 0/1 & 0/1 & 0/3 & 0/3 \\
Evasiveness & 0/1 & 0/1 & 0/3 & 0/3 & 0/1 & 0/1 & 0/1 & 0/3 & 0/3 \\
Frege proof length & 0/1 & 0/1 & 0/3 & 0/3 & 0/1 & 0/1 & 0/1 & 0/3 & 0/3 \\
Online matching & 0/1 & 0/1 & 0/3 & 0/3 & 0/1 & 0/1 & 0/1 & 0/3 & 0/3 \\
Branching programs & 0/1 & 0/1 & 0/3 & 0/3 & 0/1 & 0/1 & 0/1 & 0/3 & 0/3 \\
\bottomrule\end{longtable}
\endgroup